\pdfoutput=1
\documentclass[11pt,a4paper,oneside]{report} 
\usepackage[ngerman,american]{babel}
\usepackage[utf8]{inputenc}
\usepackage[T1]{fontenc}   
\usepackage{textcomp}   
\usepackage[hyphens]{url}
\usepackage{color} 
\usepackage{microtype}
\usepackage{relsize}
\usepackage{amsmath,amsthm,amssymb,pifont,mathtools,wasysym}
\usepackage{mathpazo}  
\usepackage{csquotes}
\usepackage[all]{nowidow}

\usepackage[
	backend=biber,
	style=authoryear-comp,
	 dashed=false,
	 sorting=nyt,
	 natbib=false,
	 doi=false,
	 url=false,
	 isbn=false,
	 backref=true,
	 maxcitenames=2,
	 hyperref=true,
	 uniquename=false
]{biblatex}
\undef\cite{} 
\DefineBibliographyExtras{english}{}
\bibliography{./ms.bib}

\usepackage{graphicx}
\usepackage{subfig}
\usepackage{listings}
\usepackage[usenames,dvipsnames,table]{xcolor}
\usepackage{float}
\usepackage{algorithm}
\usepackage[noend]{algpseudocode}
\usepackage{enumitem}
\usepackage{pdflscape}

\usepackage{eso-pic}

\usepackage[paper=a4paper,width=14cm,left=35mm,height=22cm]{geometry}
\usepackage{setspace}
\newcommand{\phv}{\fontfamily{phv}\fontseries{m}\fontsize{9}{11}\selectfont}
\newcommand{\HRule}{\rule{\linewidth}{0.5mm}}
\usepackage{fancyhdr}
\usepackage{multirow,multicol}
\usepackage{booktabs}
\usepackage{rotating}
\usepackage{tabularx}

\usepackage{afterpage}

\usepackage{algpseudocode}
\usepackage{algorithm}

\newcommand*{\ngram}{$n$-gram~}
\newcommand*{\ngrams}{$n$-grams~}
\newcommand*{\lone}{$L_1$~}
\newcommand*{\art}{state-of-the-art~}
\newcommand*{\ltwo}{$L_2$~}
\newcommand*{\pulse}{\textit{P\textsmaller{ULSE}}~}
\newcommand*{\pypulse}{\textit{PyPulse}~}
\newcommand*{\nplus}{N$^+$~}
\newcommand*{\eg}{for example~}
\newcommand*{\Eg}{For example~}
\newcommand*{\ie}{that is~}

\newcommand*{\fig}{figure~}
\newcommand*{\Fig}{Figure~}
\newcommand*{\Tab}{Table~}
\newcommand*{\tab}{table~}
\newcommand*{\eq}{equation~}

\newcommand{\otoprule}{\midrule[\heavyrulewidth]} 
\newcommand*{\OneS}{\includegraphics[width=.02\linewidth]{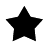}}
\newcommand*{\OneStar}{\OneS}
\newcommand*{\TwoStar}{\OneS\OneS}
\newcommand*{\ThreeStar}{\OneS\OneS\OneS}
\newcommand*{\FourStar}{\OneS\OneS\OneS\OneS}

\usepackage{hyperref}
\hypersetup{
		colorlinks,
		citecolor=black,
		filecolor=black,
		linkcolor=black,
		urlcolor=black,
		pdfinfo={
			Author={Jonas Langhabel},
			Subject={Learning a Predictive Model for Music Using PULSE}
		}
	}

\begin{document}
\begin{titlepage}
	
\begin{center}
\includegraphics[width=.6\linewidth,keepaspectratio]{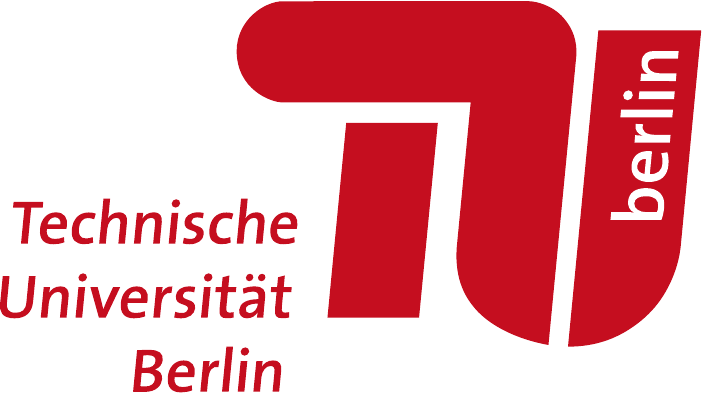}
\vfill
\HRule \\[0.5cm]
\begin{spacing}{1.7}
	{ \huge \bfseries Learning a Predictive Model for Music
		Using \pulse{}}\\[0.2cm]
\end{spacing}

\HRule \\[1.0cm]

Master Thesis in Computer Science

\bigskip
by

\smallskip
{\LARGE \textbf{Jonas Langhabel}}

\bigskip

{\large September 2017}

\bigskip

\vfill
\bigskip
supervised by
\smallskip

{\LARGE \textbf{Robert Lieck}}\\
\textbf{Machine Learning and Robotics Lab, University of Stuttgart \\ Systematic Musicology and Music Cognition, TU Dresden}
\smallskip

{\LARGE \textbf{Prof. Dr. Klaus-Robert Müller}}\\
\textbf{Machine Learning Group, TU Berlin}
\bigskip

reviewed by
\smallskip

{\LARGE \textbf{Prof. Dr. Klaus-Robert Müller}}\\
\textbf{Machine Learning Group, TU Berlin}
\smallskip

{\LARGE \textbf{Prof. Dr. Marc Toussaint}}\\
\textbf{Machine Learning and Robotics Lab, University of Stuttgart}


\end{center}

\end{titlepage}
\pagebreak 

\thispagestyle{empty}

\section*{
	\huge \begin{center}
		Statement in Lieu of an Oath
	\end{center}}
I hereby confirm that I have written this thesis on my own without illegitimate help and that I have not used any other media or materials than the ones referred to in this thesis.

\section*{
	\huge \begin{center}
		Eidesstattliche Erkl\"arung
\end{center}}
Hiermit erkl\"are ich, dass ich die vorliegende Arbeit selbstst\"andig und eigenh\"andig sowie ohne unerlaubte fremde Hilfe und ausschließlich unter Verwendung der aufgef\"uhrten Quellen und Hilfsmittel angefertigt habe.

\vfill
\begin{tabular}{lp{4cm}lp{1cm}lp{4cm}} \\
Berlin,  &   &        & &        \\
 \cline{2-2} \cline{6-6}
& (Date/Datum) & &    & & (Signature/Unterschrift) \\
\end{tabular} 
\thispagestyle{empty}

\section*{\huge \begin{center}
		Acknowledgements
\end{center}}
First and foremost, I would like to express my gratitude to Robert Lieck who provided me with this intriguing topic, put an extraordinary amount of time in my supervision, and was a great teacher to me. I would further like to thank Prof. Dr. Klaus-Robert Müller and Prof. Dr. Marc Toussaint for their supervision and support of this thesis. Special thanks go to Prof. Dr. Martin Rohrmeier for his musicological mentoring, for the invitations to his chair for Systematic Musicology and Music Cognition at TU Dresden, and for his patronage of our research paper about my work. I am grateful to my reviewers Deborah Fletcher, Christian Gerhorst, Jannik Wolff, and Malte Schwarzer for their valuable comments. Finally, I would like to thank my wife for her support that allowed me to put all my focus on this thesis, and my daughter for making my breaks worthwhile.

\selectlanguage{american}

\thispagestyle{empty}
\section*{\huge \begin{center}
	Abstract
\end{center}}
Predictive models for music are studied by researchers of algorithmic composition, the cognitive sciences and machine learning. They serve as base models for composition, can simulate human prediction and provide a multidisciplinary application domain for learning algorithms. A particularly well established and constantly advanced subtask is the prediction of monophonic melodies. As melodies typically involve non-Markovian dependencies their prediction requires a capable learning algorithm.

In this thesis, I apply the recent feature discovery and learning method \pulse to the realm of symbolic music modeling. \pulse is comprised of a feature generating operation and $L_1$-regularized optimization. These are used to iteratively expand and cull the feature set, effectively exploring feature spaces that are too large for common feature selection approaches. I design a general Python framework for \textit{P\textsmaller{ULSE}}, propose task-optimized feature generating operations and various music-theoretically motivated features that are evaluated on a standard corpus of monophonic folk and chorale melodies. The proposed method significantly outperforms comparable \art models. I further discuss the free parameters of the learning algorithm and analyze the feature composition of the learned models. The models learned by \pulse afford an easy inspection and are musicologically interpreted for the first time.

\newpage
\thispagestyle{empty}
\selectlanguage{ngerman}

\section*{\huge \begin{center}
		Zusammenfassung
\end{center}}
Prädiktive Modelle für Musik sind Gegenstand der Forschung in den Feldern der algorithmischen Komposition, der Kognitionswissenschaft und des maschinellen Lernens. Die Modelle liefern eine Basis zum Komponieren, sie können menschliches Verhalten vorhersagen und sie bieten eine interdisziplinäre Anwendung für Lernalgorithmen. Ein aussergewöhnlich beliebter und ständig vorangetriebener Teilbereich des prädiktiven Modellierens ist die Vorhersage von monophonen Melodien. Da Melodien typischerweise nicht-Markovsche Abhängigkeiten mit sich bringen, erfordert ihre Prädiktion besonders leistungsfähige Lernalgorithmen.

In dieser Thesis wende ich die kürzlich entwickelte \pulse Methode an, um symbolische Musik zu modellieren. \pulse ist eine Methode zum Aufspüren und Lernen der geeignetsten Merkmale. Dazu werden abwechselnd neue Merkmale generiert und die global Besten mithilfe von $L_1$-regularisierter Optimierung ausgewählt. Dadurch können Merkmalsräume durchsucht werden, die zu groß für gängige Merkmal-Auswahlverfahren sind. Ich entwerfe ein Python Framework für die \pulse Methode und geeignete generierende Operationen für die Erzeugung von Merkmalen für Melodien sowie zahlreiche musiktheoretisch motivierte Merkmalstypen. Die erlernten Modelle werden auf einem etablierten Korpus monophoner Volksmusik und monophoner Choräle evaluiert; die vorgestellte Methode übertrifft deutlich die besten vergleichbaren Modelle. Weiterhin diskutiere ich die freien Parameter des Lernalgorithmus und analysiere die Merkmal-Zusammensetzung der gelernten Modelle. Die mit \pulse gelernten Modelle sind einfach inspizierbar und werden zum ersten Mal musikwissenschaftlich interpretiert.

\selectlanguage{american}

\setcounter{page}{6}
\begin{spacing}{1.05}
	\tableofcontents
\end{spacing}
\newpage

\chapter*{Acronyms \& Abbreviations}
\begin{tabular}{l l}
	BWV & Bach-Werke-Verzeichnis\\
	CRF & Conditional random field\\
	CSR & Compressed sparse row matrix format\\
	CV & Cross-validation\\
	EFSC & Essen Folksong Collection\\
	EMA & Exponential moving average\\
	FNN & Feed-forward neural network\\
	GEMM & Fast general matrix multiplication\\
	GP & Gaussian process\\
	IDyOM & Information Dynamics of Music\\
	L-BFGS & Limited-memory Broyden-Fletcher-Goldfarb-Shanno algorithm\\
	LTM & Long-term model\\
	MIDI & Musical Instrument Digital Interface\\
	MIR& Music information retrieval\\
	MVS & Multiple viewpoint systems\\
	NLP & Natural language processing\\
	NPMM & Neural probabilistic melody model\\
	OWL-QN & Orthant-wise limited-memory quasi-Newton optimization\\
	PPM & Prediction by partial matching\\
	PULSE & Periodical uncovering of local structure extensions\\ 
	PyPulse & Python framework for PULSE\\
	RBM & Restricted Boltzmann machine\\
	RNN & Recurrent neural networks \\
	RTDRBM & Recurrent temporal discriminative RBM\\
	SGD & Stochastic gradient descent\\
	STM & Short-term model\\
	TEF & Temporally extended features\\
	UML & Unified Modeling Language\\
\end{tabular}
\chapter{Introduction} \label{chap:Introduction}
In a world that is growing ever more interconnected, forging links between cultures becomes an increasingly meaningful endeavor. Of all cultural contrivances, music is one that has the power to go beyond the barriers that divide us and bring people together in a unique way. Thus, musical cognition and particularly the understanding of similarities and differences in styles is not only relevant for musicians, but for everyone.

\section{About Music Modeling}

While the study of music is typically pursued in the fields of arts and humanities as well as historic and systematic musicology, music has always exerted a certain pull on researchers of machine learning. Computational modeling of music and algorithmic composition have been well established since the 1990s and constitute two appealing realms for the application of learning methods \parencite{papadopoulos1999ai}. Music provides multifaceted data with several layers of temporal structure such as rhythm, melody and harmony \parencite{lerdahl1985generative,narmour1992analysis}. This sparked the search for a grammar as in natural language processing \parencite{steedman1984generative,rohrmeier2007generative}. 
Machine learning models adopt the techniques of humans who acquire their understanding of music by culturally influenced statistical learning  \parencite{rohrmeier2012implicit,saffran1999statistical,huron2006sweet}. This stands in contrast to prior computational rule-based approaches of finding a grammar or model of music \parencite{lerdahl1985generative,narmour1992analysis,schellenberg1997simplifying}. For music and statistical language modeling, \ngram models enjoy great popularity today. However, long ranging dependencies that may encompass the entire length of a piece, cannot be captured by traditional Markovian approaches \parencite{rohrmeier2011towards}. For example, the first and last note are often identical and motifs are repeated during the piece. Thus, it is indicated to apply non-Markovian methods, which were recently shown to outperform \art melody models \parencite{cherla2015discriminative}. Many statistical models of music have been learned for songs from different cultures and styles: European, Canadian and Chinese folk music \parencite{pearce2004}, Turkish folk music \parencite{sertan2011modeling}, northern Indian raags \parencite{srinivasamurthy2012multiple} and Greek folk music \parencite{conklin2011comparative}, to name just a few. This highlights the cross-cultural interest in music modeling of local as well as foreign pieces.

Music is considered to be a well suited domain for the study of human cognition. \textcite{pearce2012music} reason that music is a fundamental and ubiquitous human trait that has played a vital role in evolution, shaping culture and human interaction. According to Pearce and Rohrmeier, musical complexity and variety constitute a scientifically interesting cognitive system. Computational models of music have been used to analyze expectation behaviorally \parencite{pearce2006expectation,pearce2012auditory} and neuroscientifically \parencite{rohrmeier2012predictive}, as well as to study human memory \parencite{agres2017information}. By learning predictive models of music, this thesis is tightly linked to the study of expectation. In psychology, the expectations in future-directed information processing (\ie the expected likelihood of future events) are referred to as predictive uncertainty \parencite{hansen2014predictive}. 

Expectation in music was reported to evoke emotion \parencite{huron2006sweet, meyer1956emotion} and tension \parencite{lehne2013influence}. Being deceived in ones expectations of a melody's continuation may not have as far-reaching implications as a similar mistake in traffic. Nonetheless, such a deception was shown to make the listener's heart rate drop \parencite{huron2006sweet}, which underlines the significance of expectation in music. Similarly, unexpected harmonies were detectable by skin conductance measurements \parencite{steinbeis2006role}. In consequence, the right prediction of future events in music is essential for computational cognitive models as well as models of music generation. 

\begin{figure}[!ht]  
	\centering
	\includegraphics[width=\textwidth]{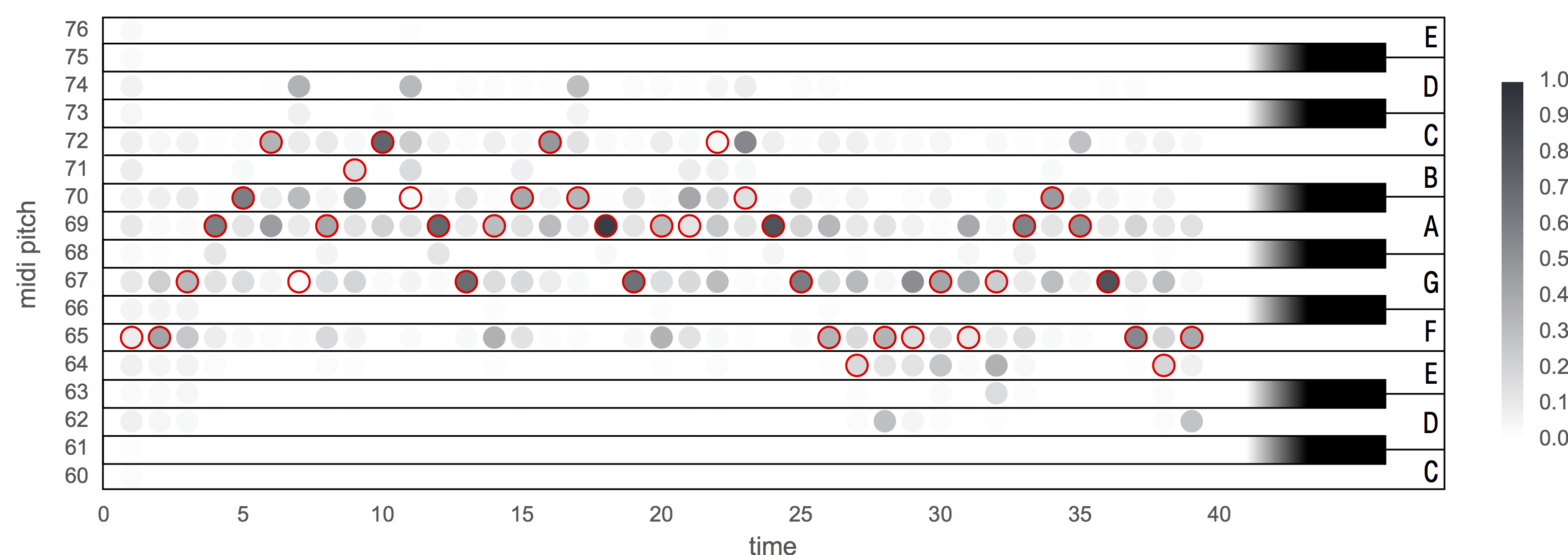}
	\caption{Piano keyboard visualization of the output of a predictive model for music. The predictive pitch distributions for J. S. Bach's chorale \textit{Erstanden ist der heil'ge Christ} (BWV 306) were generated by a model learned with \textit{P\textsmaller{ULSE}}. The red markers represent the actual pitch values. Each column represents the model's probability distribution for the next note, given all previous true notes.}
	\label{fig:bwv306_score}
\end{figure}

An event-based predictive model for symbolic music computes the conditional probability distribution over all possible future events given the past events. Formally, the conditional probability distribution $p(s_{t}|s_{0:t-1})$ is computed over all possible events $s_t \in \mathcal{X}$ at time $t$ in the song, given the context $s_{0:t-1}$ of events that have already occurred, where $\mathcal{X}$ is the sequence's alphabet or symbol space. Note that this problem is analogous to the prediction of the next letter or word of a text in statistical language models. \Fig \ref{fig:bwv306_score} visualizes the output of a predictive melody model for Bach chorale \textit{Erstanden ist der heil'ge Christ} (BWV 306). The probability distributions for the prediction of chromatic pitch events were generated by a model learned with \textit{P\textsmaller{ULSE}}. The dots in grayscale represent the probabilities for each pitch value at time $t$, given the past $t-1$ pitches that occurred in the data. The red markers visualize the actual pitches.

\section{Approach}
In this thesis, the recent \pulse method \parencite{lieck2016} is, for the first time, applied outside the realm of reinforcement learning to the task of sequential melody prediction. \pulse is an evolutionary algorithm that was shown to be successful in the domain of reinforcement learning and in the discovery of non-Markovian temporal causalities. In addition to learning a predictive model, \pulse also discovers and selects the best features for this model. To find a set of features, \pulse iteratively expands and culls the feature set until convergence, by using a feature generating operation and $L_1$-regularized optimization alternately. This allows \pulse to explore feature spaces that are too large to be explicitly listed in common feature selection approaches. At the same time, the feature generating operation and regularization factor allow the injection of top-down knowledge into the learner. The underlying conditional random field model affords an easy inspection and interpretation of the converged feature set.

I design a general Python framework for supervised learning with \pulse called \pypulse as well as a specialization of \pypulse for monophonic melody prediction. Subsequently, I explore the hyperparameter space of the method to find the best models and feature sets. The best models are compared to \art models, analyzed, and musicologically interpreted. The proposed framework operates on sequences of musical events represented by digitized musical scores in the MusicXML, \textit{**kern}, MIDI or abc format.

\section{Contribution}
This is the first application of the recently published \pulse feature discovery and learning algorithm to music. My focus lies on computational modeling of music cognition and musical styles (in contrast to algorithmic composition). I contribute to the \pulse method and machine learning community by (1) confirming \textit{P\textsmaller{ULSE}}'s capabilities through a successful application in a new domain, (2) designing and developing a \pulse Python framework, and (3) evaluating it in combination with $L_1$-regularized stochastic gradient descent. Further, I contribute to the field of computational modeling of music by (1)  introducing a new approach which outperforms the current \art algorithms significantly while (2) at the same time providing insights into the learned models, which I show to be music-theoretically interpretable. To the field of cognitive sciences I contribute by introducing a new computational surrogate model of human pitch expectation. Last but not least, I contribute to algorithmic composition by providing new \art models of musical styles which I show to be sufficient for the generation of new melodies of the respective styles.

\section{Thesis Structure}
Chapter~\ref{chap:background} is concerned with summarizing the line of research on music modeling that I will carry on, and with introducing \pulse and other algorithms that I will use subsequently. Chapters \ref{chap:pypulse} and \ref{chap:pulse-for-music} introduce and describe the new \pypulse framework and its application to music. The best \pypulse models for music are determined in chapter~\ref{chap:results}. Chapter~\ref{chap:evaluation} evaluates the learned models in depth by comparing them with prior work and analyzing the discovered feature sets.

Parts of this thesis were presented in \parencite{langhabel2017} but a large share of my work will remain exclusive to this thesis. This includes, but is not limited to, an in-depth discussion of the \pypulse framework (chapters \ref{chap:pypulse} and \ref{chap:pulse-for-music}), an examination of all free variables of the learner (chapter~\ref{chap:results}), the introduction of a larger number of musical features (\S \ref{sec:pypulse-tef} and \S \ref{sec:feature-combos}), a comparison of a model's predictions with psychological data (\S \ref{sec:psycho}), the musicological analysis of metrical-weight-based features (\S \ref{sec:feature-analysis-mk}), an analysis of the discovered feature sets' temporal extents (\S \ref{sec:temporal-analysis}), and sequence generation from the learned models (\S \ref{sec:pypulse-inference} and \S \ref{sec:results-inference}).
\chapter{Background and Related Work} \label{chap:background}

\section{Predictive Models for Music}
In this section, I review the most relevant literature for this thesis about music, specifically melody prediction. Firstly, I describe the concepts of long- and short-term models as well as multiple viewpoint systems to shed light on the underlying learning setting. Secondly, I introduce notable melody prediction models that are used to bring these concepts to life. The focus lies on the well-established \ngram models, as well as on recent better-performing connectionist approaches.

\subsection{Long-Term and Short-Term Models}
The terms long-term model (LTM) and short-term model (STM) are borrowed from the respective memory models in cognitive psychology. In the context of music prediction, they refer to the offline trained (the default setting when training any model or classifier) LTM and an online trained (during prediction time) STM that is discarded after every test song. Note however that \textcite{rohrmeier2012predictive} call the neuroscientific flavor to the STM's naming misleading. They remark that the model's concept neither matches the biological auditory sensory memory with an only second-long buffer, nor the working memory.

\begin{figure}[htbp]
	\centering
	\subfloat[][Beethoven's Ode to Joy]{\includegraphics[height=1.3cm]{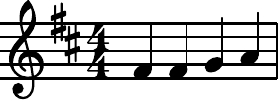}\label{fig:ode}}
	\qquad
	\subfloat[][German nursery rhyme (dataset 6, song 2)]{\includegraphics[height=1.3cm]{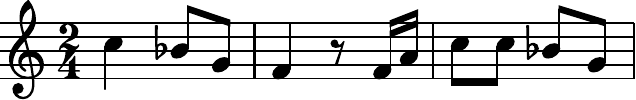}\label{fig:schlaf}}
	\caption{The beginning of two melodies. Well known (a) can be continued after sufficient exposure to classical music using our mind's LTM. Less known (b) can be carried on by memorizing motives in line with the STM concept without prior exposure.}
	\label{fig:ltm-stm-example}
\end{figure}

The concept to distinguish between an offline and online model for melody prediction was pioneered by \textcite{conklin1990prediction}. Since then, the ensemble of both models was shown to outperform pure LTMs \parencite{conklin1995, cherla_hybrid_2015, pearce2004, pearce2005, whorley2013construction}. For example, \textcite{teahan1996entropy} used the same idea to improve the compression performance of English texts by training a model on a corpus of similar texts first. In music, the LTM captures style-specific characteristics and motives while the STM captures piece-specific characteristics and motives. For each prediction of $s_t$, the STM is trained on the context $s_0, \dots, s_{t-1}$ within the current piece. \Fig \ref{fig:ltm-stm-example} invites the reader to intuitively compare the concept of LTM and STM based on music examples.

\textcite{pearce2004} also investigated a hybrid called LTM+ that is pretrained offline and continuously improved online with every new test datum that it encounters. Such models are not investigated in this thesis.

\subsection{Ensemble Methods}
Basically, LTM and STM are single models that can be combined to a mixture-of-experts. In multiple viewpoint systems however, LTM and STM may already be mixture-of-experts themselves. Such ensembles of classifiers typically boost the performance compared to each standalone classifier.

\subsubsection{Multiple Viewpoint Systems}\label{sec:MVS} 
Multiple viewpoint systems (MVS) are ensembles of music models that each have different points of view~--~\textit{viewpoints}~--~on the musical surface. They were first introduced by \textcite{conklin1990prediction,conklin1995} and have since then been applied to a range of tasks such as modeling of melody \parencite{conklin1995, pearce2005,whorley2013construction}, harmony \parencite{HedgesW16,whorley2013multiple, rohrmeier2012comparing, whorley2016music},
and classification \parencite{conklin2013multiple, hillewaere2009global}.

\begin{figure}[tb]
	\centering
	\includegraphics[width=.7\textwidth]{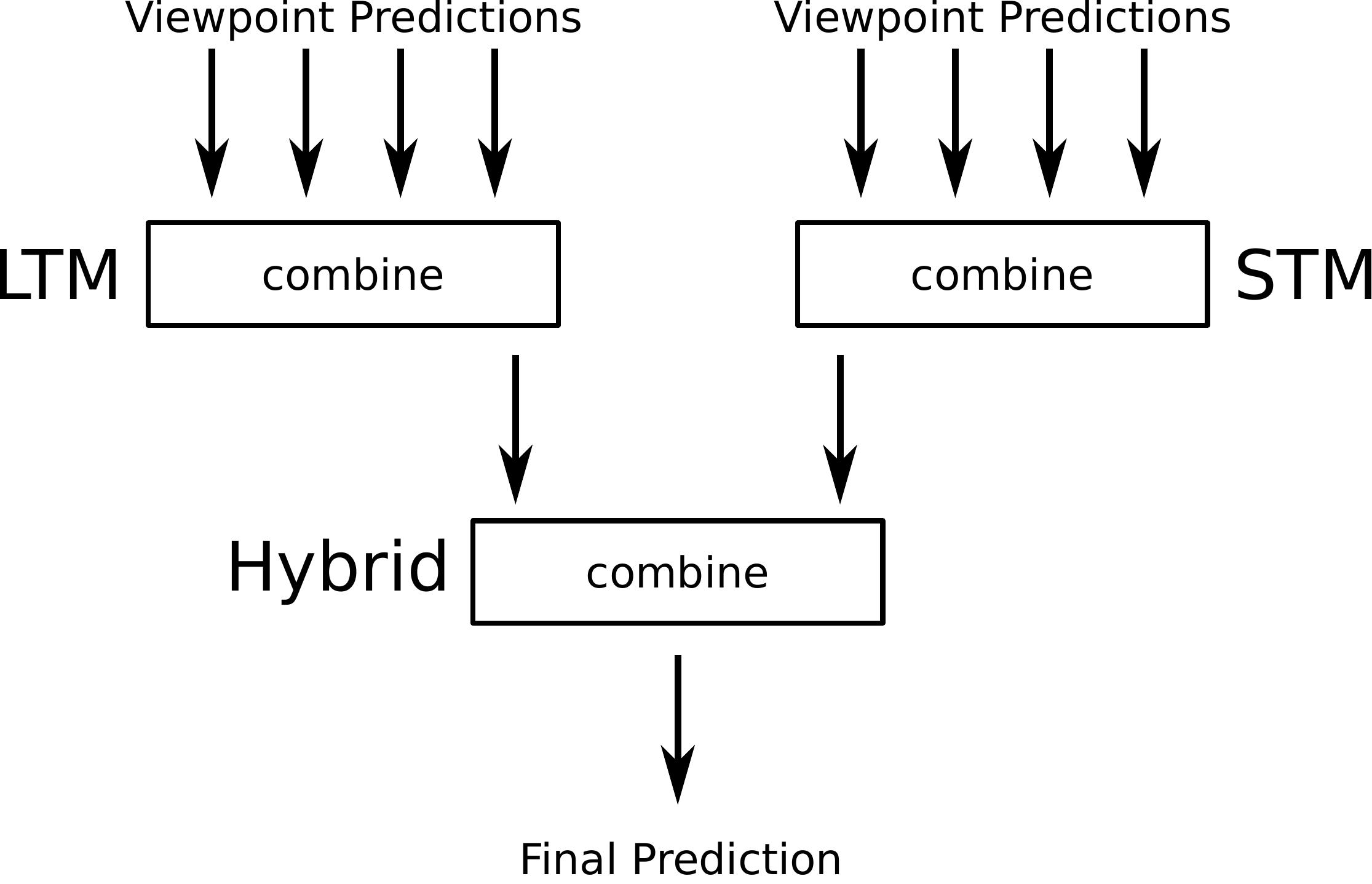}
	\caption{The multiple viewpoint system (MVS) architecture (reproduced based on \fig 3, \textcite{conklin1995}). Several predictions of long-term model (LTM) and short-term model (STM) viewpoints are combined separately in a first step, and merged into a final LTM+STM hybrid prediction in a second step.}
	\label{fig:mvs}
\end{figure}

\Fig \ref{fig:mvs} outlines the concept of MVS. The final hybrid model is a combination of the LTM and STM predictions, whereas the LTM and STM can be a combination of several viewpoint models themselves. The probability distributions are combined on a per prediction basis.

In an MVS, the sequence events for each viewpoint are of different \textit{types}. For a type $\tau$, the partial function $\Psi_\tau$ defines a mapping from the sequence events to $[\tau]$, the set of all values $\tau$ might take. A \textit{viewpoint} for $\tau$ consists of $\Psi_\tau$ and a model of sequence prediction over $[\tau]$. \textcite{conklin1995} define the following viewpoint categories: 
\begin{itemize}
	\item Basic viewpoints are taken directly from the data. The function $\Psi_\tau$ is total. Examples are `chromatic pitch' or `note duration' viewpoints.
	\item Derived viewpoints are inferred from one or several basic viewpoints. The function $\Psi_\tau$ may be partial. Examples for this viewpoint are the `sequential melodic interval' (\texttt{seqint}), the `interval from a referent' (\texttt{intfref}) or the `sequential difference in note onset' (\texttt{gis221}).
	\item Linked viewpoints operate on the Cartesian product of their constituents and introduce the capability to model correlations between viewpoints. The linked viewpoint of \texttt{intfref} and \texttt{seqint} is written as \texttt{intfref} $\otimes$ \texttt{seqint}.
	\item Test viewpoints map to $\{0,1\}$ and are used to mark locations in the sequence. An example of which is the `first event in bar' (\texttt{fib}) viewpoint.
	\item Threaded viewpoints are only defined on locations as described by a test viewpoint. Their alphabet is the Cartesian product of a test and another viewpoint. An example is the `\texttt{seqint} between first events in bars' (\texttt{thrbar}).
\end{itemize}
The challenge is to find the best performing set of viewpoints. Using the viewpoints \texttt{intfref} $\otimes$ \texttt{seqint}, \texttt{seqint} $\otimes$  \texttt{gis221}, \texttt{pitch} and \texttt{intfref} $\otimes$ \texttt{fib}, \textcite{conklin1995} reported their best result of 1.87 bits of cross-entropy (see section \ref{sec:evaluation-measure} for a description of this measure) on a dataset of 100 Bach chorales. However, this performance was computed on a single hold-out set and thus does not generalize well. \textcite{pearce2005} used the same set of viewpoints on a dataset of 185 chorales (see dataset 1 in section~\ref{sec:corpus}) using 10-fold cross-validation and reported 2.045 bits. In addition to that, Pearce employed a selection algorithm to find the best set of viewpoints and achieved a performance of 1.953 bits (smaller values are better). To train the MVS, the authors cited above used \ngram models of sequence prediction (see section~\ref{sec:ngram}).

\subsubsection{Combination Rules}\label{sec:combi-rules}
Ensembles of classifiers, mixture-of-experts and hybrid models all refer to the same concept: The predictions of several models are combined into one. Two combination rules have been applied to melody prediction models in the past, and will be considered here: \textcite{conklin1990prediction} was the first to apply a technique based on a weighted arithmetic mean (the \textit{sum rule}), which was then complemented by \textcite{pearce_methods_2004} with a geometrical mean version (the \textit{product rule}). Pearce, Conklin et al. showed that the product rule performs better than the sum rule for combinations of viewpoints within the LTM or STM. In combinations of the LTM and STM (the use case of combination rules in this work), both rules were shown to perform similarly, however, the sum rule was shown to perform better than the product rule if the combined LTMs and STMs operate on the same viewpoints. Outside the realm of music, \textcite{alexandre2001combining,kittler1998combining} who combine classifiers using above techniques found the sum rule to be more robust against erroneously high or low probability estimates than the product rule. They observed a better performance of the sum rule in the tested scenarios. \textcite{conklin1990prediction} also proposed weighting approaches for the source distributions that work on a per-distribution basis and proved to increase the performance.

For a set of models $M$ with model $m \in M$, let $p_m(s_{t}|s_{0:t-1})$ be the predictive distribution for $s_{t}$ as computed by $m$. The weighted arithmetic mean is then defined as
\begin{align}
p(s_{t}|s_{0:t-1}) &= \frac{\sum_{{m \in M}} w_m p_m(s_{t}|s_{0:t-1})}{\sum_{{m \in M}} w_m}~~,
\end{align}
and the geometrical mean with normalization constant $Z$ is defined as
\begin{align}
p(s_{t}|s_{0:t-1}) &= \frac{1}{Z} \Big[\prod_{{m \in M}} p_m(s_{t}|s_{0:t-1})^{w_m}\Big]^{\frac{1}{\sum_{{m \in M}} w_m}}.
\end{align}
The weighting strategy is shared by both methods and follows the idea that predictive distributions with lower entropy should entail a higher weight. Parameter $b\geq0$ tunes the bias attributed to the lower-entropy distribution. The divisor $\log |\mathcal{X}|$ normalizes the weights to be in $[0,1]$, with $\mathcal{X}$ being the sequence's alphabet. The weighting factor $w_m$ for model $m$ is then computed by
\begin{align}
w_m &= \bigg[\frac{-\sum_{s\in\mathcal{X}}\log p_m(s|s_{0:t-1})}{\log |\mathcal{X}|}\bigg]^{-b}~~.
\label{eq:bias}
\end{align}

\subsection[\textit{{n}}-gram Models]{\textit{\textbf{n}}-gram Models}\label{sec:ngram}
LTM, STM and MVS are general concepts that need a sequence model~--~such as \ngram models~--~to put them to life. \ngrams are especially popular as language models in machine translation, spell checking, and speech recognition. In the realm of music, they have first been used by \textcite{brooks1957experiment, hiller1959experimental, pinkerton1956information}, and since then gained great popularity, too \parencite{rohrmeier2012predictive}. \ngrams are used for almost all MVS implementations, and they are also used independently in, for example, \textcite{Rohrmeier2008statistical, Ogihara2008}. 

In a sequence, \ngrams are contiguous subsequences of length $n$. \ngram models use \ngrams for statistical learning by counting occurrences of each subsequence in the training data. They are $(n-1)$th-order Markov models, as the prediction of the next event $s_t$ depends on the last $n-1$ events only. Using maximum likelihood estimation the prediction is computed with
\begin{align}
p(s_t|s_{t-n:t-1}) = \frac{c(s_t|s_{t-n:t-1})}{\sum_{s \in \mathcal{X}}c(s|s_{t-n:t-1})}
\end{align}
where $c(s_t|s_{t-n:t-1})$ describes the occurrence count of the \ngram $s_{t-n:t}$.

The choice of the right $n$ is important as a too large $n$ leads to overfitting on the training data whereas a too small $n$ results in insufficient exploitation of the data structure. \textit{Unbounded n-grams} are variable-order Markov models that make the choice of $n$ superfluous: they try to compute the predictions $p$ based on the lowest-order context $s_{t-n:t-1}$ that matches the data and unambiguously implies $s_t$ to occur next in any training sequence $s_{t-n:t}$. If such a context does not exist, the highest order context that still matches the data is chosen. A disadvantage for \ngrams of long contexts is that they suffer from the curse of dimensionality: the number of model parameters grows exponentially with $n$.

\subsubsection{Smoothing and Escaping Methods}
In addition to overfitting for large $n$, the vanilla \ngram approach explained above has another flaw called the zero-frequency problem. If a subsequence did not occur in the training data but is encountered during prediction time, it is assigned likelihood zero. 

\textcite{pearce2004} make an in-depth empirical comparison of escaping and smoothing techniques that mitigate overfitting and the zero-frequency problem. They use the prediction by partial matching (PPM) algorithm \parencite{cleary1984data} which is prominent in data compression based on \ngram models. Cleary and Witten's original version implements \textit{backoff smoothing}: In case the context of length $n-1$ does not occur in the data, the algorithm backs off to the next shorter context length to compute the prediction. Another variant examined by Pearce and Wiggins is \textit{interpolated smoothing}, which always computes a weighted average of contexts of all length. Thus, it simultaneously reduces overfitting caused by an inappropriately chosen $n$ and the zero-frequency problem.

\textit{Escaping strategies} aim at solving the zero-frequency problem by assigning non-zero counts to newly encountered subsequences during prediction time. \textcite{pearce2004} tested several such strategies, amongst which, for example, the most basic one simply assigns count one to all unseen $n$-grams.

\textcite{pearce2004} reported unbounded \ngrams using their escaping strategy (C) and interpolated smoothing to be the best performing LTM configuration. They achieved 2.878 bits on a benchmarking corpus of monophonic chorale and folk melodies\footnote{In the remainder of this thesis I will refer to this corpus as the \textit{Pearce corpus} (also see section~\ref{sec:corpus}).}. The corresponding STM, LTM+, and hybrid of both achieved 3.147, 2.614 and 2.479 bits, respectively. \ngrams held the previous \art for STMs.

\subsubsection{IDyOM}
The Information Dynamics Of Music (IDyOM) framework\footnote{\url{https://code.soundsoftware.ac.uk/projects/idyom-project}} is a cognitive model for predictive modeling of music using MVS and unbounded \ngrams \parencite{pearce2005}. IDyOM extends the work of \textcite{pearce2004} with a larger range of viewpoints. It supports manual as well as automatic viewpoint selection. The system produces \art results for \ngram models on the task of melody prediction.

Since its introduction, the IDyOM framework was used several times in research of the cognitive sciences. For example, \textcite{pearce2012auditory,hansen2014predictive,pearce2010role} discuss IDyOM's suitability as a model for auditory expectation.

\subsection{Connectionist Approaches}
The comparison to connectionist approaches is of particular importance, as an approach based on recurrent neural networks (RNN) held the previous \art for non-ensemble methods in monophonic melody prediction. While there have been many connectionist approaches in the past \parencite{mozer1991connectionist, bosley2010learning,spiliopoulou2011comparing}, I will go into detail on only the most recent approaches that use the same benchmarking corpus and measure.

\subsubsection{RBM}
\textcite{cherla2013RBM} use a restricted Boltzmann machine (RBM) for the task of melody modeling. RBMs are a type of neural network in that, as a restriction compared to Boltzmann machines, the hidden units within one layer are not connected. Their approach outperforms \ngram LTMs on the Pearce corpus, especially for larger context sizes $n$. Furthermore, the RBM scales linearly with $n$ and $|\mathcal{X}|$ in contrast to an exponential scaling of $n$-grams. 

The authors also proposed a unified model that uses note durations additionally to pitches, as well as an arithmetic mixture model of a pitch and duration model. They report that the unified pitch and duration model performed worse than the pitch-only model, but that the ensemble performed better. The RBM pitch-only model was later reported to achieve 2.799 bits on the Pearce corpus \parencite{cherla2015discriminative}.

\subsubsection{FNN}
Feed-forward neural networks (FNN) were applied to single and multiple viewpoint melody prediction in prior work. \textcite{cherla2014multiple} examined two different architectures: (1) A FNN with a single sigmoidal hidden layer, a variable number of hidden units and input layer vectors of different length, and (2) an extension to the FNN (1) named neural probabilistic melody model (NPMM) which modifies the neural probabilistic language model and accepts several vectors as input. Each vector represents the fixed-length context of a musical viewpoint, for example, the past $n$ pitches. The viewpoints are one-hot encoded. In the NPMM several such binary input vectors are transformed to real-valued vectors of lower dimensionality within an additional embedding layer; the respective embeddings are learned from the data. The real-valued vectors then form the input to FNNs of architecture (1), all hidden units use hyperbolic-tangent activations. Thus, the prediction of an NPMM can be based on several viewpoints. The softmax output layer then returns the desired probability vector over the prediction classes\footnote{According to \textcite{gal2016dropout, gal2016uncertainty} the softmax output does not model the probability distribution over the prediction classes properly. They give examples that have high softmax outputs despite having a low model certainty. Gal and Ghahramani propose the Monte Carlo dropout method to compute the probabilities.}.

Cherla, Weyde and Garcez evaluated their models on the Pearce corpus and performed better than \ngram models but worse than the RBM on the single viewpoint task with 2.830 bits. In addition to that, they compared the performance of a single model with three input viewpoints and a mixture of single-input models of the same viewpoints on one dataset of the Pearce corpus. Both performed better than the single-viewpoint model, although the ensemble of several NPMM performed slightly better than the multiple-input NPMM.

\subsubsection{RTDRBM}
The recurrent temporal discriminative RBM (RTDRBM) was introduced by \textcite{cherla2015discriminative} as a non-Markovian approach for LTMs, and used in STM and LTM+STM hybrid settings by \textcite{cherla_hybrid_2015}. Cherla et al. combined the discriminative approach for RBMs \parencite{larochelle2008classification} with the structure of the recurrent temporal RBM \parencite{sutskever2009recurrent}, to achieve discriminative learning while capturing long-term dependencies in time series data: the conditional probabilities $p(s_{t}|s_{0:t-1})$ are learned directly while they explicitly depend solely on $s_{t-1}$.

The RTDRBM held the previous \art on LTM melody prediction for a single input type, with 2.712 bits on the Pearce corpus. It performs worse than \ngram models in the STM setting with 3.363 bits, but held the previous record in the combined LTM and STM (using \ngram STMs) with 2.421 bits.

In this section I have introduced the most relevant melody modeling literature for this thesis. While this work operates on symbolic data, much research has also been done on the prediction of raw audio data. For example, `A.I. Duett' of the \hbox{\url{https://magenta.tensorflow.org/}} project, \textcite{thickstun2016learning}, and \textcite{oord2016wavenet}, to name a few recent works.

\section{\texorpdfstring{\pulse}{PULSE}}
In the domain of reinforcement learning, delayed causalities pose special challenges to the learner. For example, a household robot leaving the fridge door open and discovering bad food the day after has to be able to conclude that this is not the cause of some directly preceding action, but of a delayed one. Therefore, such problems can only be solved with non-Markovian approaches. \textcite{lieck2016} introduced periodical uncovering of local structure extensions (\textit{P\textsmaller{ULSE}}), a feature discovery and learning method, that can find features for arbitrarily delayed or non-Markovian causal relationships. \pulse pulsatingly grows and shrinks the feature set by repeated generation of new features and selection of the fittest. It operates like an evolutionary algorithm with the exception that the fitness measure is not applied to each feature separately, but to the entire population at once. Features can be arbitrary functions that describe certain aspects of the data (see section~\ref{sec:crf}). \pulse was analyzed in both, a model-free and a model-based setting, and outperformed its competitors in the latter in a partially observable maze environment with delayed rewards.

Algorithm \ref{alg:pulse} describes \pulse in pseudo-code. In \textit{P\textsmaller{ULSE}}, a feature construction kit named \nplus is called repeatedly to incrementally build the feature set $\mathcal{F}$ (line \ref{pulse:grow}). This stands in contrast to typical feature selection methods where a large universal feature set is reduced to the features that are relevant for the data. Let $\theta = \{\theta_{f}\,|\,f\in\mathcal{F}\}$ be the respective set of feature weights. The optimization of an $L_1$-regularized objective $\mathcal{O}$ on feature set $\mathcal{F}$ and training data $D$ assigns non-zero weight $\theta_{f}$ to meaningful features $f\in\mathcal{F}$. \lone regularization is used to both select features and reduce overfitting. In line \ref{pulse:shrink}, using \textsc{shrink}, all features with zero weight are removed from the feature set. Note that features are always added with weight zero (line \ref{pulse:0weight}) so that the objective value remains the same before and after a call to \textsc{grow}. If greedily optimized, the objective value will monotonically decrease.
\begin{algorithm}[htbp]
	\caption{\pulse (reproduced, based on algorithm 1 \textcite{lieck2016})}\label{alg:pulse}
	\textbf{Input:} $N^+$, $\mathcal{O}$, $D$\\
	\textbf{Output:} $\mathcal{F}$, $\theta$
	\begin{algorithmic}[1]
		\State $\mathcal{F} \gets \emptyset$ , $\theta \gets \emptyset$
		\While {$\mathcal{F}$ not converged}
		\State \textsc{grow}($\mathcal{F}$, $\theta$, $N^+$) \label{pulse:grow}
		\State $\theta \gets$ argmin$_\theta$ $\mathcal{O}(\mathcal{F}, \theta, D)$
		\State \textsc{shrink}($\mathcal{F}$, $\theta$) \label{pulse:shrink}
		\EndWhile
		\State \Return $\mathcal{F}$, $\theta$\vspace{0.3cm}
		
		\Function{grow}{$\mathcal{F}$, $\theta$, $N^+$}
		\For {$f \in N^+(\mathcal{F})$}
		\If {$f \notin \mathcal{F}$}
		\State $\mathcal{F} \gets \mathcal{F} \cup \{f\}$
		\State $\theta_f \gets 0$ \label{pulse:0weight}
		\EndIf
		\EndFor
		\EndFunction
		
		\Function{shrink}{$\mathcal{F}$, $\theta$}
		\For {$f \in \mathcal{F}$}
		\If {$\theta_f = 0$}
		\State $\mathcal{F} \gets \mathcal{F} \setminus \{f\}$
		\EndIf
		\EndFor
		\EndFunction
	\end{algorithmic}
\end{algorithm}

For a time series prediction setting where dependencies reach back $n$ events, the authors proved that \pulse will converge to a globally optimal feature set within $n$ iterations, if: (1) \nplus uses conjunctions to expand every feature with all basis features, (2) the basis features are indicator features and describe all relevant past events, (3) the objective $\mathcal{O}$ is a strictly monotonic function of the model's goodness, and (4)  the optimization of $\mathcal{O}$ involves every feature $f \in \mathcal{F}$ and leads to a model that performs equally to an optimal predictor (see section 3.3 in \textcite{lieck2016} for details).

The `no free lunch' theorem states that there is no universal learning framework that performs well in all scenarios \parencite{wolpert_no_1995, wolpert97}. Prior knowledge about the domain of deployment has to be included into the framework to achieve a good performance. In \textit{P\textsmaller{ULSE}}, prior knowledge can be included in two ways: (1) By the definition of the \nplus operator and features, and (2) by the choice of the regularization term in the objective. The \nplus operator and objective as well as the underlying model are explained hereinafter.

\subsection{The Conditional Random Field Model}\label{sec:crf}
The model-based \pulse approach uses conditional random field (CRF) models. CRFs  \parencite{lafferty01} are a class of powerful discriminative classifiers for supervised learning. They were demonstrated to be successful in many applications \parencite{sutton2006introduction, peng2004chinese}, including music \parencite{durand2016downbeat,lavrenko2003polyphonic}.

In CRF, the data is described by feature functions $f$ which are arbitrary mappings $f: (x, y) \mapsto \mathbb{R}$, where $x \in X$ is the input data or context and $y \in Y$ is the class label or outcome.\\
A CRF computes the conditional probability $p(x|y)$ using the log-linear~model
\begin{align}
\begin{split}
p(x|y) &= \frac{1}{Z(y)} \exp \sum_{f\in\mathcal{F}} \theta_{f} f(x, y) \label{eq:CRF} \\
Z(y) &= \sum_{x^\prime\in X} \exp \sum_{f\in\mathcal{F}} \theta_{f} f(x^\prime, y)~~.
\end{split}
\end{align}
Factor $Z$ is called the partition function and normalizes the probabilities to sum up to unity. In log-linear models (also known as maximum entropy models), the linear combination of feature functions is computed in the logarithmic space which affords positive results even for negative feature values. The partition function can become a computational bottleneck, as it requires the summation over the whole space~$Y$.

\subsection{The Objective}\label{sec:objective}
As no closed form solution exists for \eq \ref{eq:CRF}, numerical optimization is used to find the best weight values $\theta$. Lieck and Toussaint used L-BFGS optimization for this task. Typically, maximum likelihood estimation, or equivalently minimization of the \textit{negative log-likelihood}, is used as optimization objective. Operating in logarithmic space has the advantage that floating point underflows (for very small likelihoods) are avoided and that the derivatives are computed over a sum instead of a product. For likelihood $\mathcal{L}$, with $\mathcal{L} = \prod_{{(x,y)\in D}} p(x|y)$, the objective $\ell(\theta;D)$ is computed as the sum of the negative log-likelihood and a convex regularization term $R(\theta)$ with regularization strength $\lambda$:
\begin{align}
	\ell(\theta;D) &= - \sum_{\mathclap{(x,y)\in D}} \log p(x|y) + \lambda R(\theta)
\end{align}
As both summands are convex, $\ell$ is convex as well, and any local optimum will be a global optimum. To facilitate feature selection, \pulse requires $R(\theta)$ to include terms that drive the weights of non-expressive features to zero. The authors relied on \lone regularization, whereas more sophisticated $R(\theta)$ could have been used to incorporate prior knowledge into the model (e.g. assuming a Gaussian prior over the weights with \ltwo regularization).

\subsection{The \texorpdfstring{\nplus}{N+} Operation}
\begin{figure}[htbp]
	\centering
	\includegraphics[width=.5\textwidth]{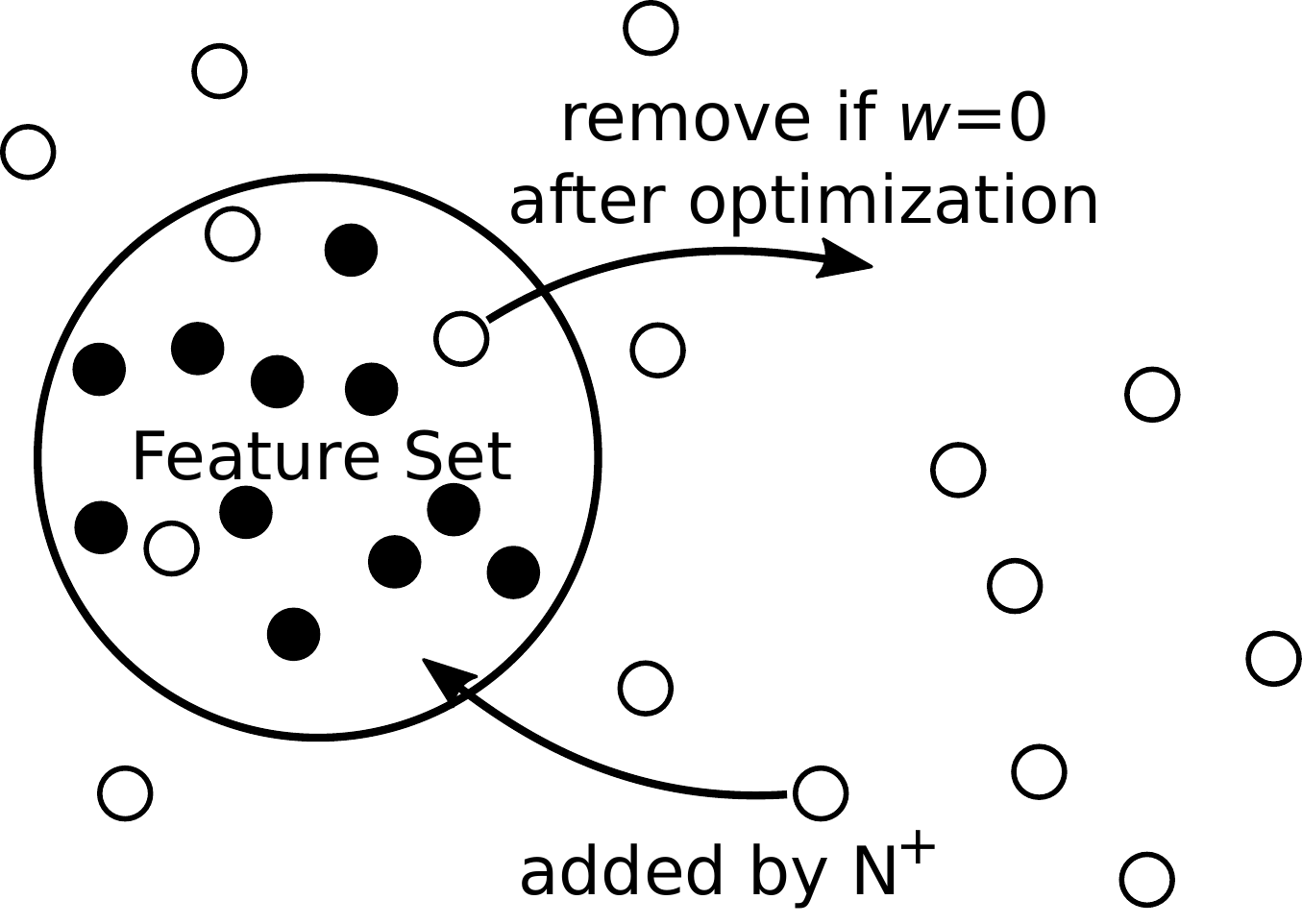}
	\caption{Interplay of the \textsc{grow} and \textsc{shrink} operators that expand (using N$^+$) and cull (using $L_1$-regularized optimization) the feature set. Full circles are features with non-zero, empty circles with zero weight.}
	\label{fig:featureset}
\end{figure}

The antagonist of the $L_1$-regularized optimization is the \nplus operation. The regularization compacts the feature set (\textsc{shrink}) and the \nplus operation expands it with new candidates (\textsc{grow}). The interplay of \textsc{shrink} and \textsc{grow} is depicted in \fig \ref{fig:featureset}. In the figure, circles describe features. The feature set is a subset of the whole  possible features space. Filled circles have non-zero weight, while empty circles have zero weight. Feature spaces can be too large to be searched exhaustively or even to be listed explicitly. The \nplus operation serves as a task-specific heuristic to generate new \textit{candidate features} and add them on probation to the feature set. \nplus bases its decisions on the \textit{active features} (those with a non-zero weight) in the shrunken feature set that have already proven beneficial.

The authors describe the generation of new candidates by creating liaisons between the active features and all elements from a set of basis features using conjunctions. Any other operation that synthesizes finite sets of features based on the active feature set is suitable, too. Such an operation may be the recombination of features using logical operands or the mutation of features by dropping out terms.

\section{Stochastic Gradient Descent}\label{sec:background-sgd}
Stochastic gradient descent (SGD) is a first-order online optimization algorithm. In batch gradient descent, the objective to be minimized (or maximized) is computed on the entire training dataset. In SGD, the gradient of the true objective is stochastically approximated using one datum (vanilla SGD) or a small subset of the training data (mini-batch SGD). Its main advantages show on very large datasets that are too big to fit in memory and cause expensive overheads in the computation or approximation of the Hessian matrix. In practice, optimizations on large datasets often converge faster using SGD compared to batch methods. The main disadvantage in using vanilla SGD is that finding a good learning rate and annealing schedule is hard. This shortcoming has been targeted by several automatic learning rate tuning methods such as AdaGrad and AdaDelta, which I use in this work.

\subsection{AdaGrad}\label{sec:adagrad}
\textcite{duchi_adaptive_2011} introduced AdaGrad, a method to automatically decay the learning rate individually per dimension of the weight vector. For each dimension and update, the \ltwo norm of all past gradients $g$ is accumulated. Subsequently, the initial learning rate $\eta$ is divided by these accumulators that increase monotonically. The underlying idea is that a history of larger gradients decreases the learning rate more than a history of smaller gradients. Thus, for dimensions with infrequently observed features or weaker gradients, the learning rate remains relatively higher.

Let $\epsilon$ be a small constant to prevent divisions by zero. For iteration $t+1$, the weight update for weight vector $w$ is
\begin{align}
	w_{t+1}&=w_{t}+\Delta w_{t}\\
	\Delta w_{t} &= - \frac{\eta}{\sqrt{\sum_{\tau=1}^{t}g_{\tau}^2 + \epsilon}} g_{t}~~.
\end{align}
The main problem inherent in this method is that the learning will stall in case of elongated training durations that cause the rate to become infinitesimally small.

\subsection{AdaDelta}\label{sec:adadelta}
AdaDelta was developed to solve the problem of diminishing gradients in AdaGrad. Furthermore, while SGD and AdaGrad require a meticulous tuning of the learning rate, AdaDelta was shown to perform similarly well without the need to tune any learning rate parameter \parencite{zeiler2012adadelta}. The history of past gradients is represented by the exponential moving average (EMA) of the squared gradients $E[g^2]$, with decay rate $\rho$. This replaces the global accumulation and prevents infinitesimally small updates. The numerator normalizes the weight updates to the same scale as the previous updates, using EMAs as well. Again, let $\epsilon$ be a small constant, then
\begin{align}
	\Delta w_{t} = - \frac{\sqrt{E[\Delta w]_{t-1} + \epsilon}}{\sqrt{E[g^2]_t + \epsilon}} g_t\label{eq:adadelta}~~.
\end{align}

\subsection[\texorpdfstring{\textit{L}$_1$}{L1} Regularization in SGD Training]{\texorpdfstring{\textbf{\textit{L}}$_\mathbf{1}$}{L1} Regularization in SGD Training}\label{sec:background-sgdl1}
The effect of \lone regularization is best described by \textcite{tibshirani1996regression}'s explanation of the \textit{lasso}: ``It shrinks some coefficients and sets others to zero, and hence tries to retain the good features of both subset selection and ridge regression.'' Its feature selecting properties stem from the diamond-like shape of the \lone ball, whose corners lie on the coordinate axes where all but one coefficient is zero. The contour lines of the stochastic gradients are more likely to touch the corners than the sides of the diamond, and thus, many coefficients become zero. Sparse models are advantageous when feature values are expensive to acquire, to increase the prediction speed in practice, and to reduce memory usage. The regularizing properties are advantageous whenever training data is not ample and maximum likelihood learning causes overfitting.

\textcite{tsuruoka09} state that it is difficult to have \lone regularization in SGD for two reasons: (1) The \lone norm is discontinuous at the orthant boundaries and thus not differentiable everywhere, and (2) the stochastic gradients are very noisy which makes the local decision whether to globally set a weight to zero or not difficult. Following the approach to add the \lone term to the objective~--~as it is done in batch methods~--~is not sufficient; it is highly unlikely that the weight updates precisely sum up to zero after optimization and thus the resulting model would not be sparse.

The following approaches seek to produce sparse models with \lone in SGD: \textcite{xiao2010dual} maintains running averages of past gradients and solves smaller optimization problems in each iteration to circumvent selecting features based on local decisions. \textcite{carpenter2008lazy,langford2009sparse,duchi2009efficient,shalev2011stochastic} all follow a two-step local approach. They first compute the updated weight without considering the regularization term, and then apply the regularization penalties under the constraint that the weights are clipped whenever they cross zero.

\textcite{tsuruoka09} point out shortcomings in the weight clipping methods and propose their \textit{cumulative penalty} approach. They compared their method to OWL-QN BFGS \parencite{andrew2007scalable} using CRF models, and found it to be similar in accuracy but faster on all benchmarked NLP tasks. In their approach, the total penalty $u$ that could have been applied to any weight is accumulated globally, as well as on a per-dimension basis the penalties $q$ that actually were applied. The resulting \lone penalty term is based on the difference of the total and actual penalty accumulators, and applied after the regular weight update. As a consequence, the gradients are smoothened out and a regularization according to the unknown real gradients is simulated.

Let $N$ be the size of the training dataset, $\lambda_{1}$ be the regularization strength for $L_1$, and $\eta$ be the global learning rate. Let further $w^i$ represent one dimension of the weight vector and let $q^i$ be the respective actual penalty accumulator. Then, the weight update for optimization iteration $t+1$ is computed with
\begin{align}
w^i_{t+\frac{1}{2}}&=w^i_{t}+\Delta w^i_{t}\\
w^i_{t+1}&=\begin{cases} \text{max}(0,w^i_{t+\frac{1}{2}} - (u_t + q^i_{t-1})& w^i_{t+\frac{1}{2}} >0\\
\text{min}(0,w^i_{t+\frac{1}{2}} + (u_t - q^i_{t-1})) & w^i_{t+\frac{1}{2}} < 0\end{cases}
\end{align}
where the total penalty accumulator $u$ and the received penalty accumulator $q^i$ are defined as
\begin{align}
u_t &= \frac{\lambda_{1}}{N} \sum_{k=1}^{t} \eta_k\label{eq:cumulative-u}\\
q^i_t &= \sum_{k=1}^{t} (w^i_{k+1} - w^i_{k+\frac{1}{2}})~~.
\end{align}
\chapter{\texorpdfstring{\textit{PyPulse}}{PyPulse}: A Python Framework for \texorpdfstring{\pulse}{PULSE}} \label{chap:pypulse}
In this chapter, I discuss the design and implementation of \textit{P\textsmaller{ULSE}}, a Python framework for \textit{P\textsmaller{ULSE}}. The framework was designed with generality in mind and supports feature discovery and learning for any kind of labeled data. The implementation boasts SGD in combination with \lone regularization to gear for large datasets, and uses Cython modules to maximize speed. Currently, the \pypulse framework is still under development and parts of it are solely realized for music-specific time series. The code will be published on \url{https://github.com/langhabel}.

\section{Design}\label{sec:pypulse-design}
The module design is specified as a UML class diagram \parencite{rumbaugh2004unified}.
The main modules are sketched in the following whereas the full diagram, including the specializations for music from chapter~\ref{chap:pulse-for-music}, is provided in appendix~\ref{appendix:uml}. Note that in the implementation for efficiency reasons the conceptual modules were melted together in several instances.

\subsection{Overview}
\begin{figure}[htbp]
	\centering
	\includegraphics[width=.8\textwidth]{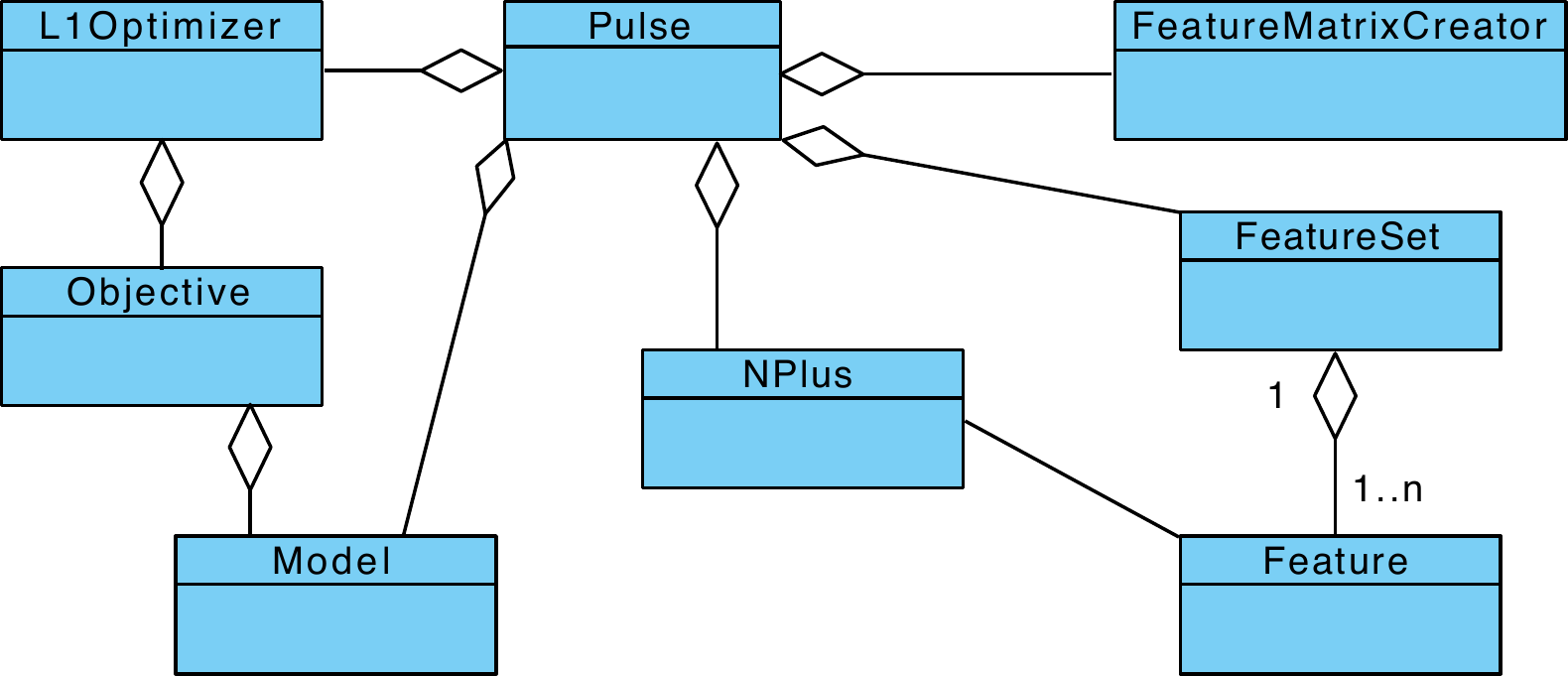}
	\caption{A UML sketch of the \pypulse framework's main modules.}
	\label{fig:uml-small}
\end{figure}

A minimalist version of the class diagram is presented in \fig \ref{fig:uml-small} and gives a structural overview of the architecture. The central \texttt{Pulse} module executes the algorithm while making use of the other modules.

\subsection{Module Descriptions}
In the following, the main modules and their chief functionalities are explained briefly:
\begin{itemize}
	\item \texttt{Pulse}: The main module has the public methods \texttt{fit()} and \texttt{predict()} for supervised training and prediction. \texttt{fit()} accepts a list of labeled data points, \texttt{predict()} accepts a data point and returns a label. \texttt{Pulse} is initialized to use a given \texttt{NPlus}, \texttt{L1Optimizer} and \texttt{Model}. During training, the learning algorithm alternatingly uses \texttt{NPlus} and \texttt{L1Optimizer} to discover the best feature set.
	\item \texttt{FeatureSet}: The \texttt{FeatureSet} is the container of the features and their respective weights. Its method \texttt{shrink()} removes all features with zero weight from the feature set.
	\item \texttt{NPlus}: The \texttt{NPlus} module has the method \texttt{grow()} which takes a \texttt{FeatureSet} object and returns an expanded instance of it. If previously empty, \texttt{FeatureSet} is initialized based on a given list of feature types. To be able to expand a feature, the implementation of \texttt{NPlus} has to have knowledge of the feature's structure.
	\item \texttt{Feature}: In CRFs, features are functions $f: (x, y) \mapsto \mathbb{R}$. In the implementation, a feature function takes a data point and label pair and returns a float value. The implementation has to maintain all relevant constants and states for the feature function's computation. 
	\item \texttt{FeatureMatrixCreator}: The job of the \texttt{FeatureMatrixCreator} is to prior to optimization compute the values of the feature function for each data point, feature in the feature set, and occurring class label. All values are stored in a three-dimensional feature matrix (see section~\ref{sec:fmat}).
	\item \texttt{Model}: This module provides the function \texttt{eval()} that, given the feature weights, the feature matrix, and for every data point a reference to the matching class label, evaluates the model.
	\item \texttt{L1Optimizer}: The function \texttt{optimize()} computes the best model weights using $L_1$-regularized optimization. As input \texttt{optimize()} takes a feature matrix, weight vector, regularization, and convergence parameters. Its actions are guided by the optimization \texttt{Objective}.
	\item \texttt{Objective}: This module declares the public functions \texttt{computeLoss()} and 
	\texttt{computeGradient()} that compute the loss value to be minimized during optimization and the gradient, respectively. Their computation requires the feature weights, the feature matrix, the number of training data points, and for every data point a reference to the matching class label.
\end{itemize}

\section{Implementation Details}\label{sec:implementation}
\begin{figure}[htbp]
	\centering
	\includegraphics[width=.5\textwidth]{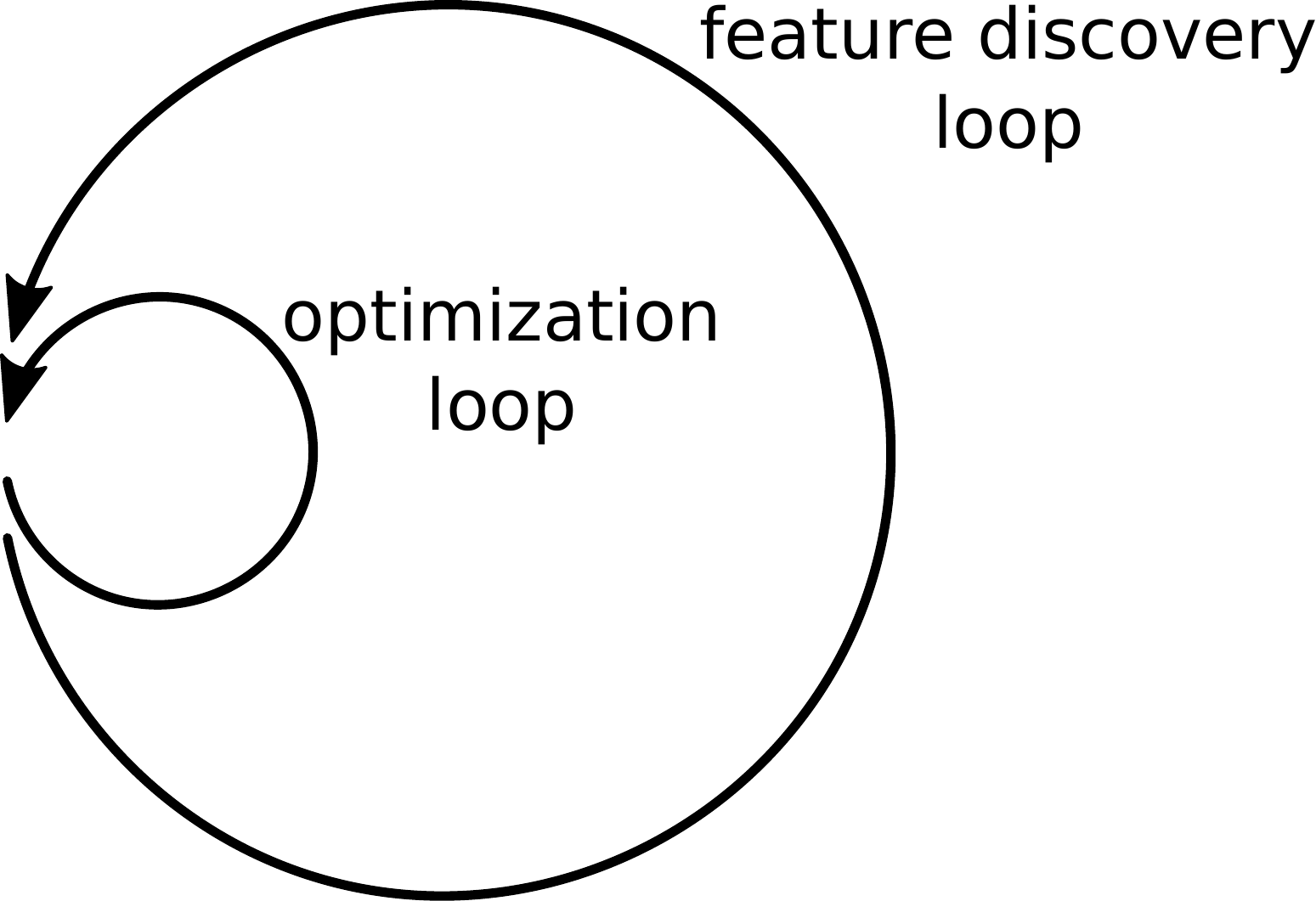}
	\caption{Interplay of \textit{P\textsmaller{ULSE}}'s main loops. The outer loop expands the feature set, the inner loop selects and learns the best features. For every outer loop iteration there are many inner loop iterations.} 
	\label{fig:outer_inner}
\end{figure}

The core of the algorithm is best depicted as two nested loops (see \fig \ref{fig:outer_inner}). For every \textit{outer loop} iteration for feature discovery there are several \textit{inner loop} iterations to select the best features using $L_1$-regularized optimization. In every iteration, the outer loop executes the sequence of function calls \texttt{grow()} -- \texttt{optimize()} -- \texttt{shrink()}.

\subsection[Implementing \texorpdfstring{$L_1$}{L1}-Regularized Optimization]{Implementing \texorpdfstring{\textbf{\textit{L}}$_\mathbf{1}$}{L1}-Regularized Optimization}\label{sec:imp-sgd}
It is pertinent to ask whether SGD or L-BFGS optimization is preferable to implement the \texttt{L1Optimizer} module. \textcite{bottou2010large,lavergne2010practical,vishwanathan2006accelerated} show that SGD with cumulative penalty \lone regularization (see section~\ref{sec:background-sgdl1}) is preferable over the Quasi-Newton L-BFGS method for large CRF models. Based upon these findings, I choose SGD over L-BFGS.

In prior attempts, I tested the $L_1$-regularized SGD optimizer AdaGrad-Dual Averaging \parencite{duchi_adaptive_2011}, as implemented in the TensorFlow machine learning framework \parencite{tensorflow2015whitepaper}. However, the observed convergence rates proved to be unsatisfying. Resorting to Theano \parencite{bergstra2010theano}, I implement $L_1$-regularized versions of the optimizers AdaGrad and AdaDelta by adapting them to \textcite{tsuruoka09}'s cumulative penalty approach (see section \ref{sec:pypulse-perfeature}), which leads to good results.

\subsection[Vectorizing Cumulative Penalty \texorpdfstring{\lone}{L1} Regularization]{Vectorizing Cumulative Penalty \texorpdfstring{\textbf{\textit{L}}$_\mathbf{1}$}{L1} Regularization}\label{sec:pypulse-perfeature}
In the following, I describe my adjustments to the cumulative penalty approach of \textcite{tsuruoka09} to make it work with adaptive stochastic gradient methods. The resulting optimizers learn $L_1$-regularized models with automatic per-feature learning rate annealing.

Regularization serves as a means to reduce overfitting and to promote better generalization performance. In the context of \textit{P\textsmaller{ULSE}}, it additionally provides a means of injecting prior knowledge into the model (see section~\ref{sec:pypulse-regularization} for more details). To inject prior knowledge, per-feature regularization is more precise than a global regularization rate that treats all features equal. This is especially the case when features describe varied properties or are of a heterogeneous expressiveness.

AdaGrad and AdaDelta maintain accumulator vectors to compute per-feature learning rates and weight updates. The cumulative penalty method maintains per-feature accumulators $q$ for the received weight penalties. Originally, it is designed to be used with a global learning rate and global \lone regularization factors only (see section~\ref{sec:background-sgdl1}). I extend the algorithm by vectorizing the total penalty accumulator~$u$. The resulting method offers per-feature learning rates and regularization factors while conserving the algorithm's essence. 

Let $u^i_t$ represent the total penalty accumulator value for dimension $w^i$ of the weight vector and iteration $t$. Let further $\lambda_{1}^i$ be the $L_1$ regularization strength for feature $i$, $N$ be the size of the training data, and $\eta^{i}=\Delta w^i / g^i$ be the respective adaptive per-feature learning rate of AdaGrad or AdaDelta. The vectorized version of equation~\ref{eq:cumulative-u} is then defined as
\begin{align}
u^i_t &= \frac{\lambda_{1}^i}{N} \sum_{k=1}^{t} \eta_k^i~~.
\end{align}

\subsection{Adding a Learning Rate to AdaDelta}
To accelerate the learning in AdaDelta, I introduce an initial learning rate parameter $\eta$ to \eq \ref{eq:adadelta}. Using the notation from section~\ref{sec:adadelta}, the extended weight update formula is
\begin{align}
	\Delta w_{t} = - \eta \frac{\sqrt{E[\Delta w]_{t-1} + \epsilon}}{\sqrt{E[g^2]_t + \epsilon}} g_t~~.
\end{align}

\subsection{Hot-Starting the Optimizer}\label{sec:hot-start}
\Fig \ref{fig:outer_inner} visualizes that for each feature discovery loop, a \textit{new} optimization is started, for a modified feature set. In every new outer loop iteration, features that graduated from the candidate to the active state are initialized with their previous non-zero weight, and the new candidates are initialized with weight zero. However, all AdaGrad/AdaDelta and cumulative penalty accumulators are reset by default. With the intent to accelerate learning, I add a \textit{hot-starting} option to the optimizers that carries over the accumulator values (total penalty accumulator, received penalties accumulator and squared gradient accumulator vectors) of the selected features to subsequent iterations.

\subsection{Convergence Criteria}
Convergence criteria aim at detecting an optimization algorithm's arrival at the optimum. The criteria that I implemented for the feature discovery and optimization loop, as shown in \fig \ref{fig:outer_inner}, are outlined in the following.

\subsubsection{Inner Loop Convergence Criteria}\label{sec:pypulse-conv-inner}
For the optimization loop, I implement two criteria. The first one is based on the convergence of the active feature set (i.e. the features with non-zero weight), and the second one is based on the convergence of the loss value (i.e. the value of the negative log-likelihood objective). The two criteria arise from different intents: The first one takes effect when the active feature set stops changing, and can be used in all but the last outer loop iteration where a convergence of the loss value is required. This motivates the second criterion that is meant to kick in later and meant to ensure a thorough training of the final feature set. Note that prior to the last outer loop iteration, we are only interested in the selected features, not their weights.

Let $i$ be the current training epoch, $\tau_{\textit{inner, loss}}$ and $\tau_{\textit{inner, active}}$ the decay rates of the exponential moving averages (EMA), $\ell_i$ the accumulated negative log-likelihood (see section~\ref{sec:objective}) over epoch $i$, and $\mathcal{F}^\textit{active}_{i}$ the active feature set at epoch $i$. Let factors $\gamma_{\textit{inner, loss}}$ and $\gamma_{\textit{inner, active}}$ be the convergence thresholds for the loss-based and active-feature-set-based criterion, respectively. The convergence criteria are defined as
\begin{align}
\textit{abs}\big(\textit{EMA}(\ell_i, \tau_{\textit{inner, loss}}) - \textit{EMA}(\ell_{i-1}, \tau_{\textit{inner, loss}})\big)&< \gamma_{\textit{inner, loss}}\label{eq:sgd-conv-loss}\\
\textit{EMA}(|(\mathcal{F}^\textit{active}_i \cup \mathcal{F}^\textit{active}_{i-1}) \setminus (\mathcal{F}^\textit{active}_i \cap \mathcal{F}^\textit{active}_{i-1})|, \tau_{\textit{inner, active}})&< \gamma_{\textit{inner, active}}~~.\label{eq:sgd-conv-active}
\end{align}

\subsubsection{Outer Loop Convergence Criteria}\label{sec:pypulse-conv}
For a well chosen \lone regularization factor, the \pulse feature discovery loop will arrive at an equilibrium between \textsc{shrink} and \textsc{grow}, and the feature set will converge. To detect such an equilibrium, three convergence criteria were implemented. With $j$ being the current outer loop iteration, $\mathcal{F}_j$ the feature set at iteration $j$, $\tau_{\textit{outer}}$ the decay rate of the EMA, and $\gamma_{\textit{outer}}$ the convergence tolerance, they are:
\begin{enumerate}
	\item[(a)] The relative convergence of the number of changing features in the feature set
	\begin{align}
	|(\mathcal{F}_j \cup \mathcal{F}_{j-1}) \setminus (\mathcal{F}_j \cap \mathcal{F}_{j-1})|< \gamma_{\textit{outer}} \cdot |\mathcal{F}_j|~~.
	\end{align}
	The symmetric difference of feature sets $\mathcal{F}_{j}$ and $\mathcal{F}_{j-1}$ between two consecutive iterations directly describes the fluctuation of features in the previous iteration. The count of changed features is compared to the threshold, which is relative to the current feature set count.
	\item[(b)] The relative convergence of the difference in feature set size 
	\begin{align}
	\textit{abs}(|\mathcal{F}_j| - |\mathcal{F}_{j-1}|) < \gamma_{\textit{outer}} \cdot |\mathcal{F}_j|	~~.
	\end{align}
	The absolute difference in feature set size between two consecutive iterations is a heuristic strategy for criterion (a). This criterion is simpler to compute than (a) but oblivious to the actual number of fluctuating features. Arrival at a constant feature set size is a necessary but not sufficient condition for feature set convergence.
	\item[(c)] The convergence of the EMA of the validation error $H$
	\begin{align}
	\textit{abs}\big(\textit{EMA}(H_{j}, \tau_{\textit{outer}}) - \textit{EMA}(H_{j-1}, \tau_{\textit{outer}})\big)< \gamma_{\textit{outer}}~~.
	\end{align}
	To stop learning after convergence of the validation error is a standard machine learning approach. However, in \textit{P\textsmaller{ULSE}}, the validation error does not always decrease monotonically. To smoothen the values, I consider the EMA of the validation error instead. Once the absolute difference of consecutive values falls below the threshold, learning is stopped.
\end{enumerate}

\subsection{The Feature Matrix}\label{sec:fmat}
The three-dimensional feature matrix $F \in (D\times\mathcal{F} \times\mathcal{X})$ quickly becomes very large (recall $D$ to be the data, $\mathcal{F}$ the feature set and $\mathcal{X}$ the space of all class labels or prediction alphabet). For example, $|D|=10,000,~|\mathcal{F}|=10,000$ and $|\mathcal{X}|=20$ already leads to a size of 8 GB for a matrix with 32 bit float values. However, if indicator features $f(x,y)\in\{0,1\}$ are used, then $F$ will typically be very sparse. The sparsity permits the storage of $F$ as a compressed sparse row (CSR) matrix which in the observed cases reduces the memory consumption by more than three orders of magnitude.

\subsection{Computation of the CRF}
I use a CRF-based approach to implement the module \texttt{Model} as described in \ref{sec:crf}. This approach was already shown to be effective by \textcite{lieck2016}. The computation of the model's gradient in the objective, specifically the matrix multiplication $F[\text{data idx}] \cdot \theta$ for every data point, is the computational bottleneck of the optimization. I decided to use single-threaded sparse matrix multiplication, as provided by Theano. Below, the investigations leading to this implementation options are described.

Space limitations enforce $F$ to be stored as a sparse matrix; nonetheless, single slices $F[\text{data idx}]$ can still be reverted to the dense representation for the computation of the model or objective. That raises the question whether a dense or sparse matrix multiplication runs faster. One inhibiting factor regarding the sparse alternative is that Theano (version 0.9.0b1) does not implement parallel sparse matrix operations on neither GPU nor CPU. Still, tests showed that single threaded sparse matrix multiplication performs faster than a parallelized dense multiplication. Furthermore, runtime profiling reported the sparse multiplication to take only 20\% (Python dot product) of the total computation time compared to 80\% (GEMM) for the parallel version (without considering the time needed to make the matrix dense). Using a GPU to run dense multiplication failed for two reasons: (a) If a slice $F[\text{data idx}]$ is copied to the GPU memory for every mini-batch, the Host-to-GPU transfer time outweighs any benefits, and (b) if instead the whole dense feature matrix is copied to the GPU once, the size of the feature matrix is restricted by the size of the GPU memory. An approach that was not tested is to compute $F[\text{data idx}] \cdot \theta$ by looping over the feature set while only updating weights for features used in the current datum.

\subsection{\texorpdfstring{\nplus}{N+} Postprocessing}
During expansion, the \nplus operator can introduce a significant number of irrelevant features. Such features slow down the optimization without providing any benefits. Thus, a postprocessing of the feature set after expansion is generally desirable. Obviously irrelevant are features with value zero for all data points and class labels, as they don't change the value of typical models (e.g. linear/log-linear models). These features are removed from the feature set after expansion and before optimization. In practice, this allows to implement and use \nplus operations that would have otherwise introduced too many irrelevant features and would have rendered optimization interminable.

\chapter{\pypulse for Music} \label{chap:pulse-for-music}
In this chapter, I describe two specializations of \textit{PyPulse}: Its adaptation to time series data and to music. The latter includes the conception of music-specific features, \nplus operators and regularization functions. The result is the highly versatile and potent \pypulse \textit{for music} framework for the prediction of musical attributes.

In line with prior research, this work uses event-based, in contrast to quantized time-based, time series data. Each event is described as a multidimensional vector of musical attributes, most importantly MIDI pitches on a chromatic scale. Indicator functions, as they are frequently used in NLP, are employed as features. Using the \nplus operator, such features can be expanded with logical operations. This work looks at different expansions with logical conjunctions. As a result, each feature can be a conjunction of other features itself. Feature selection is facilitated with a range of per-feature regularization factors, which depend on each feature's temporal extent. For the choice of the \nplus operator and regularization factors, the LTM and STM are considered separately. 

\section{Time Series Data}
This work uses event-based sequences of symbolic music data. In contrast, \pypulse is a supervised learning framework that learns data-label pairs $(x,y)$ in its CRF model. Though, the representation of a time series prediction or tagging task as a supervised learning problem is straightforward: The data points $x$ are the musical contexts $s_{0:t-1}$, for all possible time indices $t$. Based upon these contexts, the labels $y$ that represent the next time series event $s_{t}$ are to be predicted. Alternatively, any other label such as a sequence of tags could be predicted. As a typical piece of music contains more than one musical facet and possibly several voices, the time series events are multidimensional vectors of musical attributes.

Besides the \texttt{L1Optimizer} module, the \texttt{FeatureMatrixCreator} poses a computational bottleneck for the learner. Thus, the \texttt{FeatureMatrixCreator} module was optimized for time series data, parallelized, and implemented in Cython \parencite{behnel2011cython}.

\section{Temporally Extended Features}\label{sec:pypulse-tef}
What should the musical features look like? As mentioned previously, the \nplus operator requires knowledge about the features' structure to be able to read and manipulate them. I build on the concept of temporally extended features (TEF), proposed and realized by \textcite{lieck2016} in the context of reinforcement learning. A TEF for time series is a function $f: \mathcal{X}^\ast \times \mathcal{X} \mapsto \{0,1\}$, where $\mathcal{X}$ is the possibly multidimensional alphabet of the series (e.g. homophonic or polyphonic melodies) and $\mathcal{X}^\ast$ is the respective set of all possible sequences. \pypulse \textit{for music} uses two kinds of TEF: \textit{Compound features} and \textit{basis features}. Each basis feature $f_{\sigma,\nu}$ has the properties time $\sigma\in \mathbb{N}$ and value $\nu \in \mathcal{X}$ and is computed by
\begin{align}
f_{\sigma,\nu}(s_{0:t-1},s_{t})&= \mathbb{I}(v,s_{t-\sigma})\label{eq:tef}
\end{align} 
where the indicator function $\mathbb{I}(\cdot,\cdot)$ returns one if both arguments are equal and zero otherwise. Basis feature $f_{\sigma,\nu}$ thus considers the event that lies $\sigma$ steps in the past. Time $\sigma=0$ looks at the current event, which is to be predicted.
Compound features are conjunctions of features $f^i$ from a set $\varGamma$ of arbitrary basis features
\begin{align}
f(s_{0:t-1},s_{t}) = \bigwedge_{ f^i\in \varGamma} f^i_{\sigma, \nu}(s_{0:t-1},s_{t})~~.
\end{align}
Basis features do not exist on their own but only as constituents of compound features. Note that each compound feature is required to contain a basis feature $f_{0,\nu}$, as only features with $\sigma=0$ make a statement about the event $s_t$ to be predicted. A feature that operates entirely in the past has no predictive esteem. One that operates solely in the future ($\sigma=0$) models the occurrence frequencies of value $\nu$.

Three extensions of basis features are formalized in the following, having sequences of musical events in mind.
 
\subsection{Viewpoint Features}\label{sec:pypulse-viewpoint}
\textit{Viewpoint features} increase the expressiveness of TEF basis features by operating on different views on the data. They alter the definition of $f_{\sigma,\nu}$ by ahead of evaluation applying a mapping $\Psi$ from the input sequence to the viewpoint value range $\mathcal{V}$, where value $\nu \in \mathcal{V}$ (compare section~\ref{sec:MVS}). While the definition of $f_{\sigma,\nu}$ in \eq \ref{eq:tef} requires $s_{0:t-1}$ and $s_{t}$ to be of same dimensionality, this requirement is relaxed in viewpoint features: Let $\mathbb{U}\supseteq\mathcal{X}$ be the by viewpoints extended time series alphabet and $\mathbb{U}^\ast$ be the universal set of all sequences in the viewpoint domain. Let $f: \mathbb{U}^\ast \times \mathcal{X} \mapsto  \{0,1\}$ be the updated feature function and $\Psi:\mathbb{U}^\ast \times \mathcal{X} \mapsto \mathcal{V}$ be the viewpoint mapping, then
\begin{align}
f_{\sigma,\nu}(s_{0:t-1},s_{t})&= \mathbb{I}\big(v,\Psi(s_{0:t-1},s_{t})_{t-\sigma}\big)~~.
\end{align} 
Viewpoint features can be derived from one or several viewpoints themselves, as $\Psi$ can be chosen arbitrarily within its input/output value ranges. A feature may also choose to operate independently from either (or even both) of its properties $\sigma$ and $\nu$.

In \parencite{langhabel2017} we used the term \textit{generalized n-gram features} to describe compound features of one or several viewpoints. They encompass a superset of \textit{n-gram features}, that describe all temporally contiguous sequences of basis features, by additionally including all sequences that skip one or more time step. Generalized \ngrams can best be comprehended by imagining \ngrams that may have holes. Hence, they can depend on distinguished events in the past. While the space of all generalized \ngrams has size $(\mathcal{|X|}+1)^n$, in \textit{P\textsmaller{ULSE}}, typically only a tiny fraction of features has to be considered explicitly.

An overview of all implemented viewpoint feature types is given in the upper part of \tab \ref{tab:features}. They are described in the following.

\subsubsection{Pitch (P)}
\textit{Pitch features} describe the chromatic pitches of note events. They afford learning of the pitches' occurrence frequencies as well as transposition-sensitive motifs. The feature values equal the MIDI pitch values at the respective times. The value range $\mathcal{P}\subset \mathbb{N}$ encompasses all occurring pitch values. For a generic application, the equivalent of pitch features is a direct learning of the time series events, meaning the mapping $\Psi$ for P is the identity.

\subsubsection{Interval (I)}
The distance in semitones between two pitch values at times $t$ and $t^\prime \neq t$ is defined as the melodic interval. In \textit{interval features}, the interval is defined to be sequential. That means the source pitch values stem from subsequent time indices $t$ and $t^\prime = t+1$. Due to a lack of a reference pitch at time zero, they are undefined for $s_0$. Interval features are the tool of choice to describe transposition-invariant motifs. They take values of $\mathcal{I}\subset \mathbb{Z}$, where $\mathcal{I}$ is the set of all occurring intervals.

\subsubsection{Octave Invariant Interval (O)}
\textit{Octave invariant intervals} are an octave-invariant subcategory of interval features. They are computed by taking the interval feature value modulo 12, which is the number of semitones in one octave. As they are unsigned, they are less suited to describe motives. Instead, they forge links with the harmonic (vertical) intervals that make up chords. The intent behind these features is to learn broken chords which frequently make up parts of melodies.

\subsubsection{Contour (C)}
\textit{Contour features} describe melodies as either rising, falling or static. They are geared to model melodic movements on a higher level of abstraction than interval features. Their values are computed by taking the sign of the melodic interval and lie in the range $\{-1, 0, 1\}$.

\subsubsection{Extended Contour (X)}
\textit{Extended contours} refine contour features by differentiating between large (more than five semitones) and small intervals. The distinction is motivated by \textcite{narmour1992analysis}, who writes that large intervals prompt a change of the registral direction whereas small intervals suggest its continuation.

\subsubsection{Metrical Weight (M)}
\begin{figure}[t]
	\centering
	\includegraphics[width=.4\linewidth]{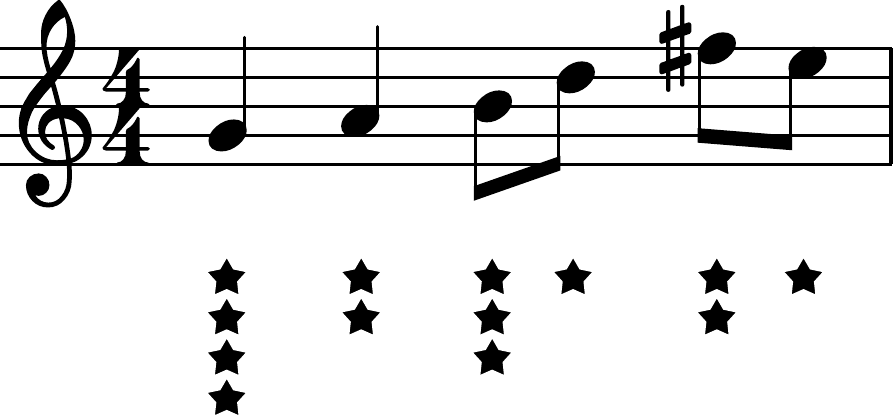}
	\caption{The metrical structure of a melody excerpt. The stars indicate the metrical weight; more stars correspond to a higher weight.}
	\label{fig:metrical-depth}
\end{figure}

The concept of the \textit{metrical weight} \parencite{lerdahl1985generative} is best understood by looking at the example in \fig \ref{fig:metrical-depth}. Several layers of incrementally finer grids are placed over the counts of each bar. The finest grid spacing is determined by the shortest note duration. In this example, the different grids are of one, two, four and eight grid points. For each grid, a note's weight is incremented if it lies on one of the grid points. The value range $\mathcal{M} \subset \mathbb{N}$, here $\mathcal{M}=\{1, 2,3, 4\}$, is determined by the depth of the metrical structure.

The metrical weight poses a special case among the presented features as it depends on several input dimensions, namely: The offset of the first bar and the note durations. It is worth noting that the metrical weight is defined for $\sigma=0$, irrespective of the target alphabet $\mathcal{X}$, as it is derived solely from the context.

\subsubsection{Negated Viewpoints (N$_\textbf{\textit{P}}$, N$_\textbf{\textit{I}}$, N$_\textbf{\textit{C}}$)}
The \textit{negated viewpoints} N$_P$, N$_I$ and N$_C$ are copies of the respective pitch, interval and contour viewpoints, with the only difference that the output of the indicator function is negated. The motivation in negated features lies in the efficient representation of causalities such as: ``If the last interval was a fifth, then the current one is not a fifth''. Negated viewpoint features break the sparsity of the feature matrix, as they are typically true almost everywhere. Because of immense memory requirements they were not evaluated.

\subsection{Anchored Features}\label{sec:anchored-features}
\textit{Anchored features} are a subclass of viewpoint features. The difference is of a semantical kind: Anchored features compute a relationship between two viewpoints $\Psi_\textit{anchor}$ and $\Psi_\textit{target}$. The viewpoint $\Psi_\textit{anchor}$ is the anchor or referent, in relation to which the viewpoint $\Psi_\textit{target}$ is considered. The anchor function $A(\Psi_\textit{anchor},\Psi_\textit{target})$ computes a kind of relation or distance measure. I use anchored features such that $\Psi_\textit{anchor}$ and $\Psi_\textit{target}$ map to pitch values, and the distance measure $A$ computes the interval between them. The benefit in computing such relative viewpoints is that it enables the learning of regularities on different scopes, for example for pieces, phrases or bars. Let value $\nu \in \mathcal{V}_A$, then
\begin{align}
f_{\sigma,\nu}(s_{0:t-1},s_{t})&= \mathbb{I}\big(v,A(\Psi_\textit{anchor},\Psi_\textit{target})_{t-\sigma}\big)~~.
\end{align}
The implemented anchored features are shown in the middle part of \tab \ref{tab:features} and described in the following. 

\subsubsection{Key (K)}
\textit{Key features} are octave invariant intervals between the current pitch and the tonic per mode. I use the Krumhansl-Schmuckler key finding algorithm \parencite{krumhansl_cognitive_1990} with key profiles from \parencite{temperley_whats_1999} for the computation of the key and tonic. Key profiles are weight vectors of dimension 12 (as many as there are choices for the tonic) for both major and minor scales. The Krumhansl-Schmuckler algorithm computes correlations between (optionally duration-weighted) note frequency counts and the key profiles. For that, the key profiles are transposed to 12 different tonics. The highest correlated key profile and its transposition determine the tonic and mode.

Having features anchored to the key has the advantage of learning regularities that are individual for each song depending on its key. Here, the regularities are motifs and pitch frequencies relative to the computed key. For each mode, the value range are the 12 degrees of the chromatic scale $\{\text{Major, minor}\} \times \{0, \dots, 11\}$.

\subsubsection{Tonic (T)}
\textit{Tonic features} are computed like key features, but ignore the mode and instead only use the tonic as anchor. Their value range is $\{0, \dots, 11\}$. This has the advantage that the learned regularities can be generalized over both major and minor keys. Additionally, ignoring the mode might abstract away from certain mistakes in the output of the Krumhansl-Schmuckler algorithm. Frequent confusions of the algorithm are the identification of the relative mode, subdominant or dominant as tonic.

\subsubsection{First-in-Piece (F)}
Frequently, the first or one of the first pitches equals the tonic of the song. \textit{First-in-piece features} simply use these pitches as estimate for the tonic. F$_i$ for $i \in \mathbb{N_+}$ computes the interval between the current and the $i$-th tone in the piece. The value range is a subset of all occurring intervals $\mathcal{I}$ in the dataset. In contrast to key and tonic features, I decided to use direction-sensitive intervals here, hoping to achieve a surplus in accuracy. This was not possible for key and tonic features, as the respective reference tonics were octave invariant already. 

\subsection{Linked Features}\label{sec:linked-features}
Compound features that do not include a basis feature for the target viewpoint at time $\sigma=0$ will compute the same value for all outcomes $s_t$. As they do not contribute to the model, I call them \textit{non-predictive}. To become \textit{predictive}, such features have to occur in compounds that contain predictive basis features. 

According to above definition, metrical weight (M) features are non-predictive. However, M features can be transformed to adopt a predictive nature. For example, the \nplus operator could be designed to generate compounds of M features and predictive features. \pulse would select the best compounds after optimization. However, M features on their own, due to their non-predictivity, would not survive the very first round of feature selection. A \textit{linked feature} is a construct to utilize non-predictive features that bypasses the dependency on N$^+$, by initializing non-predictive features in predictive compounds. In their simplest form, linked features are length-two compounds of a non-predictive and a predictive feature. More complex compounds are possible, but will not be investigated in this thesis.

The bottom part of \tab \ref{tab:features} lists the implemented linked features. In all cases, the value range is the cross product of the source types.

\subsubsection{Metrical Weight with Pitch (M$_\textbf{\textit{P}}$)}
M$_P$ features are compounds of metrical weight and pitch features. As M$_P$ features model the pitch frequencies per metrical weight, they link pitch values indirectly with their position in the bar.

\subsubsection{Metrical Weight with Key/Tonic (M$_\textbf{\textit{K}}$/M$_\textbf{\textit{T}}$)}
Similarly to above, M$_K$ and M$_T$ model the key and tonic in relation to the metrical weights. These features afford the learning of a chromatic scale degree distribution dependent on the position in the bar. The tonic version provides a higher generalization; the key version a higher precision.

\begin{table}[tb]
	\centering
	\scalebox{0.76}{
		\begin{tabular}{ccllp{0.43\linewidth}}
			\toprule
			& Abbr. & Name & Value Range $\mathcal{V}$ & Description \\
			\otoprule
			\multicolumn{1}{c|}{\multirow{1}{*}{\rotatebox{90}{\parbox{34mm}{\centering \textsc{Viewpoint \\Features}}}}} & P & Pitch & $\mathcal{P} \in \mathbb{N}$ & chromatic MIDI pitch \\
			\multicolumn{1}{c|}{} & I & Interval & $\mathcal{I} \in \mathbb{Z}$ & sequential pitch interval\\
			\multicolumn{1}{c|}{} & O & Octave invariant int. & $\{0, \dots, 11\}$ & intervals modulo 12 semitones \\
			\multicolumn{1}{c|}{} & C & Contour & $\{-1, 0, 1\}$ & registral direction, sign of interval \\
			\multicolumn{1}{c|}{} & X & eXtended contour & $\{-2, -1, 0, 1, 2\}$ & as contour but $-2/2$ if interval $>$ 5 \\
			\multicolumn{1}{c|}{} & M & Metrical weight & $\mathcal{M} \in \mathbb{N}$ & weight within metrical structure\\
			\multicolumn{1}{c|}{} & N$_P$, N$_I$, N$_C$ & Negated Pitch, etc. &$\mathcal{X}, \mathcal{I},\{-1, 0, 1\} $  & negated versions of P, I and C \\
			\midrule
			\multicolumn{1}{c|}{\multirow{1}{*}{\rotatebox{90}{\parbox{19.5mm}{\centering \textsc{Anchored\\ Features}}}}} & T & Tonic & $\{0, \dots, 11\}$ & octave invariant intervals from tonic \\
			\multicolumn{1}{c|}{} & K & Key & $\{\text{M, m}\} \times \{0, \dots, 11\}$ & octave invariant intervals from key\\
			\multicolumn{1}{c|}{} & F$_1$ & First-in-piece & $\mathcal{I} \in \mathbb{Z}$ & intervals from first-in-piece \\
			\multicolumn{1}{c|}{} & F$_{1,2,3}$ & First-three-in-piece & $\mathcal{I} \in \mathbb{Z}$ & intervals from first-three-in-piece\\
			\midrule
			\multicolumn{1}{c|}{\multirow{1}{*}{\rotatebox{90}{\parbox{15mm}{\centering\textsc{Linked\\ Feat.}}}}}& M$_P$ & Metrical weight, Pitch & $\mathcal{M} \times \mathcal{V}_P$ & combined metrical weight and pitch\\
			\multicolumn{1}{c|}{} & M$_K$ & Metrical weight, Key& $\mathcal{M} \times \mathcal{V}_{K}$ & combined metrical weight and key\\
			\multicolumn{1}{c|}{} & M$_T$ & Metrical weight, Tonic & $\mathcal{M} \times \mathcal{V}_{T}$ & combined metrical weight and tonic \\
			\bottomrule
	\end{tabular}}
	\caption{Overview of the implemented feature types. $\mathcal{P}$ is the set of all occurring pitches, $\mathcal{I}$ the set of all occurring intervals	and $\mathcal{M}$ the set of all occurring metrical weights in the data.}
	\label{tab:features}
\end{table}

\section{\texorpdfstring{\nplus}{N+} Operations}\label{sec:pypulse-nplus}
The function of \nplus in \pulse is to grow the feature set in every outer loop iteration (see section~\ref{sec:implementation}). Given the past feature sets, \nplus has the pivotal role of guiding the search through the feature space. In many cases, its operation is distributed on several sub-\nplus operators that each are active only for specific feature types or combinations thereof. Three techniques for growing the temporal extent of compound features named \textit{forwards}, \textit{continuous} and \textit{backwards expansion} are introduced in this section. It is notable that the \nplus operator satisfies \textcite{pearce2012auditory}'s claim that good models for music should select their own viewpoints. In \textit{P\textsmaller{ULSE}}, the algorithm freely constructs the most suitable features from a predefined construction kit.

The responsibilities of \nplus include the initialization and expansion of the feature set. Let $B$ be an arbitrary basis feature type:
\begin{itemize}
	\item Initialization: The initialization determines which feature types are used. N$^+_B$ is shorthand notation for the operator that adds type $B$ features with $\sigma=0$ for all $\nu \in \mathcal{V}_B$ to the feature set.
	\item Expansion: The $^\ast$ operator is used to indicate expansion of the given types. The operator N$^+_{B^\ast}$ initializes the feature set equally to N$^+_{B}$, and additionally expands features of type $B$ with features $f_{\sigma,\nu}$ of the same type for all $\nu \in \mathcal{V}_B$ and $\sigma$ dependent on the chosen strategy (see sections~\ref{sec:ltm-expansion} and \ref{sec:stm-expansion}). The notation N$^+_{({B_1}{B_2})^\ast}$ denotes that $B_1$ and $B_2$ features are initialized, and that features which contain either type are expanded with $f_{\sigma,\nu}$ for all $\nu \in \mathcal{V}_{B_1}$ and all $\nu \in \mathcal{V}_{B_2}$. Such an expansion is also called \textit{intermingled expansion}.
\end{itemize}
In the remainder of this thesis, feature type names will be used to describe the corresponding \nplus operators. \Eg $B_1 B_2 B_3$ is shorthand notation for the application of the \nplus operators N$^+_{B_1}$,  N$^+_{B_2}$ and N$^+_{B_3}$.

A grammar for music would be the ideal heuristic to construct complex features while keeping the search space at a minimum. However, grammars can to date only be generated for single pieces \parencite{sidorov2014music} with the general solution for entire styles remaining an open research question \parencite{jackendoff2006capacity, rohrmeier2011towards, rohrmeier2007generative,rohrmeier2011towards,lerdahl1985generative}. The techniques introduced in this work expand features in \textit{all} possible directions in contrast to a grammar-restricted set of directions. The specific expansion methods are explained in the following, separately for the LTM and STM.

\subsection{Long-Term Model}\label{sec:ltm-expansion}
In the LTM, training works by repeatedly running iterations of the outer loop on a training dataset until the feature set converges. Hence, \nplus can learn from previous iterations by considering the current feature set, and even from the acceptance rates of its proposed candidate features.

\subsubsection{Backwards Expansion}
\textcite{lieck2016} suggested a feature expansion method that expands features backwards in time and constructs generalized \ngram features. Lieck and Toussaint aimed at describing delayed causalities in reinforcement learning. This intention still holds in the context of melody prediction, as each note may depend on an arbitrary selection of preceding notes.

I pick up on a variation of Lieck and Toussaint's approach called `gradual temporal extension' that expands features stepwise into the past, and for which convergence has been proven. Let $\mathcal{F}_B \subset \mathcal{F}$ denote the set of all compound features of viewpoint $B$ after optimization and shrinking. The expansion in iteration $i$ is then defined by
\begin{align}
N^+_{B^\ast}(\mathcal{F}; i)=\big\{g ~\big|~\exists f \in \mathcal{F}_B, \nu \in \mathcal{V}_B: g = f \wedge f_{i+1, \nu}\big\}~~.\label{eq:backwards}
\end{align}
A global time index $i$ is incremented every outer loop iteration, and determines the time of the newly added features $f_{\sigma,\nu}$ to be $\sigma=i+1$. Features $f_{i+1,\nu}$ are added to each compound $f$ for all allowed values $\nu$ in $\mathcal{V}_B$. That means there are $|\mathcal{V}_B|$ many new features per compound of type $B$. After $n$ expansions, if no features were removed, the feature set would encompass the set of all contiguous and generalized $n$-grams. Remember that time index $i$ describes the previous time steps, consequently all expansions according to \eq \ref{eq:backwards} are done back in time.

The \nplus operator is based on assumptions and knowledge about the structure of the underlying data, to effectively serve as a search heuristic through the feature space. These assumptions are that (1) expansions of short relevant features are likely to be relevant as well, and (2) that a once irrelevant feature will not regain relevance, irrespective of the expansion. Features are assumed to be relevant if they survive \lone regularization, and irrelevant otherwise. Translated to music, this means that \nplus aims at expanding relevant motifs or patterns of musical attributes, whereas it expects that non-relevant motifs or patterns will only get less probable, if expanded.

\subsection{Short-Term Model}\label{sec:stm-expansion}
The STM distinguishes from the LTM in that the training dataset grows over time and is generally much smaller. \pulse could be applied to the STM scenario without any alterations by fitting a new \pulse instance on every time index of the song. However, this would require a huge number of outer loop iterations per song and is rather inefficient, considering that only one datum is added to the training set per call. Thus, for the STM, the outer loop is reduced to a single iteration per fit and the \nplus expansion is modified to take only the newest datum into account. This trade comes at the cost of giving up on having holes in the rendered \ngram features, but enables an efficient incremental learning of the STM. On the implementation side, \pypulse is adapted to carry over the model's state between subsequent fittings of the learner. The for the STM specialized \nplus operators are described in the following.

\subsubsection{Continuous Expansion}
The essence of continuous expansion matches that of backwards expansion, with the exception that it does not require a global iteration counter. Instead, time index $i$ is computed from the maximum temporal extent of the current compound feature by incrementing it by one:
\begin{align}
N^+_{B^\ast}(\mathcal{F})=\big\{g ~\big|~\exists f \in \mathcal{F}_B, \nu \in \mathcal{V}_B, i = \textit{max-depth}(f): g = f \wedge f_{i+1, \nu}\big\}
\end{align}
While this method affords learning of features without requiring any time or counter input, it prevents the occurrence of holes: the features learned are contiguous $n$-grams. 

Note that for every new datum $s_t$ the features are only expanded by one time step. That means that the full set of \ngram features for datum $s_t$ will only be reached, if no features were to be removed, after $n-1$ more features were added at time $t+n-1$. This is wasteful information processing, considering that the data in the STM was rare in the first place. However, it is yet to be seen that motifs and patterns that have just occurred are more relevant for the prediction of the next note, than those that lie further back in a piece.

\subsubsection{Forwards Expansion}
Forwards expansion addresses the shortcomings of continuous expansion by generating the set of all $t$-grams at song index $t$, if no features were to be removed. This is achieved by adding features to the front of the context, rather than to the tail. For this, the time indices of all features in the compound are first shifted back by one time step (operator $f_{-1}$), and then new features are added for time $\sigma=0$ and all values $\nu \in \mathcal{V}_B$. Forwards expansion is formalized as
\begin{align}
N^+_{B^\ast}(\mathcal{F})=\big\{g ~\big|~\exists f \in \mathcal{F}_B, \nu \in \mathcal{V}_B: g =  f_{-1} \wedge f_{0, \nu}\big\}~~.
\end{align}
In forwards expansion, the algorithm is given full control over the decision whether the most recent motifs and patterns, or those that lie further back, are more relevant. Thus, it constitutes the principally more capable extension method for the STM.

\section{Regularization}\label{sec:pypulse-regularization}
Regularization provides~--~in addition to the choice of the features and \nplus operator~--~another means of injecting prior knowledge into the model. Equally important, regularization also provides a means to curb overfitting. This section discusses these aspects of regularization, with respect to their realization in \pypulse \textit{for music}.

To inject prior knowledge, we can on the one hand add new regularization terms to the objective, and on the other hand change the regularization strength or factor. These modifications influence which features are removed from the set by \lone regularization and affect the weights of the features that are kept. The \nplus expansions in subsequent outer loop iterations directly depend on the selected feature sets and thus on the regularization. In consequence, the final feature set is shaped through an interplay of the \nplus operator and regularization.

Regularization curbs overfitting by penalizing the feature weights with an extra term in the objective\footnote{Recall that in the case of $L_1$-regularized SGD optimization, the penalty term is not directly added to the objective (see section~\ref{sec:background-sgdl1}).}. The goal is to find a balance between an overly precise fit (overfit) and a too loose fit on the data. Different regularization terms show different characteristics. The \lone regularization term, which is mandatory in \pulse, penalizes weights linear to their size. To discourage larger weights disproportionately, I added an \ltwo regularization term (in Bayesian terms a Gaussian prior) to the objective.

Per-feature regularization allows to penalize certain features more than others, based on knowledge of the problem domain. Specifically, it can guide feature selection more precisely. Assuming that the next tone in a melody depends more on the direct predecessors than on far back tones, I introduce a range of regularization factors $\rho(f)$ that depend on the features' maximum temporal extent. At this stage of research, all feature types are treated equally.

The maximum temporal extent $\varDelta_f$ of a feature $f(x,y)$ is the maximum time that the feature is looking back in $x$. The value $\varDelta_f$ also constitutes an upper bound to the length of the feature. Let $\lambda_{1}$ and $\lambda_{2}$ be the global regularization factors for \lone and \ltwo regularization, respectively. The per-feature regularization factor $\lambda_f$ for \lone and \ltwo regularization is then computed with
\begin{align}
\lambda_{f,1/2}= \lambda_{1/2} \cdot \rho(f)~~.\label{eq:regularization}
\end{align}
Let $\alpha > 0$ be a free parameter. The following functions $\rho$ were implemented:
\begin{align}
	\text{constant:} ~~~~&\rho(f) = 1 \\
	\text{linear:} ~~~~&\rho(f) = \alpha \cdot \varDelta_f\\
	\text{linear without zero:} ~~~~&\rho(f) = \alpha \cdot \varDelta_f + 1\\
	\text{polynomial:}~~~~ &\rho(f) = (\varDelta_f)^\alpha\\
	\text{exponential:} ~~~~&\rho(f) = (\alpha)^{\varDelta_f}\\
	\text{exponential with zero:} ~~~~&\rho(f) = \begin{cases} (\alpha)^{\varDelta_f}  & \varDelta_f > 0 \\ 0 & \text{otherwise}	\end{cases}
\end{align}
Note that in this work the anchored features are defined to have $\varDelta_f=0$. As a consequence, they remain unregularized for the linear and polynomial function, and are regularized according to factor $\lambda_{1/2}$ for the other functions.

In the STM, the regularization situation changes a lot during the progression of a song. Most notably, the amount of training data in the beginning and end of a song differs considerably. Consequently, the STM requires a time dependent regularization that is adapted to the amount of data. This was implemented by temporally decaying the regularization factor $\lambda_{1/2}$. Assuming a stronger Gaussian prior in the beginning of the song, injects the prior knowledge that it is initially more likely that new melodic patterns are introduced than that old ones are repeated. The new dynamical factor $\lambda_{1/2}$, which serves as input for the per-feature function in \eq \ref{eq:regularization}, is computed as
\begin{align}
\lambda_{1/2} = \lambda_{1/2}^{\textit{init}} \cdot \text{exp}(-\frac{t}{\tau_{1/2}})~~,
\end{align}
 given time index $t$, initial global regularization factor $\lambda_{1/2}^{\textit{init}}$, and temporal decay parameter $\tau_{1/2}$.

\section{Inference}\label{sec:pypulse-inference}
One may use inference on models of music to employ them as a cognitive model, to generate music, or to evaluate their performance. This section focuses on the latter two usages, and specifically on proving that the learned models afford a fruitful basis for music synthesis. Additionally, I intend to round off this thesis of Computer Science by making the learned models audible. My motivation however is clearly distinct from the generation of music in the sense of algorithmic composition, which is surely out of scope for this thesis (for a review of composition methods see \textcite{papadopoulos1999ai}).

In \textit{PyPulse}, predictions are made one note at a time. The synthesis of entire sequences requires additional thought. In line with my goals, I am not interesting in sampling a musical extravaganza, but rather to use the simplest method for finding a minimal cross-entropy sequence $s_{0:m}, m \in \mathbb{N}$ for a given model. I consider two methods that are briefly outlined in the following.

\textit{Beam search} \parencite{manningNLP} maintains $k$ slots (beams) holding one sequence each. In every round, each of these sequences is extended with the $k$ likeliest subsequent events. All $k\times k$ extensions then compete for the $k$ slots in the next round. This allows the algorithm to recover from greedy choices that otherwise would have lead to a local optimum. Manning reports the method to frequently perform very well, although it is not guaranteed to find the global optimum.

\textit{Iterative random walk} \parencite{whorley2013multiple} is an extension of the mathematical random walk. Starting with an initial event or sequence $s_{0:t-1}$ at time $t$, firstly, the conditional distribution $p(s_{t}|s_{0:t-1})$ is computed. Secondly, the next event is sampled with probabilities according to the computed conditional distribution. The event is concatenated to the context and the whole process is repeated. In iterative random walk, random walks are run repeatedly until sufficiently good entropy results are achieved. To prevent low probability choices that thwart the result, \textcite{whorley2016music} introduced the constraint that a sample's likelihood has to exceed a certain threshold.

\chapter{Model Selection} \label{chap:results}
In the scope of this thesis, the \pypulse \textit{for music} framework was thoroughly fine-tuned for musical data by optimizing the free variables and selecting the best regularization functions as well as \nplus operations. The impact of each hyperparameter, that means meta-variable of the learning algorithm, is discussed and its optimal value is determined.

This chapter is structured as follows: Sections~\ref{sec:corpus} and \ref{sec:evaluation-measure} introduce the corpus and evaluation measure used for assessing the model's predictive capacities. In section~\ref{sec:gp}, I outline Gaussian processes which are used for the optimization of the regularization parameters. Cross-validation is explained in section~\ref{sec:cv}. It is used in the computation of model benchmarks to ensure comparability. Section~\ref{sec:results-optimization} is concerned with the tuning of all SGD related hyperparameters to facilitate a quick learning, and with the choosing convergence criteria for the SGD and feature discovery loops. Finally, sections~\ref{sec:per-feature-reg} and \ref{sec:feature-combos} compare the various per-feature regularization terms and \nplus operations of the \textit{PyPulse for music} framework.

\section{Methodology}\label{sec:methodology}
This section introduces the evaluation corpus and measure, cross-validation, and Gaussian process based hyperparameter optimization. The introduced methodologies are used for the experiments in the remainder of this chapter and in chapter~\ref{chap:evaluation}.

\subsection{Corpus}\label{sec:corpus}
\pypulse \textit{for music} operates on symbolic music representations in the 12-tone equal temperament system, most notably on sequences of chromatic pitches. A vast amount of digitized sheet music is available online, in a variety of formats. The Center for Computer Assisted Research in the Humanities at Stanford University hosts more than 100,000 \textit{**kern} files \parencite{huron1997humdrum}, and makes them freely available. Other notable music notation formats are: abc, for which more than 500,000 digital music sheets are available, the widely supported MusicXML format, and the MIDI format which poses the de facto standard in digital music representation.

\textcite{pearce2004} selected and established a benchmarking corpus of 1,194 multifaceted melodies from the \textit{**kern} repertoire. The \textit{Pearce corpus}\footnote{\url{http://webprojects.eecs.qmul.ac.uk/marcusp/}} comprises eight sets of melodies of different style and origin (see \tab \ref{tab:datasets}). These are: Canadian folk songs and ballads from Nova Scotia, soprano lines of chorales BWV 253-438 harmonized by J. S. Bach, and folk melodies of Helmut Schaffrath's  (EFSC). The EFSC datasets are Alsatian, Yugoslavian, Swiss and Austrian folk songs, German nursery rhymes and Chinese pieces from the province Shanxi.

I parse the data using the Python musicology toolkit \textit{music21} \parencite{cuthbert2010music21}, which supports the most popular symbolic music file formats MusicXML, MIDI, \textit{**kern} and abc. Respecting the prior work on the Pearce corpus, I modify the music notation by merging all ties into single notes and deleting all rests. The alphabet $\mathcal{X}$ is defined for each of the eight datasets to be the respective set of uniquely occurring pitch values.

\begin{table}[tb]
	\centering
	\begin{tabular*}{\linewidth}{c l  c  c  c}
		\toprule
		ID & Description  & Melodies & Mean events/melody & $|\mathcal{X}|$  \\
		\otoprule
		0 & Canadian folk songs/ballads  & 152  & 56.270  & 26  \\
		1 & Bach chorales (BWV 253-438)  & 185  & 49.876  & 21  \\
		2 & Alsatian folk songs (EFSC)  & 91  & 49.407  & 32  \\
		3 & Yugoslavian folk songs (EFSC)  & 119  & 22.613  & 25   \\
		4 & Swiss folk songs (EFSC)  & 93  & 49.312 & 34  \\
		5 &  Austrian folk songs (EFSC) & 104  & 51.019  & 35  \\
		6 &  German nursery rhymes (EFSC) & 213  & 39.404 & 27  \\
		7 & Chinese folk songs (EFSC)  & 237  & 46.650  & 41  \\
		\midrule
		Total & Events: 54308 & 1194 & 45.484 & 45 \\
		\bottomrule
	\end{tabular*}
	\caption{The benchmark datasets, as first introduced by \textcite{pearce2004}, which I refer to as \textit{Pearce corpus} in the following.}
	\label{tab:datasets}
\end{table}

\subsection{Evaluation Measures}\label{sec:evaluation-measure}
The performance of a computational model for music is measured by its outputs, which are the predictive distributions. Different kinds of measures have been used in the past:
\begin{itemize}
	\item Information theoretic \textit{cross-entropy} has been used as a quantitative measure in the majority of prior research; \eg \textcite{conklin1995,pearce2004,cherla2015discriminative}, just to name a few. In melody prediction, the cross-entropy $H_c(p, q)$ between the learned model $p$ and the data distribution $q$ cannot be computed directly, as the true data distributions is unknown. Thus, typically $H_c$ is approximated by a Monte Carlo estimate between $p$ and the test dataset $D_{\textit{test}}$ with
	\begin{align}
		H_c(p, D_{\textit{test}}) &= \frac{- \sum_{x,y \in D_{\textit{test}}}  \log_2 p(x|y)}{|D_{\textit{test}|}}~~.
	\end{align}
	This approximation is directly linked to the geometrical mean $GM$ with $H_c  =  -\log_2(GM)$. 	
	
	Note that cross-entropy is a natural choice in \pypulse \textit{for music} as it matches the employed negative log-likelihood objective $\ell(\theta;D_{\textit{test}})$ with no regularization (see section~\ref{sec:objective}):
	\begin{align}
		H_c(p,D_{\textit{test}})=\ell(\theta;D_{\textit{test}}) / |D_{\textit{test}}|
	\end{align}
	Shannon's coding theorems from 1948 motivate the interpretation of the models from a data compression perspective. There, entropy describes the lower bound for the number of bits needed to encode a symbol of the alphabet $\mathcal{X}$. Lower values thus stand for better compression. In the context of prediction, a lower number of bits means that the model is more likely to predict the true outcome. 
	\item Regarding the cognitive sciences, a good model is one that accurately simulates human expectations. \textcite{agres2017information,pearce2005,pearce2006expectation} compared the outputs of computational models with psychological data.
	\item Evaluation by synthesis of new original pieces is a measure mostly used in algorithmic composition. \textcite{conklin1995} noted that a better predictive model will generally be able to generate a better piece. However, \textcite{trivino2001using} remarked that it is hard to quantify the quality of generated pieces. Instead, they used auditory experiments as criteria of goodness. \textcite{whorley2016music} evaluate their compositions by counting violations of a set of rules, in addition to using cross-entropy.
	\item The classification accuracy is a simple machine learning measure for multiclass classifiers. For classifier $g$ and data label pairs $(x,y)\in D_{\textit{test}}$, it is computed as the unweighted empirical classification error $\frac{1}{|D_{\textit{test}|}} \sum_{x,y \in D_{\textit{test}}}\mathbb{I}(g(x),y)$. \textcite{conklin2013multiple} uses this measure to classify folk tune genres and regions.
	
\end{itemize}
I will use all of the above measures in this thesis, but focus on cross-entropy as a reliable benchmark for comparing my results internally and with \art models.

\subsection{Gaussian Process Based Optimization}\label{sec:gp}
It is hard to determine the optimal hyperparameter values manually. Thus, algorithms such as grid search, random search or Gaussian process (GP) based Bayesian hyperparameter optimization are employed. Especially when the search space is of high dimensionality and when samples are expensive to obtain, GP based optimization seems most appealing as it samples based on educated guesses. GP based optimization is a Bayesian technique in which a GP prior distribution is chosen to describe the unknown function under optimization \parencite{snoek2012practical}. A GP surrogate model is maintained for the unknown function, and updated with every newly obtained sample value in the course of optimization. Based on the surrogate model's uncertainty and mean, which are known in every point, exploration of the parameter space and exploitation of the surrogate are balanced to determine the optimal next sampling location.

The overhead of computing the surrogate model is in my case easily outweighed by the sampling costs. I utilize the Scikit-Optimize framework\footnote{\url{https://scikit-optimize.github.io}} for GP based optimization based on a Matern kernel and with using expected improvement as acquisition function.

I would like to conclude by discussing the advantages and disadvantages of grid search and GP based hyperparameter optimization. Grid search persuades by shorter runtimes for coarsely spaced grids, easily interpretable results when interpolated as curves over the grid, and few hyperparameters. However, it is almost certain that the optimum is never hit precisely, which introduces noise into the benchmarks. Prior experiments using grid search showed an impaired comparability, irrespective of the grid spacing (I tested a grid of 5, 7 and 11 samples). In contrast to that, GP based optimization can be expected to find the optimum precisely, but requires a multiple in samples (compared to the aforementioned grid sizes). As samples are expensive to acquire, the increased precision comes at the cost of a higher runtime. On the downside, GP based algorithms are more complex than grid search as they have several free hyperparameters themselves.

For finding the best \nplus configuration, comparability between performances for different configurations is vital. Thus, GP based optimization was used for precisely finding the regularization parameter optima in the LTM and STM.

\subsection{Cross-Validation}\label{sec:cv}
To evaluate the test performance of a model, one should always utilize left-out data that the model has not been trained on. Thus, the dataset is split into a training and test dataset. Training the model on one dataset and evaluating it on the other one is called a split-sample or hold-out approach. Hold-out evaluation has the considerable disadvantage that, if data is not abundant, performances calculated in this way do not generalize well. Consequently, the results will vary with the choice of the two sets and their validity will be limited. Smaller training sets induce a higher bias, and smaller test sets induce a higher variance of the model.

\textcite{dietterich1998approximate} introduced $k$-fold cross-validation (CV), a model selection method that makes better use of the data than hold-out validation. The data is split into $k$ equally sized folds. In each of $k$ iterations, one varying fold is used as test set whereas the remaining $k-1$ are used as training set. The resulting $k$ performances are averaged to obtain the final CV performance.

Along with their benchmarking corpus, \textcite{pearce2004} established the use of 10-fold CV\footnote{See \url{http://webprojects.eecs.qmul.ac.uk/marcusp/} for the fold indices of the Pearce corpus.}. Using $k=10$ folds is considered to be a good balance of the bias-variance trade-off. 

To facilitate comparability with other work on the Pearce corpus, I use 10-fold CV and identical fold indices wherever I compute corpus benchmarks. Per fold, I use GP based hyperparameter optimization on a held out validation set (a small subset of the training set), to find the best values for the regularization strength $\lambda_{1/2}$. For the majority of the remaining hyperparameters, I frequently fall back on hold-out validation with grid search instead of GP based optimization minimize the computing time.

\section{Reducing Computational Expenses} \label{sec:results-optimization}
The \pypulse \textit{for music} framework ships with a whole range of free hyperparameters. In this section, the SGD optimization's hyperparameters are tuned. This is important to ensure the learning success and further aims at accelerating the pace of learning.

In section~\ref{sec:optimization-hyperparameter}, firstly, the SGD optimizers AdaGrad and AdaDelta are tuned to achieve minimal objective values within a given time frame. Secondly, the two optimizers are compared against each other with respect to these values. The convergence parameters are set in section~\ref{sec:convergence}, to prevent computational resources being spent on insignificant improvements. Last but not least, the potential in hot-starting the optimization for reducing computational expenses is explored in section~\ref{sec:experiments-hotstart}. \Tab \ref{tab:tuning-overview} gives an overview of the best found hyperparameters.

We assume all hyperparameters tuned in this section to be music specific but corpus independent. Hence, the 185 Bach chorales (dataset 1) are used for all experiments.

\begin{table}[htbp] 
	\centering
	\begin{tabular*}{\linewidth}{l | l}
		\toprule
		Realm & Best Parameters \\
		\midrule
		Optimization & AdaGrad, $\eta=1.0$, $\textit{igsav}=10^{-10}$\\
		Hot-Starting & activated \\
		Convergence SGD & $\gamma_{\textit{inner, active}}=5 \cdot 10^{-3}$, $\tau_{\textit{inner, active}} = 0.9$\\ &$\gamma_{\textit{inner, loss}}= 5 \cdot 10^{-5}$, $\tau_{\textit{inner, loss}} = 0.9$\\
		Convergence Feature Selection & by feature set fluctuation, $\gamma_{\textit{outer}}=0.01$\\
		\bottomrule
	\end{tabular*}
	\caption{The tuning results for the optimizer and convergence parameters.}
	\label{tab:tuning-overview}
\end{table}

\subsection{Tuning and Comparing AdaGrad and AdaDelta}\label{sec:optimization-hyperparameter}
This section aims at giving answers to the following questions: (1) How do the optimization parameters influence learning? (2) How fast is a relative convergence achieved? And (3), within a given number of training epochs, does AdaGrad or AdaDelta perform better? For that, AdaDelta's enhancement with a learning rate parameter as described in section~\ref{sec:imp-sgd} is considered, too.

The negative log-likelihood (see section~\ref{sec:objective}) of the training data serves as optimization objective and as performance measure in this section. The results show that, after tuning, AdaGrad achieves better objective values than AdaDelta, and does so significantly faster.
 
\subsubsection{Experiments}
The basic experimental setup was to have both optimizers minimize the objective for a limited number of 100 and 500 training epochs. The set of the 1-, 2- and 3-grams of all possible pitch sequences within this corpus was precomputed (9,723 features), without using the \nplus operator, and served as the feature set.

The free parameters of each optimizer were tuned in a combined grid search: For AdaGrad, the free parameters are the learning rate $\eta$ and the initial gradient squared accumulator value $\textit{igsav}$ (see section~\ref{sec:adagrad}), which were evaluated over the two-dimensional grid with $\eta \in \{0.01, 0.1, 1.0, 10.0\}$ and $\textit{igsav}\in\{10^{-6}, 10^{-7}, \dots,10^{-12}\}$. Vanilla AdaDelta has decay rate parameter $\rho$ and conditioning constant $\epsilon$ in the computation of the EMA (see section~\ref{sec:adadelta}). AdaDelta was evaluated on the grid over $\rho\in\{0.85, 0.9, 0.95, 0.99\}$ and $\epsilon\in\{10^{-3},10^{-4},\dots,10^{-9}\}$, while keeping the additional learning rate parameter $\eta=1.0$ to obtain the original update scheme. Subsequently, with the best values for $\rho$ and $\epsilon$, the learning rate $\eta$ was evaluated over the values $\eta \in \{1, 10, 100, 1000\}$.

\subsubsection{Results}
\Tab \ref{tab:ada} shows the results from the described experiments. The objective values on the two-dimensional parameter grid were interpolated and color encoded to provide an intuitive view on the results. Red colors represent high objective values, gray colors low values. Contour lines are drawn in intervals of 0.02. The interpolation encourages the sensible intuition that the values between the computed grid points can be assumed to continue smoothly. However, this representation disregards that an optimum may lie in between grid points. The exact objective values are given in appendix~\ref{app:sgd}.

\begin{table}[p] 
	\centering
	\subfloat[AdaGrad after 100 epochs]{
			\centering
			\begin{tabular}{c c c c c c c c c}
				\toprule
				& & \multicolumn{7}{c}{\textit{igsav}}\\
				\midrule
				& & $10^{-6}$& $10^{-7}$& $10^{-8}$& $10^{-9}$& $10^{-10}$& $10^{-11}$& $10^{-12}$\\
				\midrule
				\multirow{4}{*}{$\eta$} &\multicolumn{1}{r|}{0.01} & \multicolumn{7}{c}{\multirow{4}{*}{\includegraphics[width=8.5cm,height=1.8cm]{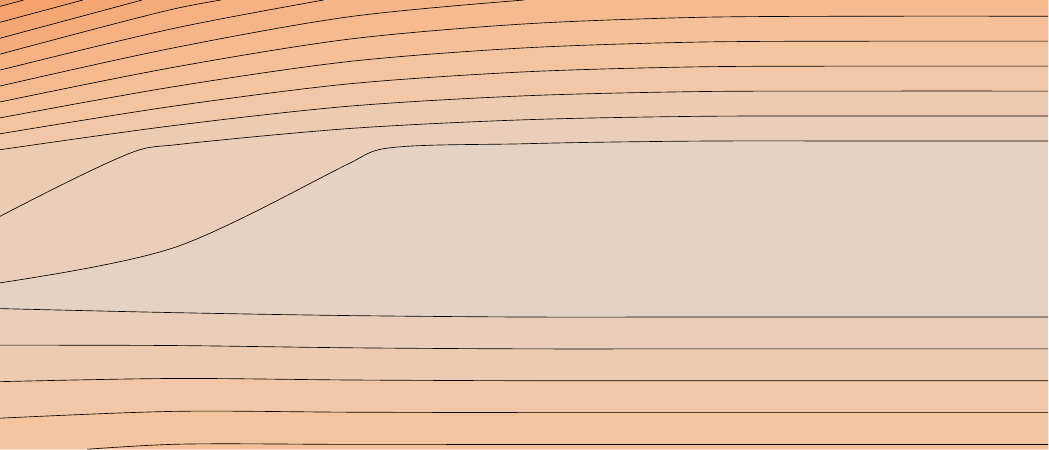}}}\\
				& \multicolumn{1}{r|}{0.1} & \multicolumn{7}{c}{} \\
				& \multicolumn{1}{r|}{1.0} & &&&&&\multicolumn{2}{c}{\textbullet~1.549~~~~~~}  \\
				& \multicolumn{1}{r|}{10.0} &  \multicolumn{7}{c}{} \\
				\bottomrule
		\end{tabular}
	\label{tab:adagrad100}}\\
\subfloat[AdaGrad after 500 epochs]{
	\centering
	\begin{tabular}{c c c c c c c c c}
		\toprule
		& & \multicolumn{7}{c}{\textit{igsav}}\\
		\midrule
		& & $10^{-6}$& $10^{-7}$& $10^{-8}$& $10^{-9}$& $10^{-10}$& $10^{-11}$& $10^{-12}$\\
		\midrule
		\multirow{4}{*}{$\eta$} &\multicolumn{1}{r|}{0.01}& \multicolumn{7}{c}{\multirow{4}{*}{\includegraphics[width=8.5cm,height=1.8cm]{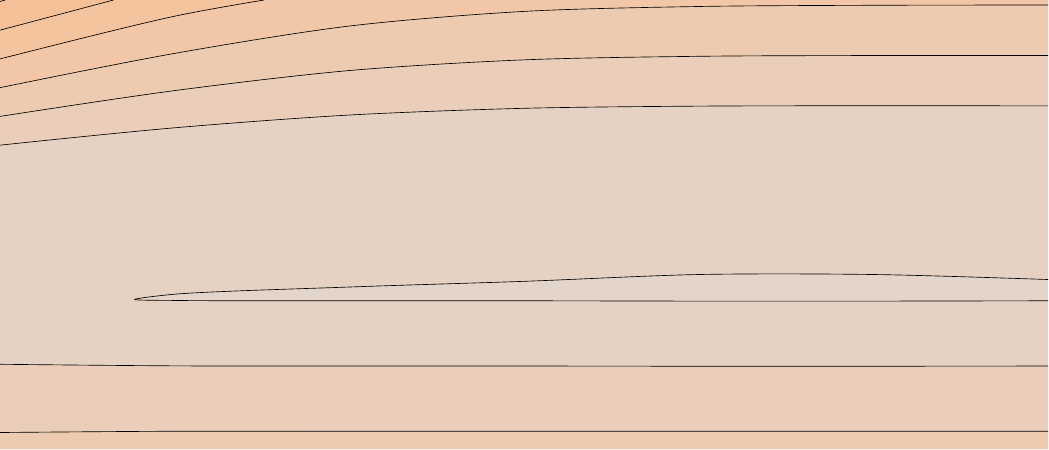}}}\\
		& \multicolumn{1}{r|}{0.1} & \multicolumn{7}{c}{} \\
		& \multicolumn{1}{r|}{1.0} & & & & & \multicolumn{2}{c}{\textbullet~1.540~~~~~~} &   \\
		& \multicolumn{1}{r|}{10.0} & \multicolumn{7}{c}{}   \\
		\bottomrule
\end{tabular}
\label{tab:adagrad500}}\\
	\subfloat[AdaDelta after 100 epochs ($\eta=1$)]{
			\centering
			\begin{tabular}{c c c c c c c c c}
				\toprule
				& & \multicolumn{7}{c}{$\epsilon$}\\
				\midrule
				& & $10^{-3}$& $10^{-4}$& $10^{-5}$ & $10^{-6}$& ~$10^{-7}$&~ $10^{-8}$&~ $10^{-9}$\\
				\midrule
				\multirow{4}{*}{$\rho$} &\multicolumn{1}{r|}{0.85} & \multicolumn{7}{c}{\multirow{4}{*}{\includegraphics[width=8.5cm,height=1.8cm]{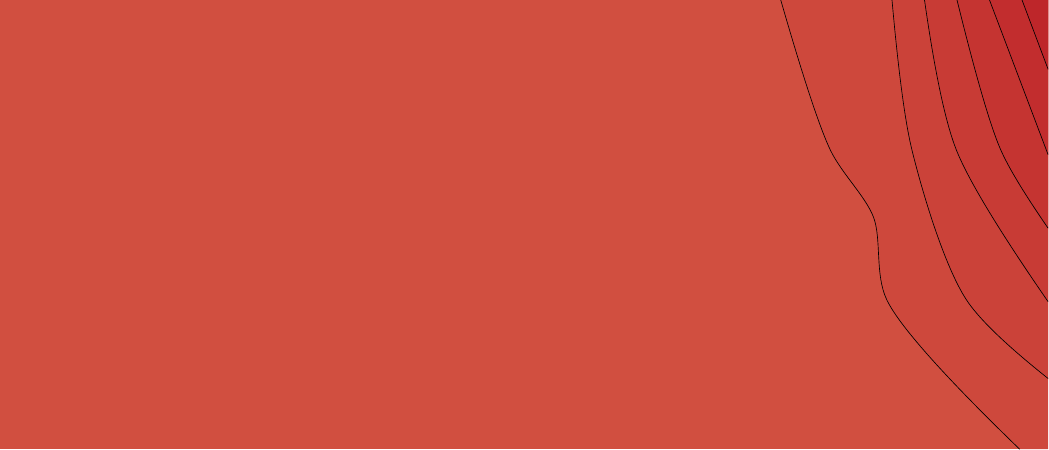}}}\\
				& \multicolumn{1}{r|}{0.90} &\multicolumn{7}{c}{}   \\
				& \multicolumn{1}{r|}{0.95} & \multicolumn{7}{c}{}   \\
				& \multicolumn{1}{r|}{0.99} & \multicolumn{2}{c}{\textbullet~2.011~~~~~~} &\multicolumn{5}{c}{}    \\
				\bottomrule
			\end{tabular}
		\label{tab:adadelta100}}\\
\subfloat[AdaDelta after 500 epochs ($\eta=1$)]{
	\centering
	\begin{tabular}{c c c c c c c c c}
		\toprule
		& & \multicolumn{7}{c}{$\epsilon$}\\
		\midrule
		& & $10^{-3}$& $10^{-4}$& $10^{-5}$ & $10^{-6}$& ~$10^{-7}$&~ $10^{-8}$&~ $10^{-9}$\\
		\midrule
		\multirow{4}{*}{$\rho$} &\multicolumn{1}{r|}{0.85} & \multicolumn{7}{c}{\multirow{4}{*}{\includegraphics[width=8.5cm,height=1.8cm]{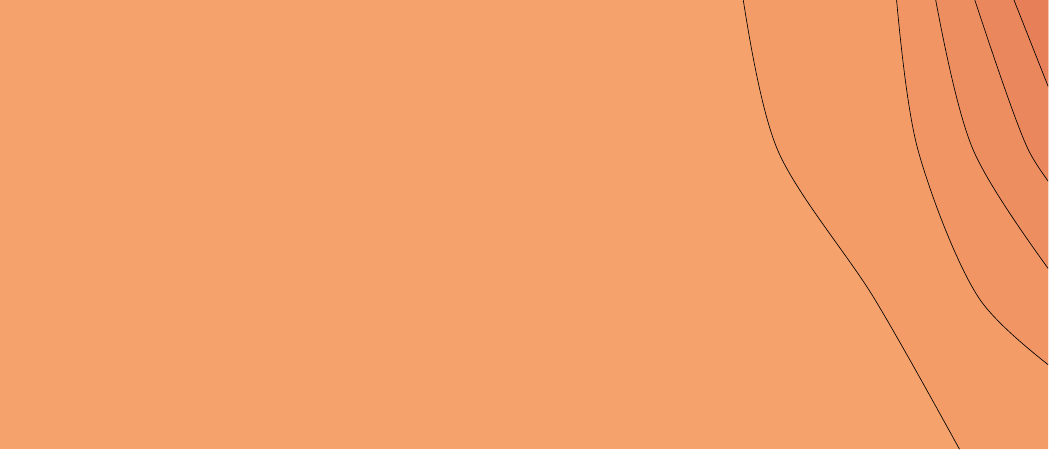}}} \\
		& \multicolumn{1}{r|}{0.90}  &\multicolumn{7}{c}{}   \\
		& \multicolumn{1}{r|}{0.95} &\multicolumn{7}{c}{}   \\
		& \multicolumn{1}{r|}{0.99} &\multicolumn{2}{c}{\textbullet~1.776~~~~~~} & \multicolumn{5}{c}{} \\
		\bottomrule
\end{tabular}
\label{tab:adadelta500}}\\
\subfloat[AdaDelta for different learning rates after 100 epochs]{
	\centering
	\begin{tabular}{c c c}
		\toprule
		\multirow{4}{*}{$\eta$} &\multicolumn{1}{r|}{1}& \cellcolor[rgb]{0.82,0.309,0.247}2.011\\
		& \multicolumn{1}{r|}{10} & \cellcolor[rgb]{0.957,0.694,0.509}1.716 \\
		& \multicolumn{1}{r|}{100} & \cellcolor[rgb]{0.929,0.796,0.69}\textbf{1.595}\\
		& \multicolumn{1}{r|}{1000} & \cellcolor[rgb]{0.945,0.78,0.655}1.605  \\
		\bottomrule
\end{tabular}
\label{tab:adadelta_lr100}}
\qquad
\subfloat[AdaDelta for different learning rates after 500 epochs]{
	\centering
	\begin{tabular}{c r c}
		\toprule
		\multirow{4}{*}{$\eta$} &\multicolumn{1}{r|}{1}& \cellcolor[rgb]{0.961,0.635,0.424}1.776\\
		& \multicolumn{1}{r|}{10} & \cellcolor[rgb]{0.945,0.78,0.655}1.616 \\
		& \multicolumn{1}{r|}{100} & \cellcolor[rgb]{0.898,0.8235,0.7608}\textbf{1.554}\\
		& \multicolumn{1}{r|}{1000} & \cellcolor[rgb]{0.945,0.78,0.655}1.601  \\
		\bottomrule
\end{tabular}
\label{tab:adadelta_lr500}}\\
\vspace{0.5cm}
\includegraphics[width=8.5cm]{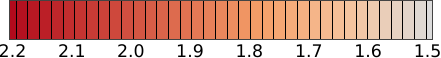}
	\caption{The objective values (negative log-likelihood) after 100 and 500 epochs of minimization using AdaGrad and AdaDelta with various hyperparameter settings. The values are color encoded according to the above legend. For the exact values please see appendix~\ref{app:sgd}.}
	\label{tab:ada}
\end{table}
\Tab\ref{tab:adagrad100} and \ref{tab:adagrad500} show the results for the AdaGrad optimization. Regarding (1), the first observation to make is that the learning rate $\eta$ has a larger influence than \textit{igsav}. Especially, values of \textit{igsav} smaller than $10^{-9}$ do not alter the results noticeably. For optimal values $\eta$ and \textit{igsav}, the gain from training for another 400 epochs is small. To answer (2), the optimization appears to have converged before or around 100 training epochs.

In \tab \ref{tab:adadelta100} and \ref{tab:adadelta500} the results for the decay rate parameter $\rho$ and conditioning constant $\epsilon$ are presented. The results confirm \textcite{zeiler2012adadelta}'s findings that AdaDelta is robust in the choice of $\rho$ and $\epsilon$. To answer (1), we can say that the parameters effect on the performance is negligible, especially if $\epsilon$ is chosen between $10^{-3}$ and~$10^{-7}$. For question (2) however, we realize that the optimization did not converge after 100 epochs and presumably also not after 500 epochs.

\Tab \ref{tab:adadelta_lr100} and \ref{tab:adadelta_lr500} show that the tuning of the added learning rate $\eta$ has a tremendous effect on the convergence speed of AdaDelta. Choosing $\eta=100$ makes the objective values significantly more competitive. Still, training for 500 instead 100 epochs improves the performance further.

Regarding (3), AdaGrad achieves immensely better results than vanilla AdaDelta, after both 100 and 500 epochs. Tuning the learning rate parameter for AdaDelta starkly reduces this gap, but still leaves AdaGrad to be the clear winner. In summary of AdaDelta's benefits, Zeiler writes that good~--~though less than optimal~--~results can be achieved without tuning any of the algorithm's parameters. In this experiment, we observe that a suboptimally tuned AdaGrad still outperforms an optimally tuned AdaDelta. We will use the AdaGrad optimizer with $\eta=1$ and $\textit{igsav}=10^{-10}$ in the following.

\subsection{Detecting Convergence}\label{sec:convergence}
The next hyperparameters we look at address the convergence of SGD optimization and the feature discovery loop. As described in section~\ref{sec:implementation}, the SGD optimization loop runs until convergence of either (1) the objective or (2) the active feature set, whatever comes first. The feature discovery loop is exited when the feature set converges, that means when it stops altering. The convergence criteria are chosen to avoid futile loop iterations to save computational expenses. An early (premature) stopping \parencite{prechelt1998automatic} of the optimization or feature discovery is avoided, as it introduces another layer of regularization which may distort the results of the regularization parameter selection in section~\ref{sec:per-feature-reg}.

The decay rate parameter $\tau_{\textit{inner, loss}}$ and the convergence threshold $\gamma_{\textit{inner, loss}}$ of the inner loop criterion (1) (see \eq \ref{eq:sgd-conv-loss}) were determined in preliminary experiments. Parameters $\gamma_{\textit{inner, loss}}= 5 \cdot 10^{-5}$ and $\tau_{\textit{inner, loss}} = 0.9$ for criterion (1) were chosen strictly to ensure a tight convergence. Typically criterion (2) (see \eq \ref{eq:sgd-conv-active}) will end the optimization much earlier than criterion (1), which then serves as a lower bound.

In the following, the three convergence criteria for the feature discovery loop (see section~\ref{sec:pypulse-conv}) are analyzed in combination with the inner loop criterion (2).

\subsubsection{Experiments}
To analyze the interplay of the outer loop criteria and the inner loop criterion (2), we consider a two-dimensional grid with value ranges $\gamma_{\textit{inner, active}},\gamma_{\textit{outer}} \in \{0.05, 0.01, 0.005, 0.001\}$ in each dimension. This is repeated for all three outer loop convergence criteria. Decay rate $\tau_{\textit{inner, active}}$ is chosen equal to $\tau_{\textit{inner, active}}$ to 0.9. As performance measure serves the cross-entropy on a randomly drawn test set of 10\% size of the full corpus. The model was configured to use exponential per-feature \lone regularization (see section~\ref{sec:pypulse-regularization}), with $\lambda_1=10^{-8.5}$, $\alpha =2.0$ and a PI*C*-\nplus configuration. 

\subsubsection{Results}
The computed test entropy values are given in \tab \ref{tab:conv}. The three tested convergence criteria show similar performances, with criterion (a) being the short winner. For all three criteria, the prime observation to make is that the influence of $\gamma_{\textit{outer}}$ on the result is smaller than that of $\gamma_{\textit{inner, active}}$. We can further observe, that the minimum value for $\gamma_{\textit{inner, active}}$ did not lead to the best performances. I speculate that this is due to a regularizing effect caused by early stopping of the optimization. Note that choosing either convergence threshold smaller leads to a big increase of execution time. 

\begin{table}[htb]
	\centering
	\subfloat[Convergence by feature set fluctuation.]{
		\begin{tabular}{cr| cccc}
			\toprule
			&& \multicolumn{4}{c}{$\gamma_{\textit{inner, active}}$}  \\
			&& 0.05 & 0.01 & 0.005 & 0.001 \\
			\midrule
			\multirow{4}{*}{$\gamma_\textit{outer}$}
			& 0.05   &2.309&2.274&2.264&2.272\\ 
			& 0.01   &2.292&2.273&2.261&2.273\\ 
			& 0.005 &2.292&2.270&2.261&2.271\\ 
			& 0.001 &2.287&2.267&\textbf{2.257}&2.269\\ 
			\bottomrule
		\end{tabular}
		\label{tab:conv-feature-diff}}\\
	\subfloat[Convergence by feature set size.]{
		\begin{tabular}{cr| cccc}
			\toprule
			&& \multicolumn{4}{c}{$\gamma_{\textit{inner, active}}$}  \\
			&& 0.05 & 0.01 & 0.005 & 0.001 \\
			\midrule
			\multirow{4}{*}{$\gamma_\textit{outer}$}
			& 0.05   &2.281&2.279&2.277&2.269\\ 
			& 0.01   &2.275&2.275&2.268&2.272\\ 
			& 0.005 &2.271&2.276&2.268&2.269\\ 
			& 0.001 &2.271&2.273&\textbf{2.266}&2.269\\ 
			\bottomrule
		\end{tabular}
		\label{tab:conv-feature-count}}\\
	\subfloat[Convergence by validation error.]{
		\begin{tabular}{cr| cccc}
			\toprule
			&& \multicolumn{4}{c}{$\gamma_{\textit{inner, active}}$}  \\
			&& 0.05 & 0.01 & 0.005 & 0.001 \\
			\midrule
			\multirow{4}{*}{$\gamma_\textit{outer}$}
			& 0.05   &2.280&2.282&2.266&2.280\\ 
			& 0.01   &2.275&2.275&2.268&2.272\\ 
			& 0.005 &2.275&2.276&2.268&2.271\\ 
			& 0.001 &2.271&2.273&\textbf{2.266}&2.269\\ 
			\bottomrule
		\end{tabular}
		\label{tab:conv-validation}}
	\caption{The test entropies for different feature discovery loop convergence criteria and convergence thresholds for the inner and outer loop.}
	\label{tab:conv}
\end{table}

On the one hand, we want to avoid a bleeding of the convergence parameters into the regularization parameters. On the other hand, stopping early saves computational resources. As a trade-off between time and performance I choose $\gamma_{\textit{inner, active}}=0.005$ to be at its optimum, but save resources by setting the less influential $\gamma_{\textit{outer}}=0.01$.

\subsection{Hot-Starting AdaGrad}\label{sec:experiments-hotstart}
We saw that tuning the SGD optimizer's parameters can have a tremendous effect on the speed of learning. Now, we will consider another means to reduce computational expenses linked to the optimization. In \pypulse, the optimizer is called in every feature discovery loop iteration. Hot-starting of the optimizer was proposed in section~\ref{sec:hot-start} as a means to speed up each run by taking over the accumulator values from the previous run. In this section, we will see that hot-starting AdaGrad can lead to a noticeable reduction of training epochs and improve the objective value.

\subsubsection{Experiments}
\pulse was run for 10 feature discovery loop iterations with and without hot-starting. The learner was configured to use exponential \lone regularization with $\lambda_1=10^{-8.5}$, $\alpha=2.0$, PI*C* \nplus and the same training and validation set as in section~\ref{sec:convergence}. The inner loop convergence parameters (optimization) were set to the optima from above and the outer loop was set to run for 10 iterations. As measure of goodness serves the objective values and number of training epochs until convergence. 

\subsubsection{Results}
From the results shown in \tab \ref{tab:hot-start}, we learn that hot-starting results in small gains in terms of the loss function as well as optimization duration. The feature set expansion stagnated from iteration 6 on; the improvements up to then are $\sim 1\%$ for the loss (in iteration 5) and $\sim 1\%$ for the number of epochs (iteration 1 to 5).

\begin{table}[tb]
	\centering
	\begin{tabular}{c| c c |c c}
		\toprule
		& \multicolumn{2}{c|}{hot-start off} & \multicolumn{2}{c}{hot-start on}\\
		iteration & objective &  epochs & objective &  epochs \\
		\midrule
		1 &  1.902 & 58 & 1.902 &58  \\
		2 &  1.678&69  & 1.678&62 \\
		3 &  1.563& 59 &1.563 &57 \\
		4 & 1.477 &57  &1.476 &55 \\
		5 &  1.441& 55 &1.433 &54 \\
		6 &  1.429&54  &1.421 &53 \\
		7 &  1.427&53  &1.417  &41\\
		8 &  1.426& 53 &1.414  &28  \\
		9 &  1.426& 53 & 1.413&20 \\
		10&1.420&90&1.412&2\\
		\bottomrule
	\end{tabular}
	\caption{The objective values and number of training epochs per feature discovery loop with (hot-start off) and without reset (on) of the AdaGrad and \lone accumulators between iterations.}
	\label{tab:hot-start}
\end{table}

From iteration 6 to 10, the number of new features ceases to grow and the effect of hot-starting shows in a faster convergence: The weights and accumulator values are close to an optimum, the repeated expansion with almost identical candidate sets is not improving the objective, and therefore the loss-based convergence criterion stops the learning. In contrast, without hot-starting, the optimizer has to relearn the accumulator and what features are the most meaningful ones.

Note that SGD convergence by the active feature set never got triggered before epoch 53. This is due to the inertia of the EMA with decay rate $\tau_{\textit{inner, active}}=0.9$ in combination with a small choice of $\gamma_{\textit{inner, active}}$. Thus, the benefits of hot-starting were in this experiment partially hidden by the small convergence threshold.

Repetitions of the experiment with different convergence parameters showed a similar picture. Having higher convergence threshold showed even stronger effects in the number of saved training epochs ($>10\%$). Due to its side-effect free benefits, I will use hot-starting in all remaining experiments and benchmarks.

\section{Reducing Overfitting}\label{sec:per-feature-reg} 
Tuning the regularization parameters means reducing overfitting. Additionally, employing per-feature regularization terms, as introduced in section~\ref{sec:pypulse-regularization}, injects top-down knowledge into the model and improves performance in general. We assume that the regularization terms are specific for music but dataset independent, and that only the global regularization factors depend on the respective dataset. The Bach chorales (dataset 1) were used for all experiments in this section.

It is shown that the LTM and STM perform best with \lone plus \ltwo regularization. Because of the temporal nature of the STM, a time dependent \lone and \ltwo regularization is additionally investigated, and thought to be good.

\begin{table}[htbp] 
	\centering
	\begin{tabular*}{\linewidth}{l | l}
		\toprule
		Model & Best Regularization Parameters \\
		\midrule
		LTM & exponential $L_1$ ($\alpha = 2.0$), constant \ltwo (per-feature \ltwo not tested)\\
		STM & exponential $L_1$ ($\alpha = 1.2$), constant \ltwo, temporal decay of \lone and $L_2$,\\
		& $\tau_1$ can be kept fixed\\
		\bottomrule                                                    
	\end{tabular*}
	\caption{Overview of the tuning results for the regularization hyperparameters.}
	\label{tab:per-feature-overview}
\end{table}

\subsection{LTM Regularization Terms}\label{sec:results-perfeature-ltm}
It is shown that for the LTM, the performance of the $L_1$-regularized \pulse model can be improved by penalizing the maximum feature depth exponentially, and by adding \ltwo regularization to the objective. 

\subsubsection{Experiments}
In a first step, to learn by what degree \ltwo regularization can improve the model, the best combination of \lone and \ltwo regularization was found on the grid $\lambda_1\in\{10^{-6}, 10^{-7}, 10^{-8}, 10^{-9}\}$ and $\lambda_2\in\{0.0, 10^{-12}, 10^{-11}, 10^{-10}, 10^{-9}, 10^{-8}, 10^{-7}\}$. In a second step, the different feature depth dependent regularization functions (see section~\ref{sec:pypulse-perfeature}) were compared for \lone regularization and different parameterizations. For that, a finer spaced grid with $\lambda_1\in\{10^{-7}, 10^{-7.5}, 10^{-8}, 10^{-8.5}, 10^{-9}\}$ was used. 

The convergence parameters were set to $\gamma_{\textit{outer}}=0.05$, $\gamma_{\textit{inner, loss}}= 5 \cdot 10^{-5}$ and $\gamma_{\textit{inner, active}}=0.01$. As \nplus configuration, PI*C* was used. The performance was measured in cross-entropy bits, on a randomly drawn test set of 10\% size of the full dataset (as in section~\ref{sec:convergence}). 

For computational reasons, the feature discovery loop was stopped after the feature set size exceeded 2,000. Hence, the found performances are worse or equal to their true optimum.

\subsubsection{Results}
The prime observation to be made from \tab \ref{tab:ltm-l1-l2} is that adding \ltwo regularization considerably improves the performance of the LTM. On the downside, \ltwo regularization increases the feature count and with it the runtime. \ltwo regularization encourages many features with small weights, whereas \lone regularization encourages smaller feature sets with larger weights. In consequence, the two counteract each other.

For the following two reasons  I use \lone regularization despite the worse performance regarding the LTM: (1) According to Occam's razor, the simpler model is preferable. The $L_1$-only model has less hyperparameters and much fewer weights. Benefits may include a better generalization performance, and an increased musicological interpretability of less but more expressive features. (2) The practicability of the model and method is at stake when the runtime exceeds the user's patience. Especially, the hyperparameter selection during 10-fold CV is immensely time-consuming.
 
\begin{table}[htbp]
	\centering
	\begin{tabular}{cc| ccccccc}
		\toprule
		&& \multicolumn{7}{c}{$\lambda_2$}  \\
		 && 0 & $10^{-12}$& $10^{-11}$ & $10^{-10}$ & $10^{-9}$ & $10^{-8}$ & $10^{-7}$ \\
		\midrule
		\multirow{4}{*}{$\lambda_1$}
		&$10^{-6}$ &2.743&2.743&2.743&2.743&2.743&2.745&2.767\\
		&$10^{-7}$ &2.386&2.386&2.386&2.386&2.381&2.384&2.452\\
		&$10^{-8}$ &2.374&2.379&2.376&2.380&2.347&\textbf{2.287}&2.376\\
		&$10^{-9}$ &2.460&2.459&2.451&2.417&2.354&2.335&2.422\\
		\bottomrule
	\end{tabular}
	\caption{Joint \lone and \ltwo regularization in the LTM.}
	\label{tab:ltm-l1-l2}
\end{table}

\Tab \ref{tab:ltm-l1-perfeature} lists the result of the per-feature regularization terms, for different parameterizations. It is striking that all depth dependent terms performed better than a constant global factor (const.). The exponential factor with $\alpha=2.0$ is the tight winner. In general, the polynomial and exponential approaches performed better than the linear ones. 

Whether features with depth zero are regularized or not was deemed relevant. However, the results show that neither shifting the linear factor up by one to regularize zero-depth features (lin-no-0), nor modifying the exponential term to not regularize zero-depth features (exp-0), performed better than the original functions.

\begin{table}[htb]
	\begin{tabular}{l| c|cccc|cccc }
		\toprule
		& const.&\multicolumn{4}{c|}{linear}&\multicolumn{4}{c}{lin-no-0} \\
		\multicolumn{1}{c|}{$\lambda_1$} & 1.0& 0.5 &  1.0 & 2.0 &  4.0& 0.5 &  1.0 & 2.0 &  4.0  \\
		\midrule
		$10^{-7}$     &2.386&2.346&2.427&2.485&2.533&2.441&2.490&2.516&2.563\\
		$10^{-7.5}$  &\textbf{2.360}&2.302&2.313&2.367&2.442&2.313&2.344&2.393&2.463\\
		$10^{-8}$     &2.374&2.340&2.301&\textbf{2.294}&2.327&2.343&2.307&\textbf{2.300}&2.333\\
		$10^{-8.5}$  &2.368&2.333&2.357&2.328&2.300&2.359&2.330&2.354&2.301\\
		$10^{-9}$     &2.460&2.441&2.377&2.328&2.333&2.378&2.346&2.330&2.328\\
		\bottomrule
	\end{tabular}	
	\begin{tabular}{l| ccc| cccc | c}
		\toprule
		& \multicolumn{3}{c|}{polynomial} &\multicolumn{4}{c|}{exponential} & exp-0 \\
		\multicolumn{1}{c|}{$\lambda_1$} & 0.5 &  2.0 & 4.0 & 1.2 & 1.5 & 2.0 & 2.5 & 2.0\\
		\midrule
		$10^{-7}$    &2.387&2.460&2.511&2.407&2.463&2.499&2.518&2.493\\
		$10^{-7.5}$  &2.324&2.363&2.455&2.306&2.328&2.382&2.426&2.378\\
		$10^{-8}$    &2.351&2.308&2.392&2.363&2.287&2.301&2.333&2.300\\
		$10^{-8.5}$  &2.343&\textbf{2.276}&2.333&2.331&2.321&\textbf{2.274}&2.290&\textbf{2.289}\\
		$10^{-9}$    &2.423&2.307&2.302&2.420&2.362&2.318&2.292&2.301\\
		\bottomrule
	\end{tabular}
	\caption{Comparison of different per-feature regularization functions and parameterizations in the LTM.}
	\label{tab:ltm-l1-perfeature}
\end{table}

\subsection{STM Regularization Terms}\label{sec:results-perfeature-stm}
From the analysis of combined \lone and \ltwo regularization, we learn the significance of \ltwo for the STM. Furthermore, we see that a temporal decay of the regularization parameters improves the performance.

\subsubsection{Experiments}
The many free parameters of the STM were evaluated one-by-one, instead of jointly.
I determined,
\begin{enumerate}
	\item[(1)] whether \ltwo regularization is beneficial on a combined grid over $\lambda_{1}$ and $\lambda_{2}$ with $\lambda_1\in\{10^{-2}, 10^{-3}, \dots, 10^{-7}\}$ and $\lambda_2\in\{10^{-2}, 10^{-3}, 10^{-4}, 10^{-5}, 0.0\}$,
	\item[(2)] the best \lone per-feature regularization term and parameter $\alpha$,
	\item[(3)] the best \ltwo per-feature regularization term and parameter $\alpha$,
	\item[(4)] whether temporal decay of $\lambda_1$ and $\lambda_2$ improves the performance, and
	\item[(5)] the impact of each of the parameters $\lambda_1^\textit{init}$, $\lambda_2^\textit{init}$, $\tau_1$ and $\tau_2$ on a four-dimensional grid over $\lambda_1^\textit{init}\in\{10^{-2}, 10^{-3}, 10^{-4}, 10^{-5}\}$, $\lambda_2^\textit{init}\in\{10^{-1}, 10^{-2}, 10^{-3}, 10^{-4}\}$ and $\tau_1,\tau_2\in\{1, 10, 100\}$.
\end{enumerate}
Note that for (2) and (3) only the exponential term was evaluated based on the experience from the LTM per-feature regularization results. For the value ranges see tables~\ref{tab:stm-l1-perfeature} and \ref{tab:stm-l2-perfeature}. Experimental setup for (5): For each parameter $\varphi\in \{\lambda_1^\textit{init}, \lambda_2^\textit{init}, \tau_1, \tau_2\}$ and value within its value range, a three-dimensional grid search was performed over the remaining parameters and their value ranges respectively. The best achieved performance for each value of $\varphi$ was stored. 

For computational reasons, the training/test dataset was made up of only 10 out of 185 randomly chosen Bach chorales. The P*IC-\nplus configuration was employed.

\subsubsection{Results}
(1): It is striking how much performance was gained by adding \ltwo regularization (see \tab \ref{tab:stm-l1-l2}). Compared to the LTM, the gain is more than four times larger. This can be explained as follows: In the beginning, due to very little training data, the STM is falsely certain in its beliefs. That means the probability vectors have high peaks (low entropy). A wrong guess seriously impairs the performance, as every class besides the peak has very low likelihood. Adding \ltwo regularization enforces a Gaussian prior over the weight distribution, and the probability vectors become more leveled (higher entropy). Hence, wrong guesses have a reduced impact with \ltwo regularization.

\begin{table}[htb]
	\centering
	\begin{tabular}{cc| ccccc}
	\toprule
	&& \multicolumn{5}{c}{$\lambda_2$}  \\
	&& 0 & $10^{-5}$ & $10^{-4}$ & $10^{-3}$ & $10^{-2}$ \\
	\midrule
	\multirow{4}{*}{$\lambda_1$}
	&$10^{-2}$ &4.327&4.328&4.325&4.286&4.263\\
	&$10^{-3}$ &3.677&3.688&3.641&3.579&3.965\\
	&$10^{-4}$ &4.534&4.325&3.581&\textbf{3.288}&3.857\\
	&$10^{-5}$ &6.328&4.597&3.628&3.299&3.835\\
	&$10^{-6}$ &7.223&4.722&3.716&3.300&3.838\\
	&$10^{-7}$ &8.647&4.872&3.731&3.310&3.842\\
	\bottomrule
	\end{tabular}
	\caption{Joint \lone and \ltwo regularization in the STM.}
	\label{tab:stm-l1-l2}
\end{table}

(2) and (3): \Tab \ref{tab:stm-l1-perfeature} and \ref{tab:stm-l2-perfeature} reveal that \lone regularization performs best with exponential per-feature factor and $\alpha=1.2$. Further, we see that \ltwo regularization performs best with a global factor. These results are in accordance with the intent behind the implementation of feature depth dependent regularization; to guide feature selection (see section~\ref{sec:pypulse-regularization}). \ltwo regularization does not drive weights to zero, and should thus penalize all features equally.

\begin{table}[htbp]
	\centering
	\begin{tabular}{l| c |cccc}
		\toprule
		& const. &\multicolumn{4}{c}{exponential} \\
		\multicolumn{1}{c|}{$\lambda_1$} & 1.0 & 1.2 & 1.5 & 2.0 & 2.5 \\
		\midrule
		$10^{-5}$  &3.299&3.248&3.263&3.275&3.293\\
		$10^{-6}$  &3.300&\textbf{3.246}&3.251&3.252&3.254\\
		$10^{-7}$  &3.310&3.267&3.277&3.262&3.252\\
		\bottomrule
	\end{tabular}
	\caption{Per-feature \lone regularization with $\lambda_2=10^{-3}$ in the STM.}
	\label{tab:stm-l1-perfeature}
\end{table}

\begin{table}[htbp]
	\centering
	\begin{tabular}{l| c |cccc}
		\toprule
		& const. &\multicolumn{4}{c}{exponential} \\
		\multicolumn{1}{c|}{$\lambda_2$} & 1.0 & 1.2 & 1.5 & 2.0 & 2.5 \\
		\midrule
		$10^{-2}$  &3.807&3.825&3.849&3.865&3.873\\
		$10^{-3}$  &\textbf{3.246}&3.246&3.264&3.277&3.283\\
		$10^{-4}$  &3.581&3.493&3.456&3.435&3.438\\
		$10^{-5}$  &4.377&4.318&4.250&4.193&4.156\\
		\bottomrule
	\end{tabular}
	\caption{Per-feature \ltwo regularization with $\lambda_1=10^{-6}$ and exponential $\lambda_1$ per-feature factor with $\alpha=1.2$ in the STM.}
	\label{tab:stm-l2-perfeature}
\end{table}

(4): Adding a temporal decay to both regularization vectors lead to a considerable improvement from $3.246$ to $2.964$ bits. 

\begin{figure}[htb]
	\centering
	\includegraphics[width=.7\textwidth]{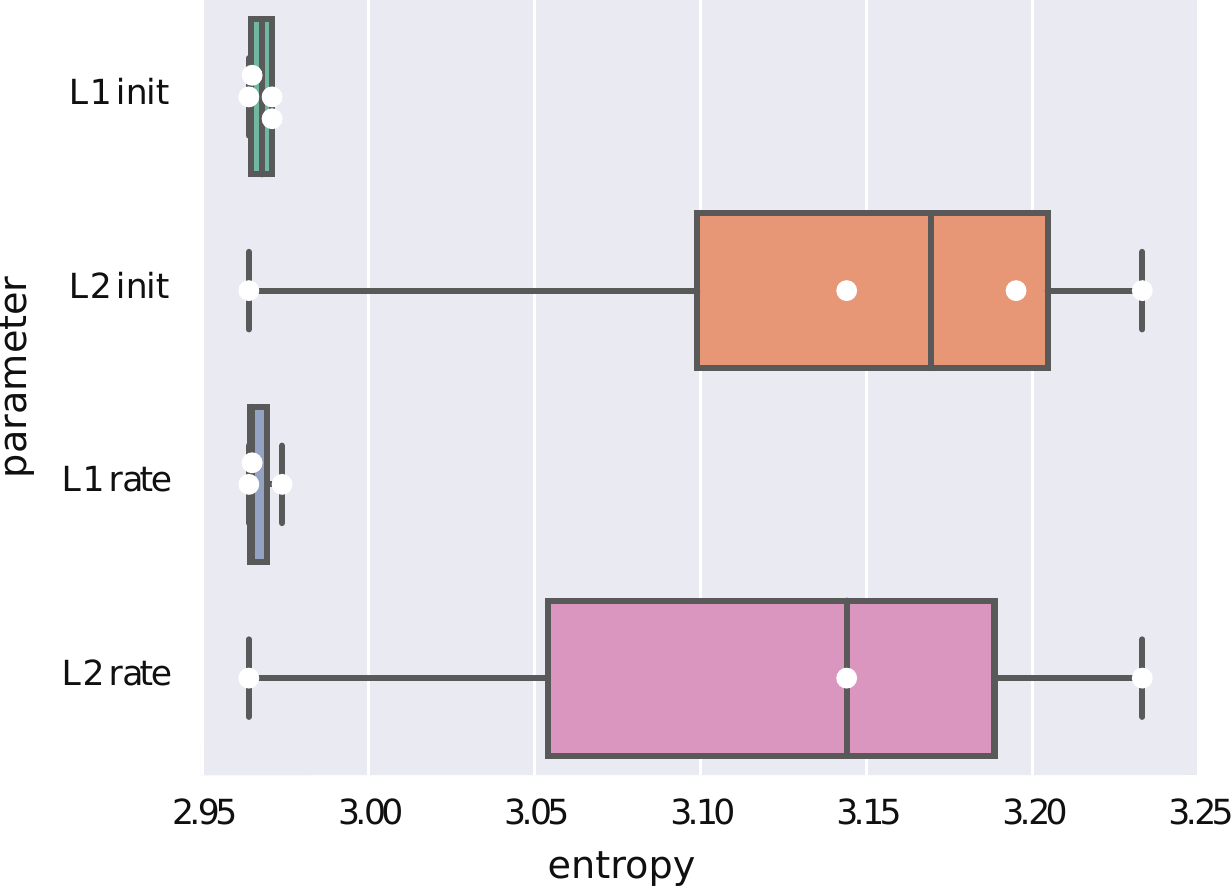}
	\caption{Box plots of the best attainable performances if one STM regularization parameter is held constant (for several values), and grid search is done over the remaining three hyperparameters. The white dots represent the respective achieved entropy values.}
	\label{fig:stm_4dim}
\end{figure}

(5): For each experiment, one parameter $\varphi\in \{\lambda_1^\textit{init}, \lambda_2^\textit{init}, \tau_1, \tau_2\}$ was held constant while performing hyperparameter optimization over the remaining three. The box plot in \fig \ref{fig:stm_4dim} visualizes the mean and variance of the best achieved performances. From the small variances of $\lambda_1^\textit{init}$ and $\tau_1$, we conclude that the optimization results barely depended on the value of either. To reduce the hyperparameter search space, we will keep $\tau_1=100$ constant in the following. Consequently, the \pypulse STM will from hereon have the free parameters $\lambda_1^\textit{init}$, $\lambda_2^\textit{init}$ and $\tau_2$.

\section{Comparing \texorpdfstring{\nplus}{N+} Operators and Feature Combinations}\label{sec:feature-combos}
This section compares several \nplus operators that use a variety of feature combinations, to find the best performing model. I aimed at covering a relevant subspace of all possible LTM and STM \nplus operators. An automated search for the best combination was not possible because of high computational costs. The best feature combinations found are given in \tab \ref{tab:features-combos-overview}. The LTM experiments were conducted on the Bach chorales, the STM experiments on the German nursery rhymes dataset.

\begin{table}[tb] 
	\centering
	\begin{tabular}{l | l}
		\toprule
		Model & Features \\
		\midrule
		LTM & PI*C*KM$_K$\\
		STM & PI*F$_1$\\
		\bottomrule                                                    
	\end{tabular}
	\caption{The best found \nplus configurations per model.}
	\label{tab:features-combos-overview}
\end{table}

\subsection{LTM}\label{sec:ltm-configs}
The space of all \nplus operators is searched in two steps. First, the best set of viewpoint features (see section~\ref{sec:pypulse-viewpoint}) was uncovered to be PI*C*, by testing musically sensible combinations. Next, this set was used as basis for further extension with anchored and linked features (see sections~\ref{sec:anchored-features} and \ref{sec:linked-features}). Overall, PI*C*KM$_K$ is the best performing feature combination.

\subsubsection{Experiments}
\Tab \ref{tab:ltm-nplus} lists all tested \nplus combinations. For computational reasons, two compromises had to be made: (1) Only \lone regularization was used, with the exception of two proofs of concept in \tab \ref{tab:ltm-viewpoint-l12}. (2) The benchmarks were run on one out of eight datasets of the Pearce corpus only. 

The Bach chorales were chosen as benchmarking dataset, as they contain a large number of well formed melodies, and are the most widely used set. As performance measure served the average cross-entropy over 10 CV folds. In every fold, 10\% of the training data was left out to validate $\lambda_1\in[10^{-9},10^{-6}]$ using GP optimization for 30 samples.

\subsubsection{Results}
The entropy values for all tested LTM configuration are listed in \tab \ref{tab:ltm-nplus}. Amongst them, PI*C*KM$_K$ is the overall winner. The remainder of the results allow for several observations and insights (for a music theoretic model analysis please see \ref{sec:feature-analysis}):
\begin{itemize}
	\item PI*C*, the highest performing configuration in \tab \ref{tab:ltm-viewpoint-benchmark}, has a persuasive musical interpretation and motivation: P learns tone profile, I* learns motifs in a transposition invariant way and C* learns melody contours.
	\item The superiority of I* to P* mirrors findings for melody learning in humans. Children start off with learning melodies based on absolute pitches, but will evolve to memorize melodies using relative pitches as they become adults \parencite{saffran2001absolute}.
	\item We observe that O* performs surprisingly poorly on its own. It is now evident that octave invariant intervals alone are unsuitable for learning melodies.
	\item Adding anchored features that learn tonic or key specific tone profiles was a natural candidate to improve the performance. It is interesting to see, that using the first note(s) as tonic estimator performs very similar to using the tonic as computed by the Krumhansl-Schmuckler algorithm. However, additionally using the computed mode leads to a much better performance.
	\item M$_K$ was designed based on the hypothesis that there are correlations between metrical weights and scale degree frequencies. Compared to pure K features, we can observe an improvement, and hence corroborate this hypothesis. Similarly, but to a smaller extent, M$_P$ features improve the performance and hint at a correlation between pitch values and the position in bar.
	\item In several instances having more features or higher value ranges lead to worse results, despite a potential higher expressiveness. \Eg F$_{1,2,3}$ compared to F$_1$, PI*X* compared to PI*C*, P*I* compared to PI*. Moreover, all combinations expanded in an intermingled fashion: P(IC)* compared to PI*C*, (PI)* compared to P*I* and (PIC)* compared to PI*C*. \textcite{ng2004feature} showed that for feature selection scenarios in logistic regression, the number of training samples needs to grow at least logarithmically with the number of irrelevant features. I suspect that a similar relationship holds for log-linear models, and that the fairly small datasets inhibit the full potential of several~--~especially the intermingled~--~configurations.
	\item The difference in feature set size between configurations is striking. For example, P* learned a model of 1236 features (averaged over the CV folds), while PI*C*K reached approximately the same training entropy and a much better test entropy with only 384 features. The best performing configuration PI*C*KM$_K$ converged to a set of 447 features. I attribute a low number of features to a well balanced set of expressive features. For example, P* is not as suited to memorize melodies as I*. A beneficial side effect of having smaller feature sets is a significantly reduced runtime.
	\item Additionally using \ltwo regularization turned out to achieve only marginal improvements in performance, paid for with much higher training durations ($\sim 1.5-2\times$ as long). In consequence, the number of features for P* rose up to 1,583.
\end{itemize}

\begin{table}[htb]
	\centering
	\begin{tabular}{cc}
		\subfloat[Viewpoints features.]{
			\begin{tabular}{c | c | c | c}
				\toprule
				\multicolumn{3}{c|}{\nplus}  & Entropy \\
				\midrule
				\multicolumn{3}{c|}{P}& 3.617\\
				\multicolumn{3}{c|}{P*}& 2.393\\  
				\multicolumn{3}{c|}{I}&3.020 \\
				\multicolumn{3}{c|}{I*}& 2.382\\
				\multicolumn{3}{c|}{O}&3.719 \\
				\multicolumn{3}{c|}{O*}& 3.149\\
				\midrule
				\multirow{3}{*}{P*}&\multicolumn{2}{c|}{I}& 2.387\\
				&\multicolumn{2}{c|}{I*}&2.307 \\
				&\multicolumn{2}{c|}{O*}&2.331 \\
				\midrule
				\multirow{6}{*}{P}&\multicolumn{2}{c|}{O*}& 2.588\\
				&\multicolumn{2}{c|}{(IC)*}& 2.305\\\cline{2-3}
				&\multirow{4}{*}{I*}&--& 2.302\\
				&&C& 2.300\\
				&&C*& \textbf{2.295}\\
				&&X*& 2.297\\
				\midrule
				\multicolumn{3}{c|}{(PI)*}& 2.330 \\
				\multicolumn{3}{c|}{(PO)*}& 2.366\\
				\multicolumn{3}{c|}{(PIC)*}& 2.331\\
				\bottomrule
			\end{tabular}
			\label{tab:ltm-viewpoint-benchmark}}\qquad\qquad&
		\begin{tabular}{c}
			\subfloat[Anchored features.]{ 
				\begin{tabular}{c| c | c}
					\toprule
					\multicolumn{2}{c|}{\nplus} & Entropy \\
					\midrule
					\multirow{8}{*}{PI*C*} &F$_1$& 2.263\\
					&F$_{1,2,3}$& 2.264\\  
					&T& 2.266\\
					&K& 2.209\\  \cline{2-3}
					&M$_P$& 2.289\\
					&M$_K$& 2.188\\
					&TM$_T$& 2.250\\
					&KM$_K$& \textbf{2.187}\\
					\bottomrule
				\end{tabular}
				\label{tab:ltm-anchor-benchmark}}\bigskip\\
			\subfloat[Results for combined \lone and \ltwo regularization.]{ 
				\begin{tabular*}{3.2cm}{c | c}
					\toprule
					\nplus & Entropy \\
					\midrule
					P* & 2.384 \\
					PI*C* & \textbf{2.290}\\
					\bottomrule
				\end{tabular*}
				\label{tab:ltm-viewpoint-l12}}
		\end{tabular}
	\end{tabular}
	\caption{Entropy values for the Bach chorales dataset using 10-fold CV and GP optimized \lone regularization for different \nplus configurations in the LTM. Each path from left to right in the \nplus column describes one configuration.}
	\label{tab:ltm-nplus}
\end{table}

Besides cross-entropy, another metric worth considering is the empirical classification error. Its advantage is its intuitiveness. Averaged over all CV folds, $47.24\%$ of the pitches in the test set have been correctly guessed by PI*C*KM$_K$. P* classified $43.40\%$ of all pitches correctly. Most misclassifications naturally occur in the beginning of each song where the contexts are smallest.

\subsection{STM}\label{sec:stm-configs}
Using the same approach as for the LTM, we assess different STM \nplus configurations. This time the German nursery rhymes dataset was used, which proved to be the easiest dataset to predict for the STM \parencite{pearce2004}. In an additional preliminary step, forwards and continuous expansion were compared. Then, building on knowledge gained from the LTM, a smaller set of feature combinations was chosen and evaluated. The \nplus operator PI*F$_1$ with forwards expansion performed best.

\subsubsection{Experiments}
Firstly, the best \nplus mode was determined by running benchmarks for forwards and continuous expansion using P*. Secondly, several viewpoint and anchored features were combined and the performances compared. Due to high computational costs, it was infeasible to optimize the hyperparameters for every CV fold. Thus, the hyperparameters  $\lambda_1^\textit{init}\in [10^{-5},10^{-2}]$, $\lambda_2^\textit{init}\in[10^{-3},10^{-1}]$, $\tau_2\in[5,12]$ (see section~\ref{sec:results-perfeature-stm}) were selected via GP based optimization for 40 samples on the whole corpus. Let $\tau_2=100$ be fixed. The reported benchmarks are the average over the 10 test set performances in 10-fold CV.

The German nursery rhymes dataset was used as benchmarking corpus because the best results of prior work on the STM have been achieved for this dataset. Moreover, it is rich in repetitions which makes it a fruitful application ground for the STM.

\subsubsection{Results}
At first, we consider the differences between the \nplus expansion modes. The best performance achieved for the forwards mode was 2.737, whereas the continuous mode only achieved 3.055 bits. The expectation that continuous expansion would perform reasonably well has not been met (see section~\ref{sec:pypulse-nplus}). Forwards expansion, which captures motifs instantly in contrast to continuous expansion, proved to be the superior strategy.

The benchmarks for different \nplus operators using forwards expansion are given in \tab \ref{tab:stm-feature-combos}. The first observation to make is that the STM benefits from an absolute learning of motifs with P* features, in contrast to the LTM, in which a relative learning of pitch sequences performed better. It is sensible to assume that the LTM generalizes better between songs by describing melodies in a relative manner, and that the STM memorizes a single song better by storing motifs using absolute pitches.

\begin{table}
	\centering
	\begin{tabular}{c | c}
		\toprule
		\nplus & Entropy \\
		\midrule
		P*& 2.737\\
		I*& 3.061\\  
		\midrule
		P*I& 2.685\\
		PI*& 2.657\\
		P*I*& 2.677\\
		\midrule
		PI*C & 2.684 \\
		PI*C*& 2.705\\
		PI*F$_1$ & \textbf{2.590} \\
		PI*K & 2.675 \\
		PI*C*F$_1$ & 2.624\\
		\bottomrule
	\end{tabular}
	\caption{Entropy values for different STM \nplus configurations on the German nursery rhymes dataset. The benchmarks were computed using 10-fold CV and GP optimized \lone and \ltwo regularization with temporal decay.}
	\label{tab:stm-feature-combos}
\end{table}

Similar to the concept that a fish cannot ponder about the significance of water, the concept of musical keys only makes sense when songs are compared to one another. Thus, in the STM, the concept of musical key does not exist, and it is surprising that the combination of PI* with F$_1$ resulted in the overall best performance. The \nplus operator F$_1$ describes nothing more than P, namely the value of a pitch in reference to the first tone or to MIDI pitch zero, respectively. I make the conjecture that this finding is caused by the GP optimizer exploiting a local optimum and by 40 samples not being enough to sufficiently explore a three-dimensional space. Moreover, this could mean that other reported results are not globally optimal neither. Due to a lack of time, this was not further investigated.

\section{Comparison of Hybrid Models}\label{sec:hybrid-configs}
Ensembles of classifiers have been shown to surpass their source models in the past. For example, MVS (see section \ref{sec:MVS} ) using the mean and product rule (see section~\ref{sec:combi-rules}) to combine single viewpoint models, or the top Netflix Prize performer \textcite{bell2008bellkor} combing over a hundred models building on \textcite{wolpert1992stacked}'s stacking method. In this section, I analyze the combination of \pulse LTMs and STMs as well as LTMs with LTMs. The combinations were performed using the mean and product rule using different parameterizations.

All tested hybrid models were found to perform better than each of the source models. \pulse LTMs were also combined with \ngram (C*I) and (X*UI)-STM (see \textcite{pearce2004} for an explanation of the shorthand model identifiers).

The combination of the PI*C*KM$_K$-LTM with the \ngram (C*I)-STM using the mean rule performed best. The combination of the P* and the I* LTMs outperformed the joint PI*-LTM.

\subsection{LTM+STM}
Augmenting LTM with STM predictions is a natural way to improve performance. The LTM captures style dependent patterns, whereas the STM learns song specific motifs. The combination of the best \pulse LTM with the best \pulse STM (see sections~\ref{sec:ltm-configs} and \ref{sec:stm-configs}) and with two \ngram STMs taken from \textcite{pearce2004} are analyzed in this section.

The combination rules come with the free parameter $b$ which assigns a bias to the lower entropy distribution (see \eq \ref{eq:bias}). In preliminary experiments, the effect of separate parameters $b$ for the LTM and STM, as well as the effect of a time-dependent shift to the LTM in the beginning and the STM in the end, were investigated. Both approaches were outperformed by hybrids using the original single parameter $b$.

\subsubsection{Experiments} 
The following hybrids were combined: (1) The \pulse PI*C*KM$_K$-LTM with the \pulse PI*F$_1$-STM, (2) the \pulse PI*C*KM$_K$-LTM with the \ngram (C*I)-STM, and (3) the \pulse PI*C*KM$_K$-LTM with the \ngram (X*UI)-STM. The mean and product combination rule from section~\ref{sec:combi-rules} were used. For the LTM+STM hybrids the entropies were computed on the whole Pearce corpus, for the LTM+LTM hybrid the Bach chorales were used.

Bias parameter $b$ was determined for each combination rule over the grid $b \in \{0, 1, 2, 3, 4, 5, 6, 16, 32\}$ on the training set of each CV fold (in \textcite{pearce2004,cherla_hybrid_2015} $b$ was determined over the same grid but on the test set).

The distributions for the \ngram STMs were obtained with IDyOM (version 1.4). Note that the pitch sequences parsed by IDyOM did not match the \textit{**kern} files in five out of 54,308 events. This disaccord was resolved by making the affected IDyOM distributions uniform.

\subsubsection{Results}
The results for the LTM+STM hybrids (1) to (3) are given in \tab \ref{tab:results-ensembles}. In all cases, the mixture performed much better than either of the source models. The \textit{P\textsmaller{ULSE}}+\pulse hybrid performed better than the \textit{P\textsmaller{ULSE}}+\ngram (X*UI) but worse than the \textit{P\textsmaller{ULSE}}+\ngram (C*I) hybrid. It is intriguing, that the \ngram (C*I)-STM, that performs worse on its own, leads to an overall better performance when employed in a hybrid. I make the conjecture that this is due to the heterogeneity of the source models, which facilitates a larger gain of information during the combination, compared to combining models of the same breed.

\begin{table}[t]
	\centering
	\scalebox{0.95}{
	\begin{tabular}{l | c | c | c}
		\toprule
		Ensemble & Rule & Entropy  & Source Entropies \\
		\midrule
		\multirow{2}{*}{\pulse LTM PI*C*KM$_K$ + \pulse STM PI*F$_1$}  & m & 2.387 &\multirow{2}{*}{2.542 / 3.092}\\
		& p & 2.397&\\
		\midrule
		\multirow{2}{*}{\pulse LTM PI*C*KM$_K$ + \ngram STM (C*I)}  & m& \textbf{2.357} &\multirow{2}{*}{2.542 / 3.152}\\
		& p & 2.394&\\
		\midrule
		\multirow{2}{*}{\pulse LTM PI*C*KM$_K$ + \ngram STM (X*UI)}  & m& 2.407 &\multirow{2}{*}{2.542  3.149}\\
		& p & 2.400&\\
		\midrule
		\midrule
		\multirow{2}{*}{\pulse LTM P*+ \pulse LTM I*} & m& 2.299 & \multirow{2}{*}{2.393 / 2.382} \\
		& p & 2.282&\\
		\bottomrule
	\end{tabular}}
	\caption{Entropy values for ensemble models using mean (m) and product (p) combination rule. The values are computed on the whole Pearce corpus for the LTM+STM hybrids and on the Bach chorales dataset for the LTM+LTM hybrid.}
	\label{tab:results-ensembles}
\end{table}

\subsection{LTM+LTM}
Combining models of the same paradigm, such as LTMs with LTMs or STMs with STMs, stands in contrast to a joint approach as pursued with \textit{P\textsmaller{ULSE}}. The former approach is used by \ngram MVS as depicted in \fig \ref{fig:mvs}. Here, the joint approach of \pulse is tested by benchmarking a joint model against the mixture of its constituent models.

\subsubsection{Experiments}
The LTMs P* and I* were combined using the product and mean rule, and the same approach was used for setting parameter $b$ as above. Then, the results were compared to the PI* LTM, the best joint model that uses pitch and interval features.

\subsubsection{Results}
I expected that a well crafted joint model would outperform an ensemble of its source models, as the joint model was tuned to maximize the joint performance whereas the source models were tuned to maximize each source's performance.
It is striking and counter to my expectations that the LTM ensemble performed better than the joint model PI* (2.302 bits). 

This finding allows me to draw the conclusion that (1) the best reported performance may be outperformed by an ensemble of smaller \pulse models with the same features, or (2) \pypulse is not operating in its optimum yet, as the joint model should be able to learn a superset of statistical patterns compared to the sub-models. Considering the number of features of each model, we see that PI* converged to only 633 features (average over the CV folds), compared to 791 of I* and 1,236 of P*. It seems that the worse performance of the joint model might be caused by the feature culling of \lone regularization and the \textsc{shrink} operator. I leave it up to future work, to analyze whether weaker \lone regularization closes the gap between the mixture and joint model.


\chapter{Evaluation}\label{chap:evaluation}
In this chapter, I assess the \pypulse learning algorithm for music that was presented and tuned in the previous chapters. The assessment is performed in two steps. Firstly, the best \pypulse models are compared with \art results in monophonic melody prediction and cognitive modeling. Secondly, the learned feature sets and weightings are analyzed musicologically and structurally.

\section{Literature Comparison}
Results gain real significance only when considered in relation to other work. This section chiefly reports the results of a quantitative comparison with \art models.

Section~\ref{sec:comparison-sota} compares the best \pypulse LTM, STM and hybrid models to the state-of-the-art. Ensemble models (\ngram MVS) and joint \pypulse models using corresponding feature types are compared in section~\ref{sec:comparison-mvs}. In section~\ref{sec:psycho}, \pypulse's suitability as cognitive model of expectation is evaluated. Please refer to section~\ref{sec:methodology} for an introduction of the utilized corpus, the evaluation measure, Gaussian process (GP) based hyperparameter optimization and cross-validation (CV).

\subsection{Comparison with State-of-the-Art Methods}\label{sec:comparison-sota}
The best \pypulse performances for LTM, STM and hybrid models were compared to the \art models for melody prediction. All three \pulse models outperformed the \art models significantly.

\subsubsection{Experiments}
The best \pypulse models were evaluated on the eight datasets of the Pearce corpus. The models were the PI*C*KM$_K$-LTM (see section~\ref{sec:ltm-configs}), the PI*F$_1$-STM (see section~\ref{sec:stm-configs}) and the hybrid of the PI*C*KM$_K$-LTM and (C*I) $n$-gram STM (see section~\ref{sec:hybrid-configs}). The corpus cross-entropy served as the performance measure. It is computed as the average entropy over the eight dataset entropies.

The hyperparameters were determined as follows: In the LTM, the global \lone regularization factor $\lambda_{1}$ was determined per fold on a small held out part (10\%) of the training set, using GP based optimization. Parameter $\lambda_{1}$ was the only one tuned on a per-dataset (and fold) basis. All remaining hyperparameters were tuned on the Bach chorales dataset as described in section~\ref{sec:results-optimization} and \ref{sec:results-perfeature-ltm}. In the STM it was planned to tune $\lambda_1^\textit{init}$, $\lambda_2^\textit{init}$ and $\tau_2$ on a per-dataset basis which turned out to be computationally infeasible. Only for the German nursery rhymes dataset, the hyperparameters were tuned on the actual dataset. For the other datasets fixed values $\lambda_1^\textit{init}=10^{-5}$, $\lambda_2^\textit{init}=0.01$ and $\tau_2=8$ were assumed based on preliminary runs. The remaining hyperparameters were tuned on the German nursery rhymes dataset as described in section~\ref{sec:results-perfeature-stm}.

\subsubsection{Results}
We compare the \art performances with the \pulse LTM, STM and hybrid performances in \tab \ref{tab:sota}. The \pulse LTM and hybrid model improve the \art by a leap larger than that between the RTDRBM \parencite{cherla2015discriminative,cherla_hybrid_2015} and \ngram \parencite{pearce2004} models. The \pulse LTM performed 0.17 bits, the \pulse hybrid 0.064 bits better than the record holder RTDRBM. The \ngram STM has been surpassed for the first time; the \pulse STM performed 0.047 bits better.

\begin{table}[!h]
	\newcommand*{\none}{--}
	\centering
	\begin{tabular}{l|cccc}
		\toprule
		& LTM &LTM$^+$ & STM & Hybrid \\
		\midrule
		\pulse & \textbf{2.542}&\none & \textbf{3.092} &  \textbf{2.357}\\
		\midrule
		RTDRBM & 2.712&2.756 & 3.363 & 2.421 \\
		RBM & 2.799 & \none & \none & \none \\
		FNN & 2.830& \none & \none & \none \\
		$n$-gram& 2.878&2.614 & 3.139 & 2.479 \\
		\bottomrule
	\end{tabular}
	\caption{Benchmark of the best melody prediction models \textit{P\textsmaller{ULSE}}, RTDRBM \parencite{cherla_hybrid_2015}, RBM \parencite{cherla2013RBM}, FNN \parencite{cherla2014multiple} and \ngrams \parencite{pearce2004}.}
	\label{tab:sota}
\end{table}

All \pypulse LTM and STM configurations that have been benchmarked on the entire Pearce corpus are reported in \tab \ref{tab:all-benchmarks}. Note that the comparison in \tab \ref{tab:sota} compares single (or joint) model performances with each other. A comparison of \pypulse with the best performing ensemble methods is undergone in the next section.
\begin{table}[htb]
	\centering
	\scalebox{0.86}{
		\begin{tabularx}{1.14\textwidth}{c | *5{>{\centering}X} | ccc}
			\toprule
			& \multicolumn{5}{c|}{LTM} & \multicolumn{3}{c}{STM}\\
			Dataset&P*&PI*C*&PI*C*K&PI*C*M$_K$&PI*C*KM$_K$&P*&PI*C&PI*F$_1$\\
			\midrule
			0          &2.792&2.685&2.493&\textbf{2.489}&2.490&3.126&3.118&\textbf{3.041}\\
			1          &2.393&2.295&2.209&2.188&\textbf{2.187}&3.228&3.106&\textbf{3.047}\\
			2          &3.206&2.889&2.664&2.672&\textbf{2.659}&3.177&3.074&\textbf{2.996}\\
			3          &2.770&2.605&2.476&2.498&\textbf{2.489}&3.595&3.570&\textbf{3.396}\\
			4          &3.149&2.870&2.691&\textbf{2.696}&2.715&3.219&3.183&\textbf{3.087}\\
			5          &3.446&3.182&2.956&\textbf{2.951}&2.952&3.342&3.305&\textbf{3.218}\\
			6          &2.402&2.331&2.234&\textbf{2.202}&2.203&2.737&2.684&\textbf{2.590}\\	
			7          &2.920&2.770&2.656&2.653&\textbf{2.646}&3.567&3.442&\textbf{3.363}\\
			\midrule
			Average&2.885&2.703&2.547&2.544&\textbf{2.542}&3.249&3.185&\textbf{3.092}\\
			\bottomrule
	\end{tabularx}}
	\caption{\pulse benchmarks on the Pearce corpus.}
	\label{tab:all-benchmarks}
\end{table}

\subsection{Comparison with \textit{n}-gram MVS}\label{sec:comparison-mvs}
In this section we compare \textit{PyPulse} models to \ngram models with matching feature and viewpoint sets. The \ngram models under investigation are:
\begin{enumerate}
	\item[(1)] Single model \ngrams using linked viewpoints (see section~\ref{sec:MVS}), the direct \ngram equivalents to \pypulse models.
	\item[(2)] The best reported melody prediction ensembles, namely \ngram MVS \parencite{pearce2005}, using the best values from literature and IDyOM.
\end{enumerate}
While \pulse is clearly the winner in (1), the comparison (2) provides mixed results. I give arguments that \pulse models have the potential to outperform either.

\subsubsection{Experiments}
\pypulse models were compared to ensemble and linked viewpoint \ngram MVS that used the same viewpoints. Specifically, a \pypulse P*-LTM, STM and LTM+STM hybrid was compared to the best viewpoint pitch \ngram models; and a \pypulse PI*C*-LTM and STM were compared to the best pitch-interval-contour \ngram ensemble, as well as linked viewpoint models. The best \ngram performances were taken from \textcite{pearce2004} or generated using the IDyOM framework\footnote{\url{https://code.soundsoftware.ac.uk/projects/idyom-project}}.

\subsubsection{Results}
All results are given in \tab \ref{tab:art-mvs}. From \tab \ref{tab:art-p} we learn that \ngrams with the single viewpoint pitch outperform the equivalent \pulse LTM, STM and hybrid. However, if we restrict ourselves to comparing the datasets that the majority of hyperparameters was tuned on, then \pulse is leading in the LTM case (dataset 1) and both models perform similarly in the STM case (dataset 6). I assume that the \ngrams can be outperformed with a dataset specific hyperparameter tuning for the LTM, and that both models will be tied in the STM. We can also interpret this experiment to be a comparison of generalized \ngrams to $n$-grams; the same results apply here.

The results for the comparison of pitch-interval-contour viewpoint with PI*C* feature models as shown in \tab \ref{tab:art-pic} are intriguing: If we compare \pulse to its direct single model \ngram equivalent, the pitch-interval-contour linked viewpoint models, then \pulse is winning with a large margin. In fact, the linked \ngram model deteriorated to the pitch-alone version. This emphasizes the capabilities of the \pulse method to join several viewpoints within a single model.

In contrast to that, we observe that \pulse loses the comparison with the \ngram ensemble models. However, if we again consider only the results of dataset~1 for the LTM and dataset 6 for the STM, then \pulse outperforms the ensemble methods. I assume that single \pulse models have the potential to beat LTM and STM ensembles if the hyperparameter tuning is improved. It is left to future work to validate my hypothesis that unified \pulse models can outperform ensembles of \ngrams with the corresponding feature and viewpoint sets.

\begin{table}[htb!] 
	\centering
	\subfloat[The best models based on P viewpoints/features only.]{
		\scalebox{0.88}{
			\begin{tabular}{c | cc | cc | cc}
				\toprule
				& \multicolumn{2}{c|}{LTM} & \multicolumn{2}{c|}{STM}  & \multicolumn{2}{c}{LTM+STM hybrid}\\
				Dataset& \pulse & $n$-gram$^{i, c}$& \pulse & $n$-gram$^{p,x}$ & \pulse & $n$-gram$^{p,c}$\\
				\midrule
				0          &\textbf{2.792}&2.863&3.126&\textbf{2.977}&2.509&\textbf{2.468}\\
				1          &\textbf{2.384}&2.443&3.228&\textbf{3.117}&2.368&\textbf{2.34}7\\
				2          &3.206&\textbf{3.089}&3.177&\textbf{3.090}&2.632&\textbf{2.540}\\
				3          &2.770&\textbf{2.720}&3.595&\textbf{3.411}&2.691&\textbf{2.588}\\
				4          &3.149&\textbf{2.982}&3.219&\textbf{3.137}&2.628&\textbf{2.454}\\
				5          &3.446&\textbf{3.307}&3.342&\textbf{3.244}&2.804&\textbf{2.651}\\
				6          &\textbf{2.402}&2.427&2.737&\textbf{2.731}&2.114&\textbf{2.106}\\
				7          &\textbf{2.920}&3.097&3.567&\textbf{3.406}&2.706&\textbf{2.681}\\
				\midrule
				Average&2.885&\textbf{2.866}&3.249&\textbf{3.139}&2.557&\textbf{2.479}\\
				\bottomrule
		\end{tabular}}
		\label{tab:art-p}}
	\bigskip
	
	\subfloat[The best models based on P, I and C viewpoints/features. The \ngram performance is computed once for an ensembles of viewpoints and once for a single linked viewpoint.]{
		\scalebox{0.88}{
			\begin{tabular}{c | ccc | ccc }
				\toprule
				& \multicolumn{3}{c|}{LTM} & \multicolumn{3}{c}{STM} \\
				Dataset& ~\pulse ~& ~$n$-gram$^{i,c,e}$& $n$-gram$^{i,c,l}$& ~\pulse~ & ~$n$-gram$^{i,c,e}$ & $n$-gram$^{i,c,l}$\\
				\midrule
				0          &\textbf{2.685}&2.724&2.878&3.118&\textbf{3.032}&3.377\\
				1          &\textbf{2.290}&2.358&2.458&3.106&\textbf{3.053 }&3.418\\
				2          &2.889&\textbf{2.847}&3.105&\textbf{3.074}&3.086 &3.570\\
				3          &2.605&\textbf{2.580}&2.745&3.570&\textbf{3.529} &3.924\\
				4          &2.870&\textbf{2.687}&2.995&3.183&\textbf{3.140} &3.602\\
				5          &3.182&\textbf{3.053}&3.358&3.305&\textbf{3.223}&3.735\\
				6          &2.331&\textbf{2.300}&2.450&\textbf{2.684}&2.703&3.128\\
				7          &\textbf{2.770}&2.892&3.129&3.442&\textbf{3.373}&3.950\\
				\midrule
				Average&2.703&\textbf{2.680}&2.890&3.185&\textbf{3.142}&3.588\\
				\bottomrule
		\end{tabular}}
		\label{tab:art-pic}}
	\caption{\pulse with viewpoint feature configurations compared to \ngram models with matching viewpoints.\\\rule{10.2cm}{0.4pt}\\
		{\footnotesize $^i$  Generated using IDyOM.\\$^c$ Escape method C and interpolated smoothing (C*I).\\$^x$ Escape method X, update exclusion, and interpolated smoothing (X*UI).\\$^p$ \textcite{pearce2004}.\\$^l$  Linked viewpoint cpitch-cpint-contour.\\$^e$  Ensemble of viewpoints cpitch, cpint, and contour.}}
	\label{tab:art-mvs}
\end{table}

\subsection{Comparison with Psychological Data}\label{sec:psycho}
In the past sections, I have thoroughly assessed \textit{PyPulse}'s quantitative performance as a predictive model of melody. Here, I make an attempt to confirm its suitability as cognitive model of pitch expectation. For that we use \pypulse to compute an entropy profile and compare it with \textcite{pearce2006expectation}'s results to the simulation of human predictive uncertainty. Entropy profiles are the entropies (i.e. the negative base-2 logarithms of the likelihoods) ascribed to the respective next notes in a song. Pearce and Wiggins used an \ngram MVS, highly tuned to the task, to simulate an entropy profile obtained from an experimental study by \textcite{manzara1992entropy}. In the study, a betting paradigm was used that, given a context, asked participants to distribute their bets on the most likely continuations.

The results confirm a general suitability of \pypulse to the task of cognitive modeling, although the achieved performance was worse than prior work using \ngram MVS. However, the reported results are of a preliminary nature and should be considered to be  a proof of concept.

\subsubsection{Experiments}
The best performing \pypulse LTM from section~\ref{sec:ltm-configs} (PI*C*KM$_K$) was trained on the Bach chorales dataset minus chorale 126 (BWV 379). The resulting entropy profile for test chorale 126 was compared to the behavioral entropy profile of \textcite{manzara1992entropy} and the \ngram MVS profile, as they were reported by \textcite{pearce2006expectation}. The goodness of fit was measured by the proximity of the model and the human profile.

\subsubsection{Results}
\begin{figure}[htb]
	\centering
	\includegraphics[width=\textwidth]{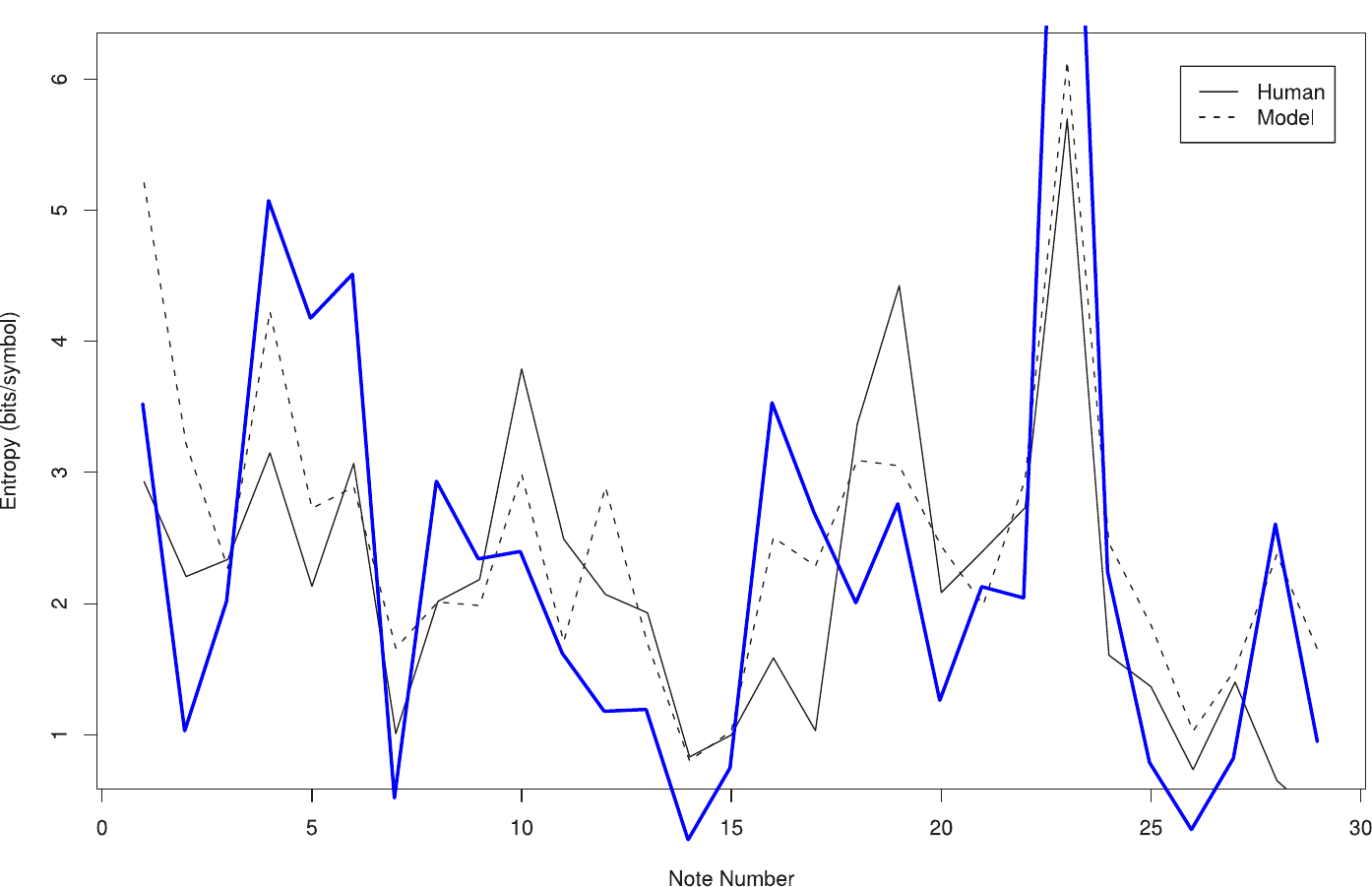}
	\caption{Entropy profile of a \pulse LTM (blue) plotted on top of \fig 7 from \textcite{pearce2006expectation} which shows the entropy profiles of human pitch expectation in comparison to those of an \ngram MVS model for the chorale \textit{Meinen Jesum laß' ich nicht, Jesu} (BWV 379).}
	\label{fig:bwv379pearce}
\end{figure}

\Fig \ref{fig:bwv379pearce} shows the entropy profiles of Manzara et al. and Pearce et al., superimposed by the \pulse profile in blue. From the curves it can be confirmed that the human and \pulse entropy profiles are positively correlated. The profile contours of both, the \ngram and \pulse model, deviate from the human contour in five notes ($n$-gram: note 9, 12, 19, 21, 28; \textit{P\textsmaller{ULSE}}: note 9,13, 18, 22, 28). The \ngram entropy values are closer to the human values for the majority of notes.

I consider this experiment to be a proof of concept and to be of a preliminary nature. The evaluation on a single benchmarking piece is of limited expressiveness and cannot be considered to generalize well. Thus, a diligent quantitative analysis was omitted. Moreover, the three entropy profiles were obtained based on different preconditions. The training corpus for humans is unknown, but presumably was very large. \textit{P\textsmaller{ULSE}} was trained on only 184 Bach chorales, and the $n$-gram models were trained on a corpus of 152 Canadian folk songs, 184 Bach chorales, and 566 German folk songs. The number of prediction classes differed as well: 20 for humans, 21 for \textit{P\textsmaller{ULSE}}, and is of unknown size for the $n$-grams. Last but not least, the \pulse model was not specifically tuned to gain the highest performance on this very test piece. On the contrary, the viewpoints of the \ngram MVS were selected to fit the entropy profiles optimally. Such a model and feature selection for \pypulse was left for future work.

Generally, the best performing model entropy-wise is not necessarily the best cognitive model of auditory expectation. \textcite{pearce2012auditory} remark that the statistical models' memories never fail. \textcite{rohrmeier2012predictive} similarly criticize that \ngram models outperform the learning skills of humans and thus provide an inaccurate surrogate.

\section{Analysis of the Learned Models}
A black box learner might bristle with performance, but it does not provide any further insights. One chief advantage of the \pulse method is that it affords the extraction of valuable music theoretic insights from the learned models. After the training one can observe how the data was memorized and, furthermore, if the algorithm was primed with the right feature building blocks, one can gain new understanding of the underlying data.

In this section, several trained models are analyzed by submitting the discovered feature sets to a musicological analysis (section~\ref{sec:feature-analysis}), and to a technical analysis with a focus on their temporal composition (section~\ref{sec:temporal-analysis}). Last but not least, the models are brought to life through the generation of sequences (section~\ref{sec:results-inference}).

\subsection{Musicological Analysis}\label{sec:feature-analysis}
The musicological analysis covers (1) the estimation of each feature type's contribution to the model predictions, (2) the music theoretic interpretation of the weightings of zero-order features, (3) the extraction of the motifs with heaviest weights, and (4) the correlation between the metrical weight and tonic triad.

In the following PI*C*K-LTM are used, trained on the Bach chorale and Chinese folk tunes datasets. These datasets were chosen due to their cultural and regional distinction to elicit multifaceted models. The regularization factor $\lambda_{1}$ was set to the over the CV folds averaged \lone factor from the \nplus benchmark in section~\ref{sec:ltm-configs}.
 
\subsubsection{Relevance of Feature Types}
The used models were given four different types of features to learn the data: pitch, interval, contour, and key features. To further understand each type's relevance, it is important to know how much weight~--~figuratively as well as mathematically~--~is given to each type. \Tab \ref{tab:analysis-weights} shows the accumulated weights given to all features of each type, in percent of all assigned weights to any type. This metric presumes that a feature type with a higher weight has a higher influence on the prediction. We observe that interval features are assigned more than 70\% of all weights in either model. It is surprising to see that K features have a higher importance for Chinese folk tunes than the Bach chorales, considering that keys are a western concept.
\begin{table}[htb]
	\centering
	\begin{tabular}{c | cc}
		\toprule
		Feature Type&Bach Chorales& Chinese Folk Tunes\\
		\midrule
		P~~~&0.084&0.125\\
		I*~~&0.744&0.702\\ 
		C*&0.056&0.042\\
		K~~&0.116&0.131\\
		\midrule
		Total \#Features&312&399\\
		\bottomrule
	\end{tabular}
\caption{The weight allocation to each feature type in percent of the total allotted weights.}
\label{tab:analysis-weights}
\end{table}

\subsubsection{Weight Analysis for Zero-Order Features}
\begin{figure}[!htb]
	\centering
	\subfloat[][Chinese folk melodies dataset ]{\raisebox{0.04cm}{\includegraphics[width=.48\linewidth]{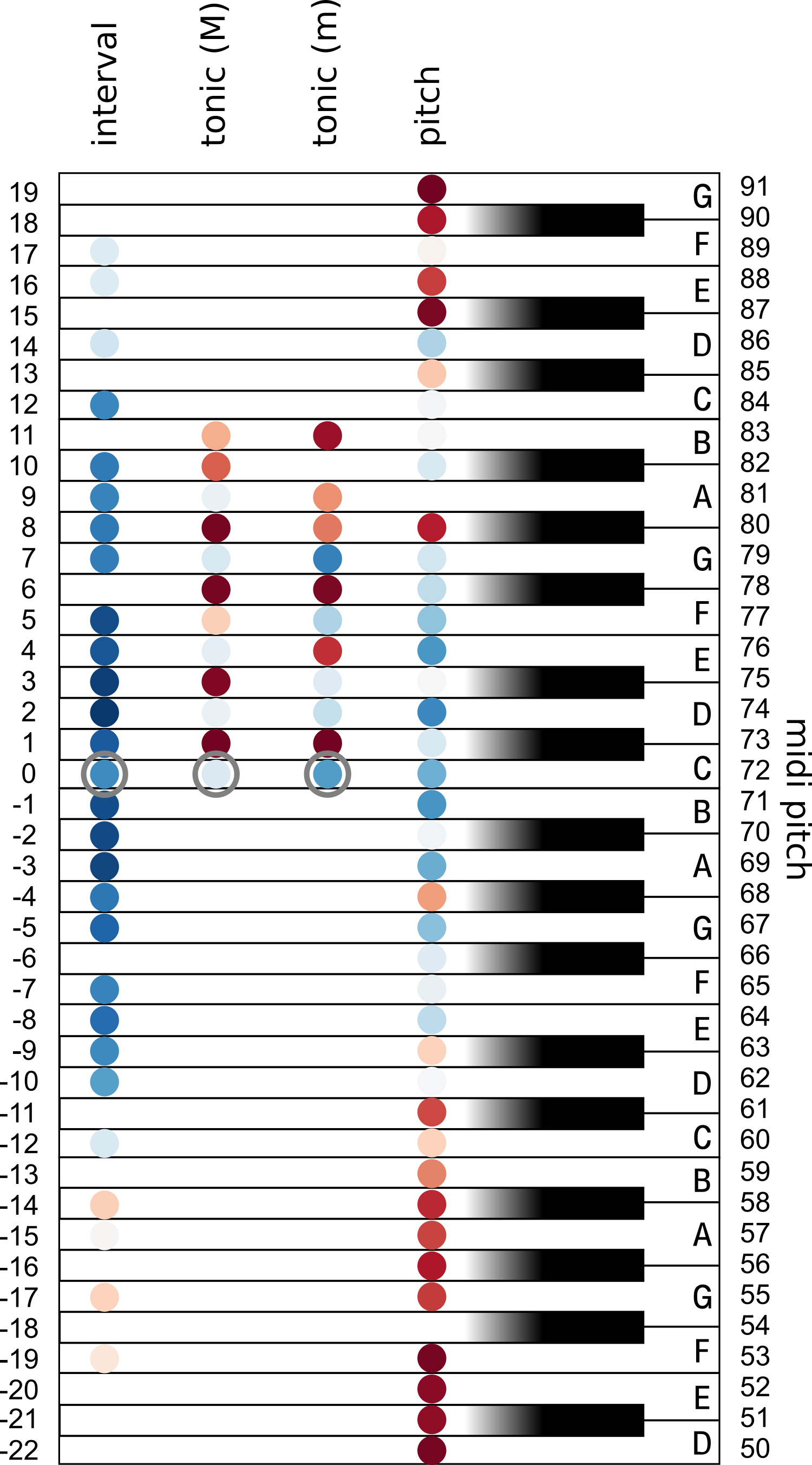}\label{fig:piano_ko_chinese}}}
	\quad
	\subfloat[][Bach chorales dataset]{
		\includegraphics[width=.48\linewidth]{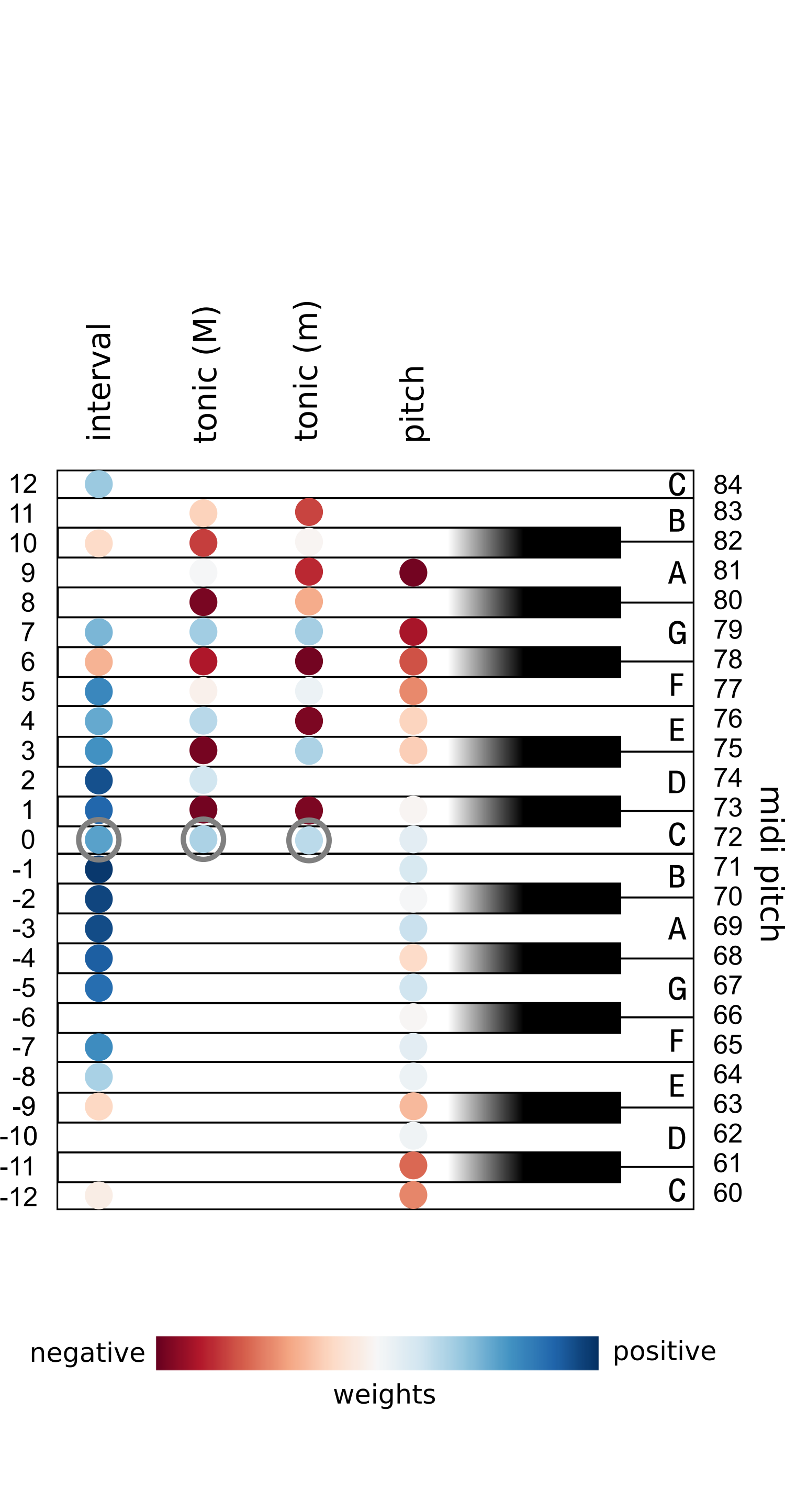}}\label{fig:piano_ko_chorales}
	\caption{Qualitative plot of the feature weights for length-one P and I viewpoint features as well as K anchored features of the PI*C*K-LTM . For the values of I and K features, C5 was chosen as reference tone. The color scale is normalized for every feature type to make use of the maximum range (red to blue, lowest negative to highest positive weight). Positive weights ascribe a high likelihood to the corresponding musical event, negative weights attribute low likelihood.}
	\label{fig:piano_features}
\end{figure}

Zero-order or length-one viewpoint features carry a special status, as they model the occurrence frequencies of the respective viewpoint values. \Fig \ref{fig:piano_features} plots the heat encoded weights for each length-one feature value in the model. Note that gaps in the plots are features that did not occur in the dataset or were regularized away.

Let us consider the weights of the pitch features first. We can infer the regularity that in both, Chinese and Western styles, the center of the register is preferred while exceptionally high or low pitches are discouraged. We further observe that the range of pitches is much larger for Chinese tunes then for Bach chorales.

Looking at the interval features' weights, we see that in both models smaller intervals are preferred over larger ones, which is in accordance with general principles of voice leading \parencite{huron2001tone}. For the Chinese folk tunes model, the discovered interval range is distinctly larger. In the Chinese model, large ascending steps are preferred over large descending steps, whereas there is no distinct predilection in the direction of small intervals. In contrast to that, small intervals are encouraged to descend in the Chorales model. That is consistent with \textcite{vos1989ascending} that reported small intervals  in western music to have the tendency to go down. We further observe that the Bach chorale model discourages the tritone step ($\pm$6 semitones) which was seldomly used to express anger or sadness in Western music, and did not even occur in the Chinese melodies in the first place. Unisons (0) are less attractive than small steps in both models.

It is striking to see that key profiles have been learned by the features of type K. The tonic features for major (M) and minor (m) represent interval frequencies relative to a key. In consequence, the model can use song specific absolute pitch frequencies. We can observe almost identical weightings in both models. The tones of the major and minor triad are preferred; those outside the respective diatonic scale are discouraged, as it was typical during Bach's times. The minor (m)/major (M) triad are tonic (0 semitones), minor third (3 semitones)/major third (4 semitones) and fifth (7 semitones).

\subsubsection{Exemplary Motifs}
Motifs are contiguous transposition-invariant melody snippets (i.e. $n$-gram). As learned from \tab \ref{tab:analysis-weights}, features generated by \nplus operator I* are of particular importance for the prediction. This section can merely give a taste of the motifs learned, due to the large number of features. Hence, \tab \ref{tab:analysis-motifs} shows only the motif of highest weight for lengths three to five, respectively (feature length two to four). 

\begin{table}[htb]
	\centering
	\begin{tabular}{l|c|c|c}
		\toprule
		Dataset & \multicolumn{3}{c}{Motifs}\\
		\midrule
		\raisebox{0.44cm}{\parbox{1.4cm}{\small Bach\\ Chorales}}&\includegraphics[height=1cm]{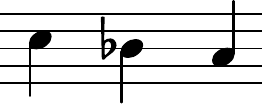}&\raisebox{0.05cm}{\includegraphics[height=1.08cm]{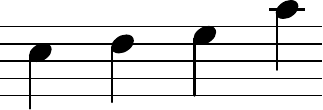}}&\raisebox{0.05cm}{\includegraphics[height=0.82cm]{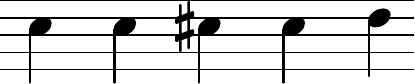}}\\
		\midrule
		\raisebox{0.44cm}{\parbox{1.7cm}{\small Chinese\\Folk Tunes}}&\includegraphics[height=1cm]{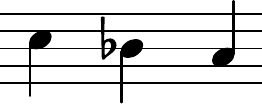}&\includegraphics[height=0.95cm]{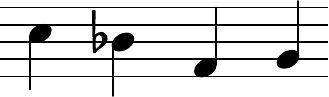}&\includegraphics[height=0.88cm]{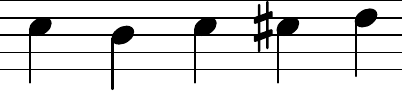}\\
		\bottomrule
	\end{tabular}
	\caption{The type I* contiguous compound features of highest weight for feature length two to four, instantiated with start pitch C5.}
	\label{tab:analysis-motifs}
\end{table}

At first sight, the listed motifs do not appear to be the likeliest ones in the respective genres. However, one should keep in mind that the motifs are CRF features and not $n$-grams. Given a context $x$ and an outcome $y$, each model prediction for $y$ is assembled from the set of features that evaluate to true for $(x,y)$. In a motif, $x$ are all intervals but the last one, and $y$ is the last interval. In consequence, a heavy weighted feature that matches $x$ puts in its weight to promote $y$ to be the next interval, amongst all other features that match $x$ and promote the same or different outcomes. Bearing that in mind, the stepwise continuation that five out of six features advocate seems sensible.

Furthermore, we can discover two simple claims of \textcite{narmour1992analysis}'s Implication-Realization model: In the second motif for the Chinese folk tunes, a large interval implies a change of direction. In the first and third interval of both models, small intervals imply the continuation of direction.

\subsubsection{Correlations Between Metrical Weight and Tonic Triad}\label{sec:feature-analysis-mk}
We will now assess the linked features of type M$_{K}$, specifically the interplay of metrical structure and key. For that we employ a PI*C*KM$_{K}$-LTM learned on the Bach chorales dataset.

It is pertinent to ask: What is the advantage of M$_K$ compared to K? From a mathematical point of view the advantage stems from a higher spatial precision. In addition to K, M$_K$ also represents where in the metrical structure the chromatic scale degrees of the reference key lie. From a music theory point of view these features are motivated by findings that there is a predilection for tones of the tonic triad to lie on heavier counts \parencite{caplin1983tonal}.

\begin{figure}[htb]
	\centering
	\includegraphics[width=\linewidth]{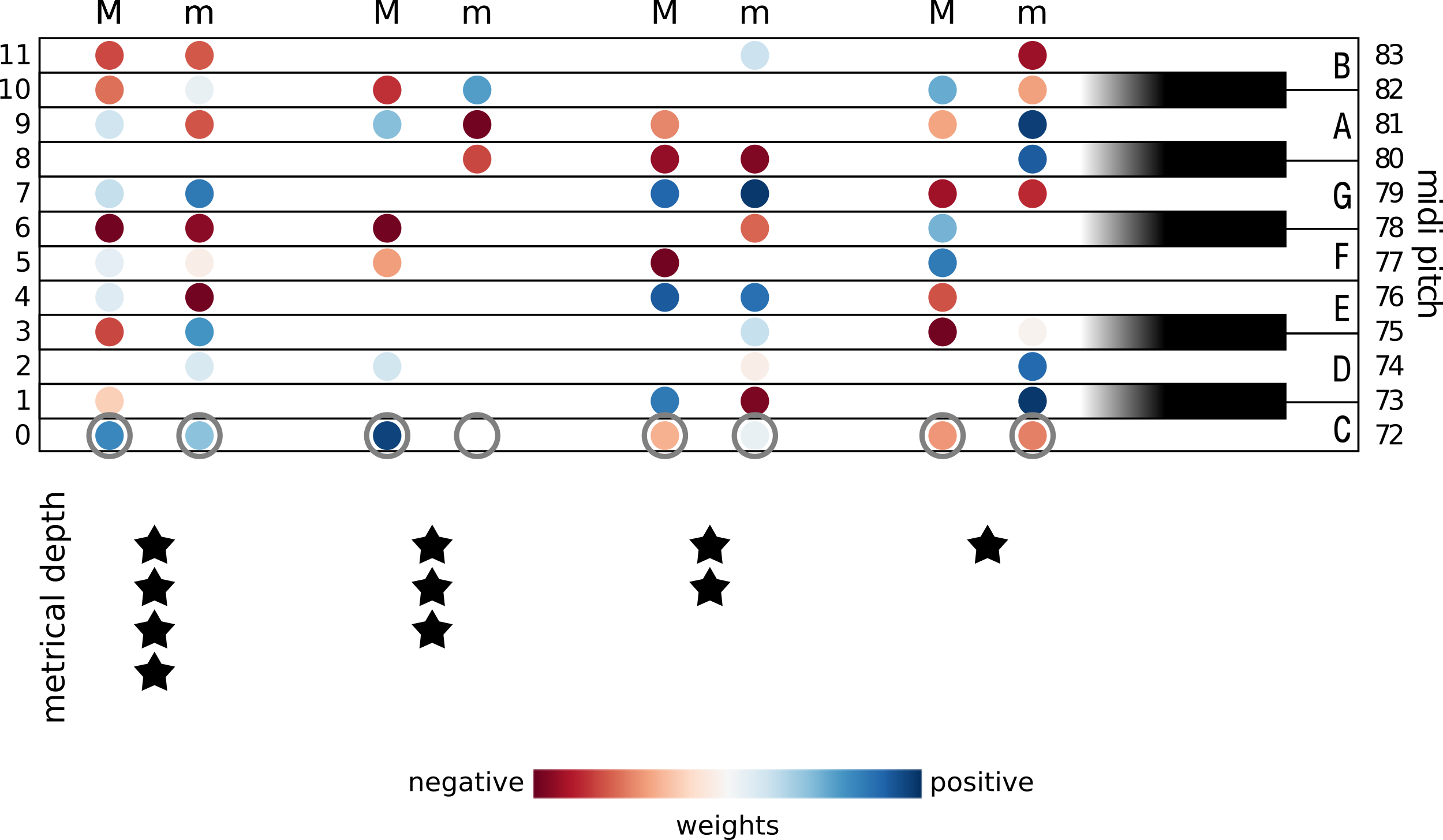}
	\caption{The weight distribution as learned for the linked metrical weight and key features M$_K$. The weights of the intervals relative to the major (M) and minor (m) tonics are given for every metrical weight, if the feature was selected.}
	\label{fig:mdepth_key_chorales}
\end{figure}

We make the following relevant observation from the M$_K$ features for the different metrical weights (see \fig \ref{fig:mdepth_key_chorales}):
\begin{itemize}[align=right,itemindent=0cm,leftmargin=2cm]
	\item[\FourStar:] The tones of the major and minor triad are encouraged.
	\item[\ThreeStar:] No particular prevalence can be deduced. Furthermore several M$_K$ features did not occur or were regularized away which indicates a neutrality towards the occurrence of scale degrees for this metrical weight.
	\item[\TwoStar:] All tones of the triads are encouraged with the exception of the tonics. In particular the dominant (fifth diatonic scale degree) which in music theory is said to be of lighter metrical weight than the tonic.
	\item[\OneStar:] In striking contrast to the stronger beats in the metrical structure, this one firmly disfavors the tones of the major and minor triad.
\end{itemize}

In conclusion, we saw that in several instances in-scale tones were encouraged to lie on the heavier metrical weights of depth four and two, and especially, that the weakest beats in the structure strongly discourage all in-scale tones.

\subsection{Temporal Model Analysis}\label{sec:temporal-analysis}
For any system that is designed to predict the continuation of a sequence, it is intriguing to know how much of the available context the system uses in practice. This section will answer this question, examine the temporal weight distribution, and analyze the extent of use of generalized \ngram features.

Conceptually, \pulse can built features of arbitrary depth, but the \nplus operations that we use limit the feature size by the number of outer loop iterations. The constructed features are generalized \ngrams which are compounds of viewpoint features (see section~\ref{sec:pypulse-viewpoint}). A generalized \ngram feature differs from an \ngram feature by not being contiguous, meaning that it has gaps or holes.

\begin{figure}[!p]
\vspace{-0.7cm}
\centering
\subfloat[PI*C*K]{
\includegraphics[height=3.7cm]{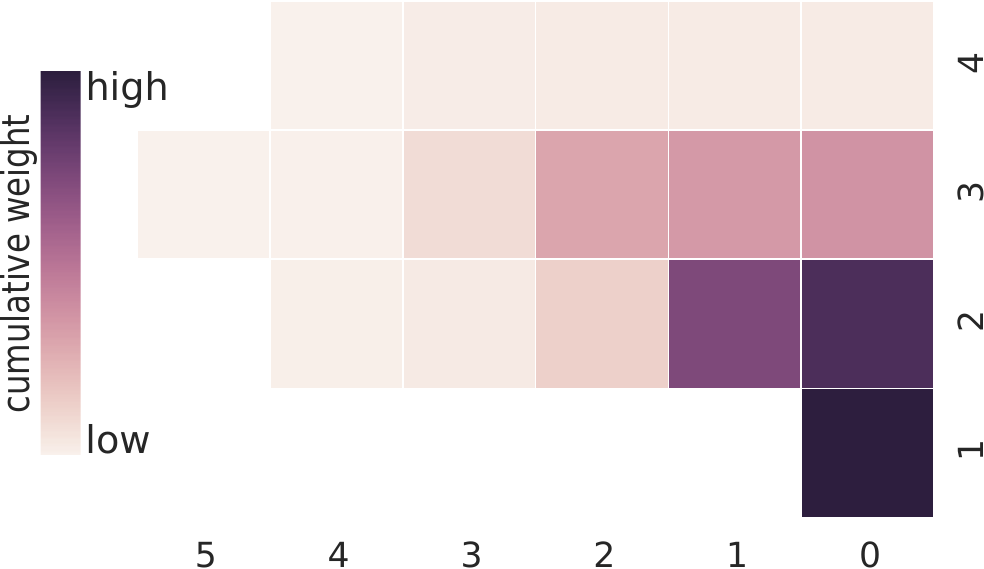}}\quad
\subfloat[PI*C*KM$_K$]{
\includegraphics[height=3.7cm]{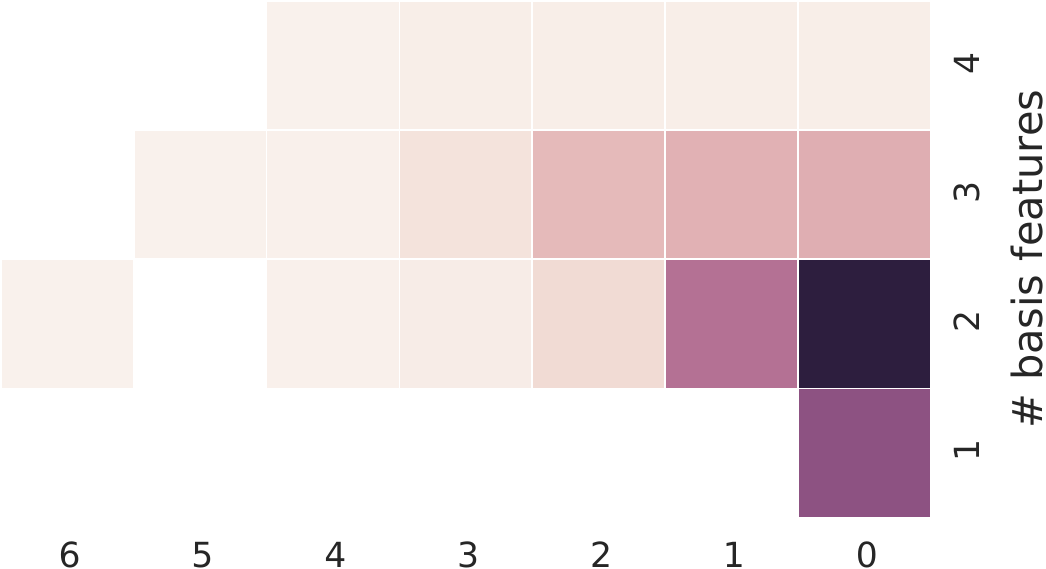}}\\
\subfloat[P*]{
\includegraphics[height=5.05cm]{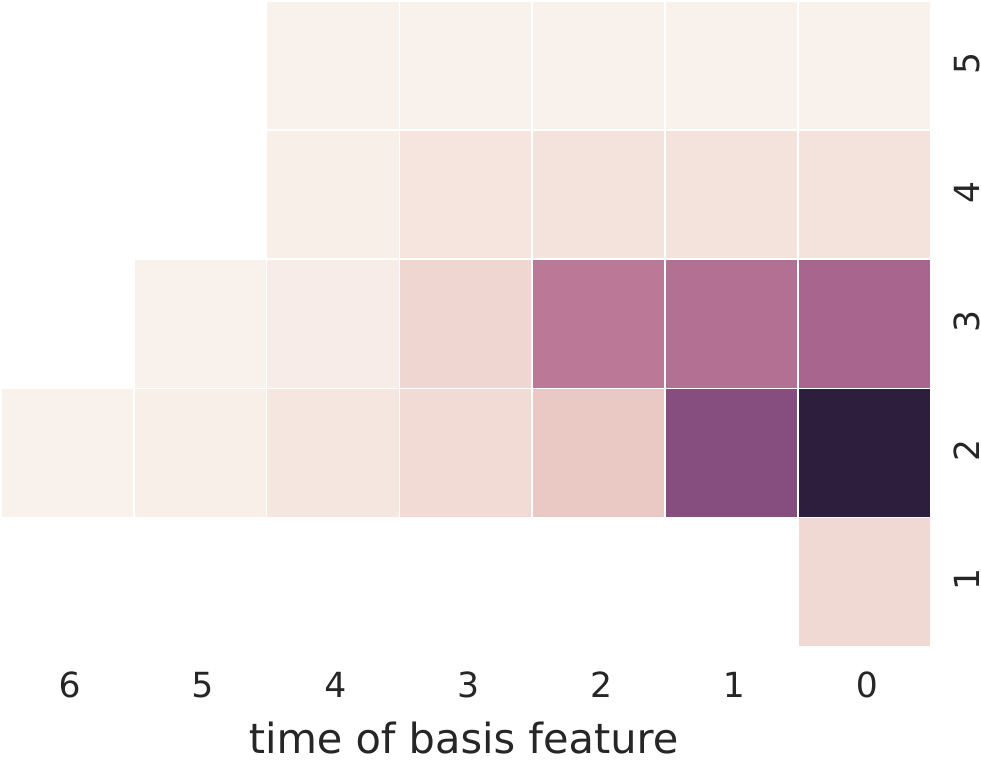}}\quad
\subfloat[PI*C*]{
\includegraphics[height=5.05cm]{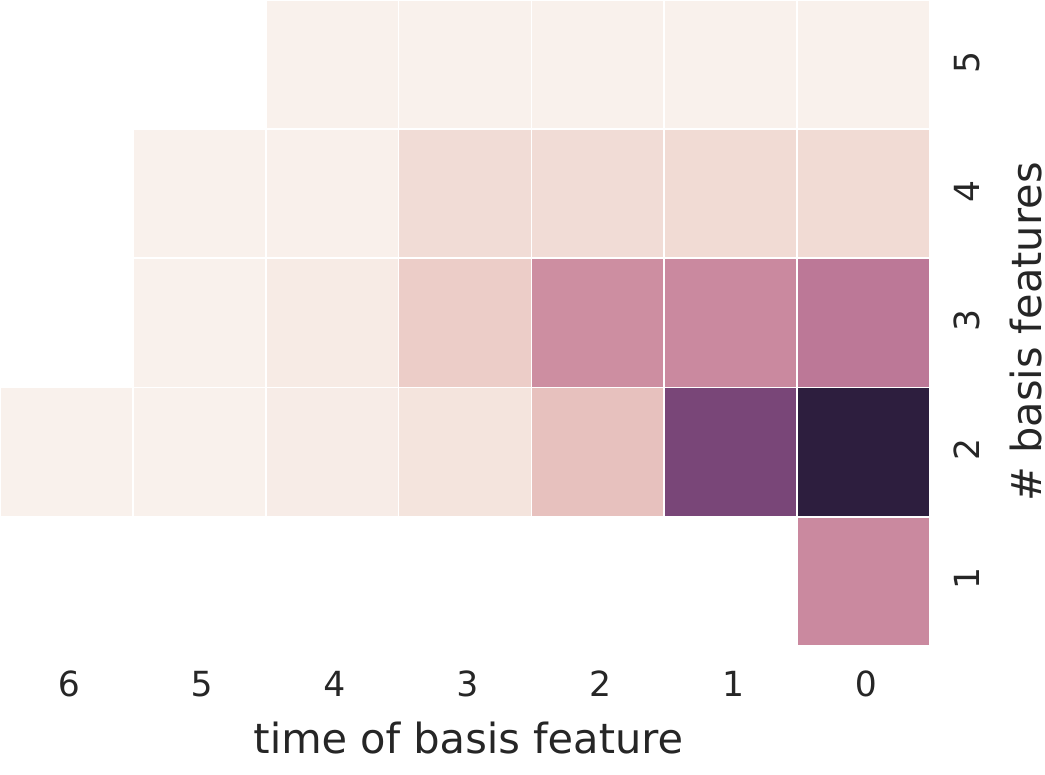}}
\caption{\small This plot shows the feature weight distribution over features of different length and temporal extent. Every field contains the sum of the absolute weights of all basis features $f_{\sigma,\nu}$ that have the respective temporal depth $\sigma$ and occur in a compound of the respective number of basis features. The models were learned on the Bach chorales dataset.}
\label{fig:heatplots-featuredepth}
\vspace{0.7cm}
	\subfloat[PI*C*K]{
		\hspace{0.0cm}\includegraphics[height=3.7cm]{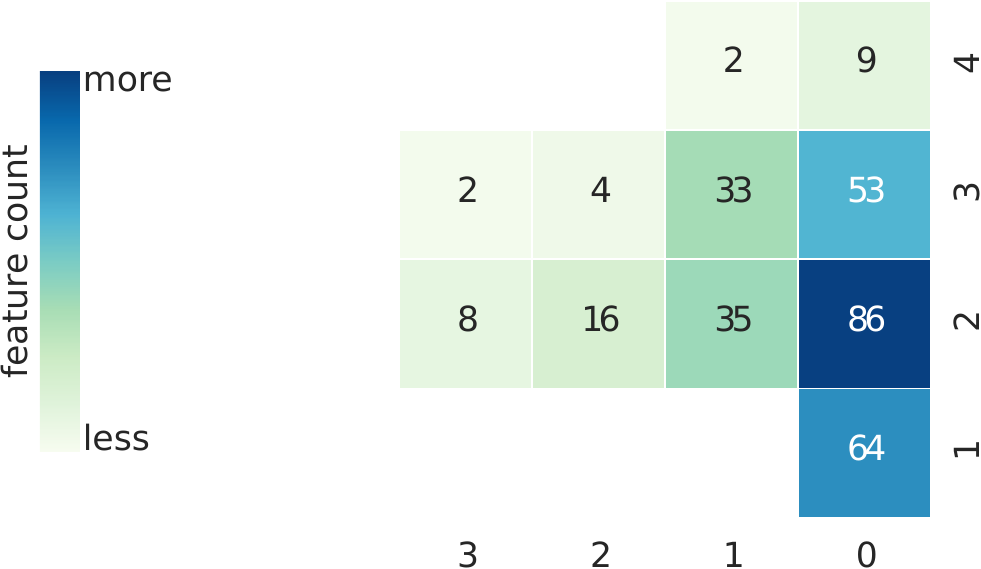}}\quad
	\subfloat[PI*C*KM$_K$]{
		\includegraphics[height=3.7cm]{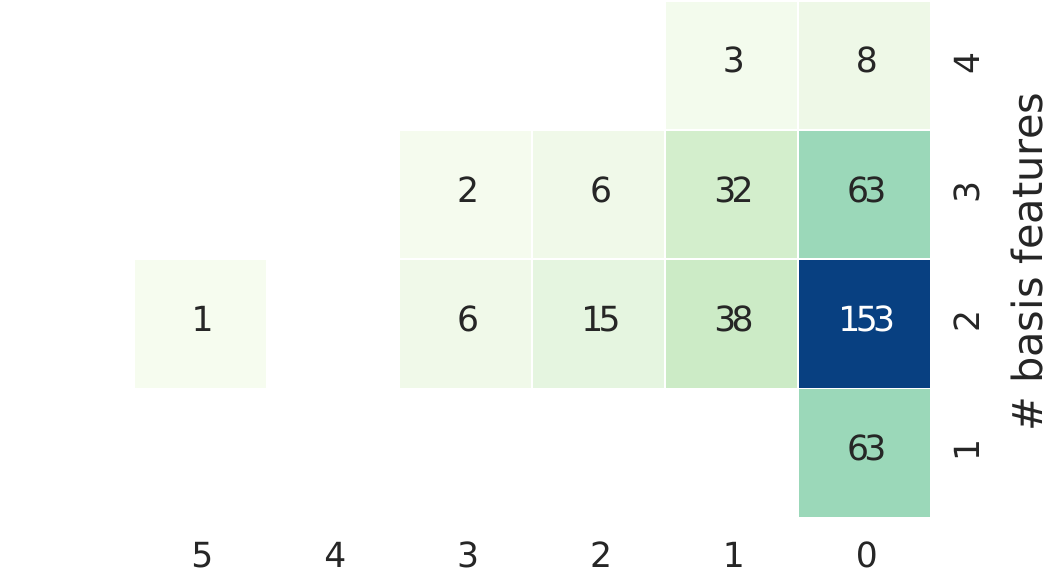}}\\
	\subfloat[P*]{
		\includegraphics[height=4.91cm]{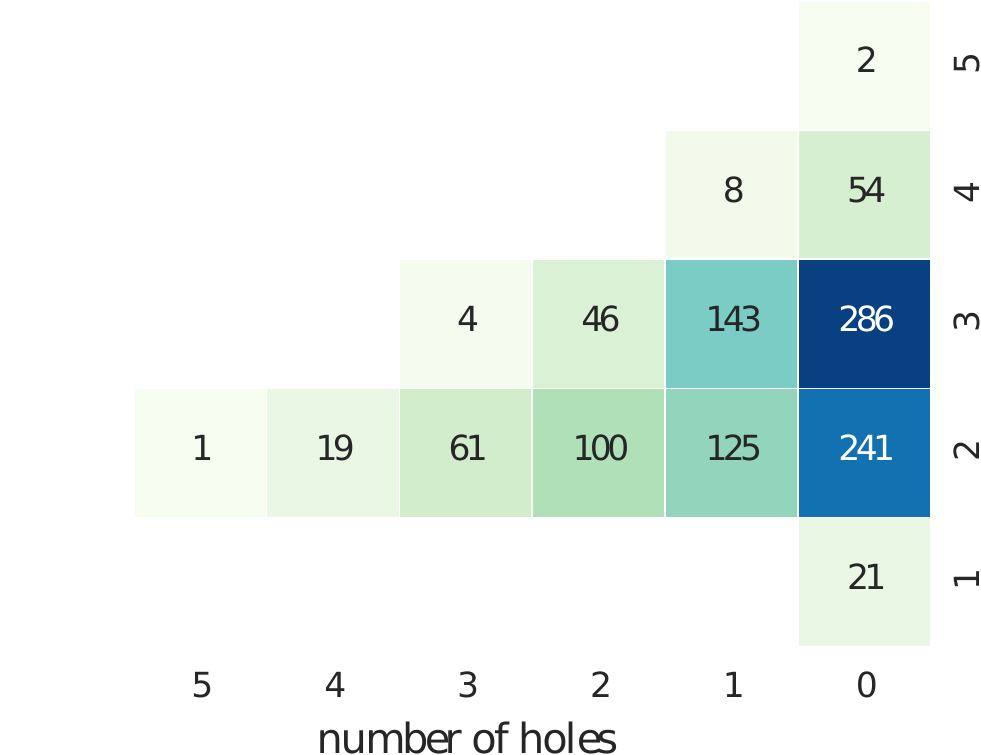}}\quad
	\subfloat[PI*C*]{
		\includegraphics[height=4.91cm]{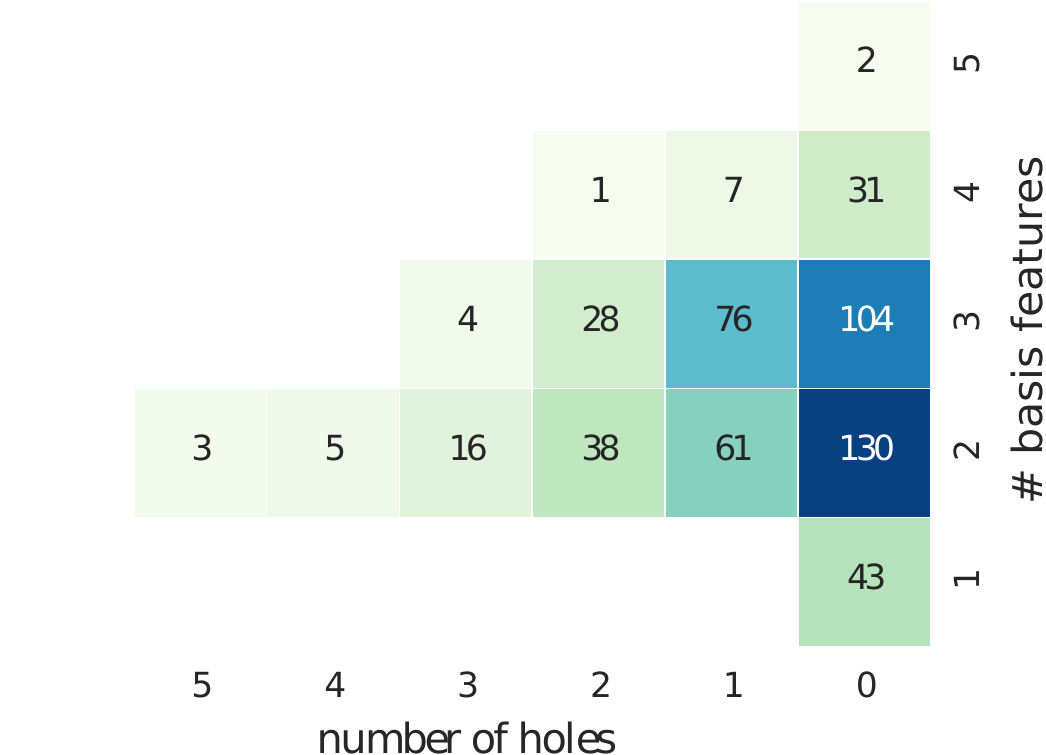}}
	\caption{\small This plot visualizes the distribution of generalized \ngram features' length and number of holes. Each row represents the length of a compound feature, each column the number of holes in the compound. The fields contain the accumulated number of features per length and number of holes. \ngram alike features have zero holes.}
	\label{fig:heatplots-holes}
\end{figure}

\subsubsection{How Much Temporal Context Does \pypulse \textit{for Music} Use?}
The temporal context is made up of the past musical events that \pulse bases its predictions on. We answer the question under consideration by analyzing \fig \ref{fig:heatplots-featuredepth}. The figure visualizes the temporal weight distribution of a set of compound features as follows: Each compound consist of one or several basis features $f_{\sigma,\nu}$ which each carry a time component $\sigma$ (for a detailed explanation see section~\ref{sec:pypulse-tef}). The basis features $f_{\sigma,\nu}$ are attributed the weight of their respective compound feature. For all basis features of all compound features, the absolute weights are accumulated in the bin for time $\sigma$ (x-axis) and the total number of basis features in its parent-compound (y-axis). The accumulated values are color encoded and plotted in a matrix. The plot is repeated for the four \nplus configurations P*, PI*C*, PI*C*K and PI*C*KM$_K$, trained on the Bach chorales dataset.

The first observation to make~--~and the answer to the question~--~is that the maximum temporal extent over all models is six steps in the past. The feature length is limited by five. The majority of the weight is assigned to features with time of two or less and with length three or less. Note that length-one features are compelled to have time zero, and that anchored and linked features were defined to have zero temporal extent. 

The differences between the four models are minuscule. We can observe that the more complex models (a) and (b) cope with shorter features of less temporal extent. Also, in (a) and (b), the length-one features are assigned a relatively higher weight, which is presumably caused by the additional length-one features K and M$_K$.

\subsubsection{Frequency of Holes in Generalized \textit{n}-gram Features}
Do generalized \ngram features have holes in practice? To answer this question we consider \fig \ref{fig:heatplots-holes}. The matrix has the two dimensions number of holes (x-axis), and number of basis features (y-axis) of compound features. The number of holes is defined as the count of missing basis features compared to the respective contiguous $n$-gram. The heat encoded values represent the count of compound features with a given number of holes and basis features. I analyze the same models as above.

The column labeled with zero holes depicts all compound features that correspond to standard $n$-grams. The values in the other columns list the number of generalized \ngram features. We can compute that the generalized \ngram version on average makes up 37\% of the feature set.

We observe further that the majority of generalized \ngrams has length two, and up to five holes. This is a comprehensible observation, as sets of generalized 2-grams are flexible and can describe the same as any longer features. \Eg ``two steps ago there was a C5, now there is a C5'' and ``one step ago there was an A4, now there is a C5'' is a versatile way of saying ``two steps ago there was a C5, one step ago there was an A4, now there is a C5''.

Up on comparing the different models we note that (a) and (b) mostly differ in the count of zero hole features with length two. The reason for this is that the linked M$_K$ features have length two but are defined to have time zero, and thus no holes. We further  observe that the number of features decreases from (c) to (d) and (d) to (a) despite that feature types have been added. This can be explained by the higher expressiveness of the added types that afford a more efficient representation and learning. \Eg repeated learning of an interval in different transpositions requires more P* than I* features.

\subsection{Sequence Generation}\label{sec:results-inference}
In the final experiment, the learned models are made audible. For this the two inference and sampling methods presented in section~\ref{sec:pypulse-inference} are applied and the generated sequences compared and discussed. The comparison in performance between the two method provides mixed results. The results of both methods resemble specimen of the respective style, although no musical extravaganzas should be expected.

\subsubsection{Experiments}
Sequences were generated from a PI*C*KM$_K$-LTM trained on the Bach chorales, German nursery rhymes, as well as Chinese folk tunes, using beam search and iterative random walk. The number of beams was set to $k=5$ and the threshold level of iterative random walk to $0.65$. All sequences were primed with the first 7 tones of a melody of the same genre that has not been part of the training set, following the example of \textcite{conklin1995}. For music generation, \textcite{whorley2016music} linked a higher audible quality with sequences of lower entropy. Resting upon these findings, entropy was used as performance measure, instead of a more difficult rule based or audible evaluation (see section~\ref{sec:evaluation-measure} for a more detailed description of evaluation measures).

\subsubsection{Results}
The best sequence inferred from the PI*C*KM$_K$ Bach chorale model attained an entropy of 1.112/1.130 bits for the beam search/iterative random walk method, the German nursery rhymes model attained 1.170/1.210 and the Chinese folk tunes model 1.529/1.314. Examples of generated melodies are provided in appendix~\ref{appendix:generated-melodies}. The results are mixed: Beam search performed slightly better in two, and iterative random walk performed much better in one out of three test cases.

In theory, iterative random walk has the advantage that it can generate an arbitrary number of sequences, whereas beam search only returns as many sequences as there are slots per initialization. In practice, however, we observe that sampled sequences with minimal entropy are almost identical in iterative random walk. Furthermore, the best results of both methods are nearly equal. In these cases, the deterministic operation of beam search becomes an advantage as it guarantees stable runtimes that are also faster than those of iterative random walk.

Various statistical patterns can be observed in the sequences: For example, the chorale models favored small steps with major seconds. Unisons were the most frequent intervals, followed by perfect fifth and fourth. The most frequent motifs were tone repetitions and perfect fifths up followed by a perfect fourth down. The melody line meandered around B4. The Chinese folk tunes model had a preference to downwards movement in the chromatic scale. In contrast to the chorales, very large intervals occurred as well.

Aurally, the generated melodies were pleasant to the ear, with the exception of the melodies that were generated from the German nursery rhymes model. In this specific model, the melodies instantly converged to a continuous repetition of the same intervals. I can only speculate at this point that this was caused by the repetitive character of the dataset.

The listener can intuit the genre of each generated melody, although the lack of note durations is making such an endeavor difficult. The melodies' nonobservance of the musical scale, despite the usage of key features, is noticeable. In summary, I conclude that the audible impression of the generated models affirms the suitability of \pypulse as base model for algorithmic composition.

\chapter{Conclusion}\label{chap:conclusion}

\section{Thesis Review}
The task of monophonic melody prediction is a subfield of musical style modeling, which involves melody, harmony and rhythm. In this thesis, the \pulse feature discovery and learning method was adapted to the realm of music to predict time series of pitches, while operating on feature spaces that are too large for standard feature selection approaches for conditional random fields (CRFs). The resulting learning algorithm outperformed the \art methods for long-term, short-term and hybrid models.  

An object oriented framework for \pulse was designed and realized for the task of music modeling, using $L_1$-regularized stochastic gradient descent (SGD) optimization. The presented framework is the best performing single-model framework for multiple viewpoint systems (MVS) to date. The achieved increase in performance for the long-term model and hybrid models is similar to that reached by \textcite{cherla_hybrid_2015} compared to \textcite{pearce2004}. For the first time, the proposed method outperformed the \art for short-term models since it was set by \textcite{pearce2004}. Furthermore, the method affords interpretable models, which were shown to reflect music theoretic insights.

Compared to standard approaches for feature selection using CRFs, the \textit{P\textsmaller{ULSE}} framework can search and select features from a feature space that is significantly larger. The explicit listing of the entire feature space is circumvented by alternatingly expanding and culling a small set of candidate features. Effectively, the \nplus operator serves as a heuristic for the search through features spaces of possibly infinite size.

On the downside, \pypulse is much slower than its \ngram competition and requires a laborious hyperparameter tuning. Thus, having a capable computing infrastructure is a prerequisite for \textit{PyPulse}'s application. 

While \pypulse was shown to outperform hybrids of long- and short-term models, it has not yet been proven to outperform ensembles of several single-viewpoint \ngram models with viewpoints corresponding to the \pulse features. The reported results give grounds for assuming that \pulse could principally outperform its $n$-gram-ensemble counterpart, but to date this is practically out of limits due to the expensive hyperparameter optimization. In addition, it is pertinent to ask, whether the proposed generalized \ngram features are more suited to melody prediction than standard $n$-grams, as the conducted experiments provided mixed results. Further, I would like to point out that the decision to utilize SGD over quasi-Newton optimization methods based on \textcite{lavergne2010practical}'s benchmarks, was made in the anticipation of very large models. The models proved to be of medium size and thus a quasi-Newton method might perform faster.

\section{Future Research}
There are several open leads for future research that were not possible to tackle in this work. First and foremost, measures to reduce the runtime require further investigation. I propose the following approaches:
	\begin{itemize}
		\item Tweaking the log-linear model and objective: The size of the prediction space is limited because the computation of the partition function $Z$ (see equation~\ref{eq:CRF}) requires the summation over the entire space which rapidly becomes very costly. Methods that avoid the computation of $Z$ are explained in \textcite{sutton2012introduction}.
		\item Tweaking the optimization: On the one hand, the speed of OWL-QN BFGS \parencite{andrew2007scalable} should be evaluated on the task of learning a musical model to either confirm or negate the superiority of SGD. On the other hand, I consider the present optimization problem in which only a small number of features is updated per step suitable to be parallelized using Hogwild \parencite{recht2011hogwild}.
		\item Tweaking hyperparameter optimization: Instead of speeding up the learning process, the hyperparameter optimization~--~being the computational \hbox{bottleneck} of the framework~--~can be targeted directly. Especially for cases with a large number of hyperparameters, the gradient descent based hyperparameter learning approach by \textcite{foo2008efficient} appears promising. Besides, \textcite{langhabel2016nips} propose a surrogate based method to optimize hyperparameters faster by transfer learning from prior optimizations with similar parameters. Surrogate based methods could speed up hyperparameter tuning significantly by generalizing over the cross-validation folds and even musical styles.
		\item Tweaking the feature matrix creator: Currently, not all viewpoint feature types are precomputed. Further precomputing will gain further improvement in performance.
		\item Changing the objective in hyperparameter optimization: Large speedups can be achieved by selecting the hyperparameters with the focus on speed instead of performance.
	\end{itemize}
From the musicological side, I deem the following augmentations and future research most interesting:
	\begin{itemize}
		\item Two improvements can be made on the utilized benchmark corpus: (a) To afford proper modeling, the rest events should be considered, and (b) the target alphabet should be a contiguous interval of pitches rather than the subset of events that occurs in the dataset.
		\item Following \textcite{pearce2010unsupervised}, it would be interesting to see if \pulse is a reliable model for neurophysiological data.
		\item Following the lead of prior research, the next step would be the application of \pulse to homophonic or polyphonic melody modeling. \pulse poses an intriguing new case as it operates on a single model in contrast to MVS that are typically implemented as ensembles of different viewpoint models.
		\item A lack of time prevented the exploration of a Metropolis-Hastings based approach to generate globally optimal sequences. Note however, that for algorithmic composition this method is suboptimal, similar to the methods analyzed in this thesis, as the likeliest sequences constitute rather conservative samples of the model. Computational creativity could be introduced by injecting events of distinguished unexpectedness using Bayesian surprise \parencite{abdallah2009information} to find the required balance between the known and unknown for the listener's ear \parencite{meyer1956emotion}.
	\end{itemize}
The model itself provides ample possibilities for future investigations:
	\begin{itemize}
		\item Currently, the model uses binary feature functions. As the underlying CRF allows the usage of arbitrary feature functions, the benefits of using real valued functions should be evaluated.
		\item Additionally, feature functions of higher complexity could be used. The choice of neural networks as feature functions appears particularly attractive as their capabilities in music related tasks were already substantiated \parencite{thickstun2016learning,cherla2015discriminative}. The \nplus operator's objective could be the discovery of network architectures. Training can be done by backpropagating through \textit{P\textsmaller{ULSE}}'s optimization gradients.
		\item The single model approach of \pulse can be complemented by learning a joint short- and long-term model, which is opposed to current approaches where both models are optimized separately and the resulting distributions are combined. This means that currently, the optimization objective is not the maximum performance of the combined but of each separate model. It would be preferable to directly optimize the joint hybrid performance.
	\end{itemize}
\vspace{0.5cm}
\begin{center}
	\huge\twonotes
\end{center}
\begin{center}
	It is my hope that the framework presented in this thesis will serve as a basis for future research on interpretable cognitive models of music using \textit{P\textsmaller{ULSE}}.
\end{center}

\newpage
\thispagestyle{empty}
\newpage
\printbibliography[heading=bibintoc, title={Bibliography}] 

\appendix

\chapter{UML Class Diagram}\label{appendix:uml}
Please view on the next page.

The colored containers are specializations of the general model: for melody prediction and time series data (green), CRF models (orange), and SGD optimization (red), respectively. A discussion of the general model can be found in section \ref{sec:pypulse-design}, the melody prediction specializations are discussed in chapter \ref{chap:pulse-for-music}.
\newpage
\begin{landscape}
\afterpage{\null\thispagestyle{empty}\newpage}
\AddToShipoutPicture*{%
	\put(60,0){%
		\parbox[b][\paperheight]{\paperwidth}{%
			\vfill
			\centering
			\includegraphics[width=.99\paperwidth,height=.99\paperheight,%
			keepaspectratio]{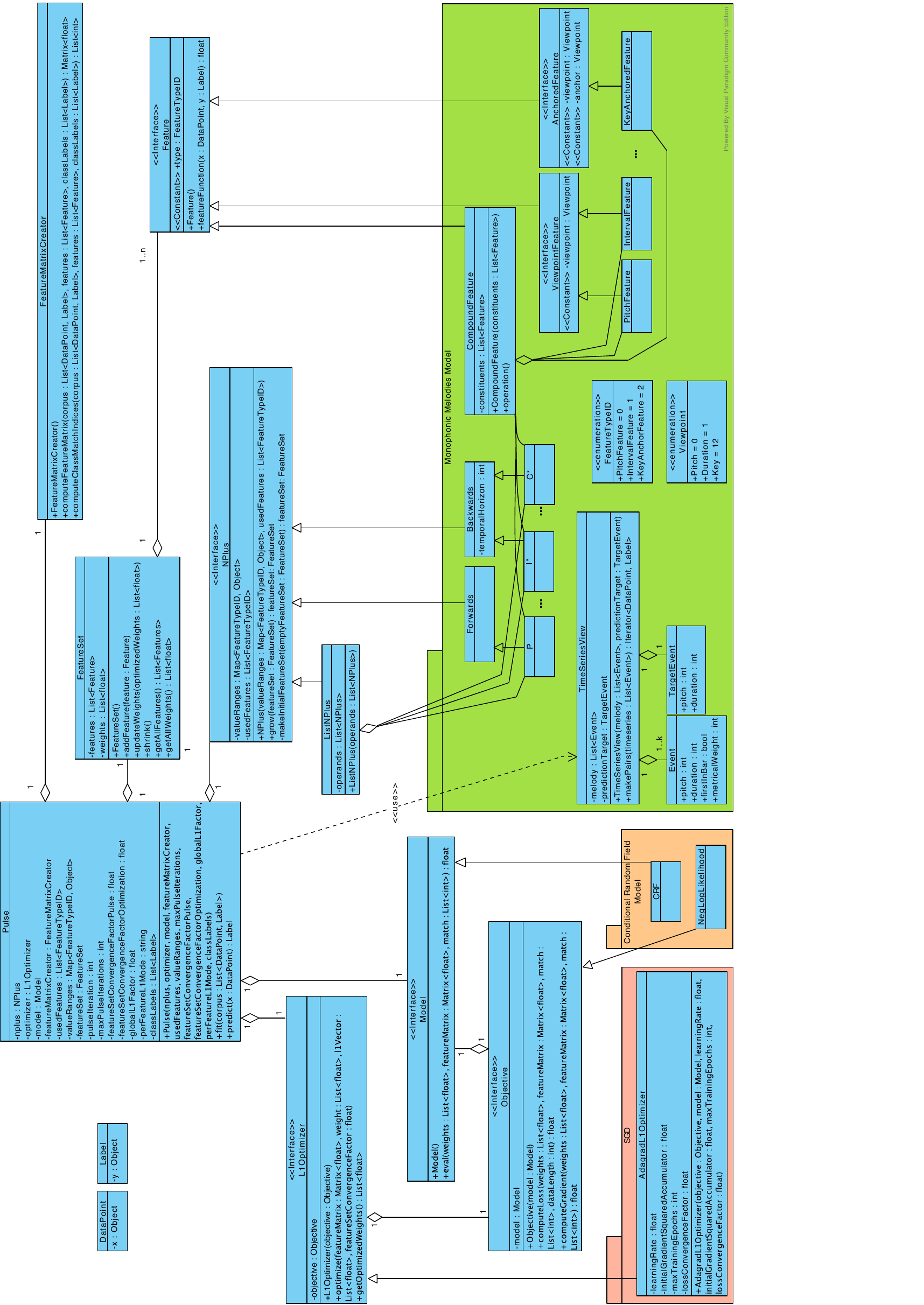}%
			\vfill
}}}
\end{landscape}

\chapter{SGD Hyperparameter}\label{app:sgd}
\Tab \ref{tab:ada} visualizes the AdaGrad and AdaDelta hyperparameter optimization as contour plots. The exact performances are stated here, serving as a reference.

\section{AdaGrad}
\begin{table}[h!] 
	\centering
	\subfloat[AdaGrad after 100 epochs]{
	\centering
	\begin{tabular}{c c c c c c c c c}
		\toprule
		& & \multicolumn{7}{c}{\textit{igsav}}\\
		\midrule
		& & $10^{-6}$& $10^{-7}$& $10^{-8}$& $10^{-9}$& $10^{-10}$& $10^{-11}$& $10^{-12}$\\
		\midrule
		\multirow{4}{*}{$\eta$} &\multicolumn{1}{r|}{0.01} & 1.7881 & 1.7286 & 1.6948 & 1.6796 & 1.6744 & 1.6732  & 1.6730 \\
		& \multicolumn{1}{r|}{0.1} & 1.6005 & 1.5747 & 1.5615 & 1.5553 & 1.5533 & 1.5529 & 1.5529   \\
		& \multicolumn{1}{r|}{1.0} & 1.5550 & 1.5516 & 1.5499 & 1.5491 & 1.5488 & \textbf{1.5487} & 1.5488   \\
		& \multicolumn{1}{r|}{10.0} & 1.6374 & 1.6426 & 1.6429 & 1.6428 & 1.6428 & 1.6428 & 1.6428   \\
		\bottomrule
	\end{tabular}}\\
	\subfloat[AdaGrad after 500 epochs]{
		\centering
		\begin{tabular}{c c c c c c c c c}
			\toprule
			& & \multicolumn{7}{c}{\textit{igsav}}\\
			\midrule
			& & $10^{-6}$& $10^{-7}$& $10^{-8}$& $10^{-9}$& $10^{-10}$& $10^{-11}$& $10^{-12}$\\
			\midrule
			\multirow{4}{*}{$\eta$} &\multicolumn{1}{r|}{0.01}& 1.6610  & 1.6287 & 1.6117 & 1.6047 & 1.6026  & 1.6021 & 1.6020\\
			& \multicolumn{1}{r|}{0.1} & 1.5568 & 1.5482 & 1.5444 & 1.5429 & 1.5425 & 1.5425 & 1.5426   \\
			& \multicolumn{1}{r|}{1.0} & 1.5411 & 1.5397 & 1.5396 & 1.5396 & \textbf{1.5395} & \textbf{1.5395} & 1.5396   \\
			& \multicolumn{1}{r|}{10.0} & 1.5849 & 1.5855 & 1.5855 & 1.5855 & 1.5855 & 1.5855& 1.5855  \\
			\bottomrule
		\end{tabular}}
	\caption{The objective values (negative log-likelihood) after 100 and 500 epochs for AdaGrad with various hyperparameter settings.}
	\label{tab:adagrad}
\end{table}
\clearpage
\section{AdaDelta}
\begin{table}[h!]
	\centering
	\subfloat[AdaDelta after 100 epochs ($\eta=1$)]{
		\centering
		\begin{tabular}{c c c c c c c c c}
			\toprule
			& & \multicolumn{7}{c}{$\epsilon$}\\
			\midrule
			& & $10^{-3}$& $10^{-4}$& $10^{-5}$ & $10^{-6}$& $10^{-7}$& $10^{-8}$& $10^{-9}$\\
			\midrule
			\multirow{4}{*}{$\rho$} &\multicolumn{1}{r|}{0.85} & \textbf{2.0105}  & \textbf{2.0105} & 2.0106 & 2.0107 & 2.0124 & 2.0286 & 2.1361 \\
			& \multicolumn{1}{r|}{0.90} & \textbf{2.0105} & \textbf{2.0105} & 2.0106 & 2.0107 & 2.0118 & 2.0227 & 2.1014   \\
			& \multicolumn{1}{r|}{0.95} & \textbf{2.0105} & \textbf{2.0105} & 2.0106 & 2.0106 & 2.0112 & 2.0167 & 2.0606  \\
			& \multicolumn{1}{r|}{0.99} & \textbf{2.0105} & \textbf{2.0105} & \textbf{2.0105} & 2.0106 & 2.0107 & 2.0118 & 2.0216 \\
			\bottomrule
		\end{tabular}}\\
	\subfloat[AdaDelta after 500 epochs ($\eta=1$)]{
		\centering
		\begin{tabular}{c c c c c c c c c}
			\toprule
			& & \multicolumn{7}{c}{$\epsilon$}\\
			\midrule
			& & $10^{-3}$& $10^{-4}$& $10^{-5}$ & $10^{-6}$& $10^{-7}$& $10^{-8}$& $10^{-9}$\\
			\midrule
			\multirow{4}{*}{$\rho$} &\multicolumn{1}{r|}{0.85} & \textbf{1.7760} &\textbf{1.7760}  &1.7761 &1.7762 &1.7772 &1.7882 &1.8775 \\
			& \multicolumn{1}{r|}{0.90}  & \textbf{1.7760}& \textbf{1.7760} & 1.7761& 1.7761& 1.7768& 1.7839& 1.8472  \\
			& \multicolumn{1}{r|}{0.95} &\textbf{1.7760} & \textbf{1.7760} & 1.7761 & 1.7761 & 1.7764 & 1.7799 & 1.8129   \\
			& \multicolumn{1}{r|}{0.99} &\textbf{1.7760} & \textbf{1.7760} & \textbf{1.7760} & 1.7761 & 1.7768 & 1.7768 & 1.7833   \\
			\bottomrule
		\end{tabular}}\\
	\caption{The objective values (negative log-likelihood) after 100 and 500 epochs for AdaDelta with various hyperparameter settings.}
	\label{tab:adadelta}
\end{table}

\chapter{Generated Melodies}\label{appendix:generated-melodies}
\section{Generated Bach Chorales}
The model used for  generation was learned from the Bach chorales dataset with the PI*C*KM$_K$-LTM. The sequences were inferred using beam search with $k=5$ slots.

\vspace{1cm}
\begin{figure}[htb]
	\centering
	\includegraphics[width=\textwidth]{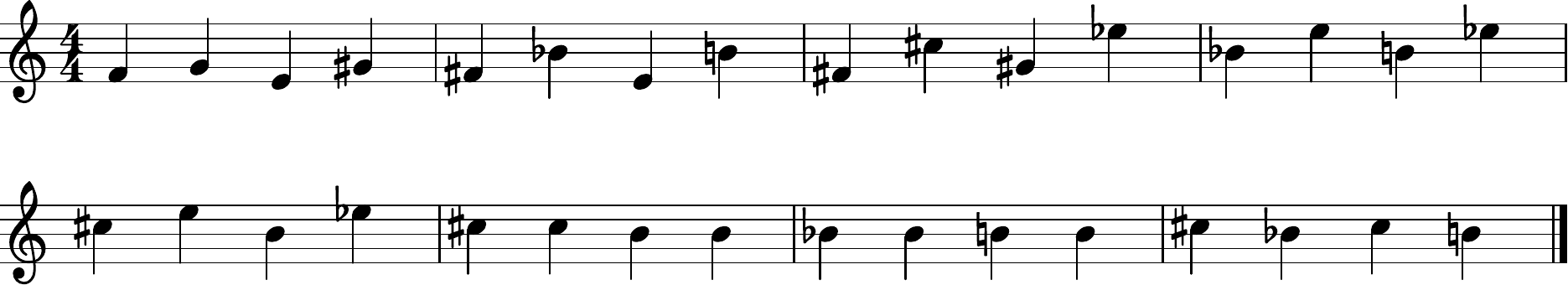}
	\caption{A generated Bach chorale initialized with start pitch value 65.}
	\label{fig:generated-chorale-no-init}
\end{figure}
\vspace{1.2cm}
\begin{figure}[htb]
	\centering
	\includegraphics[width=\textwidth]{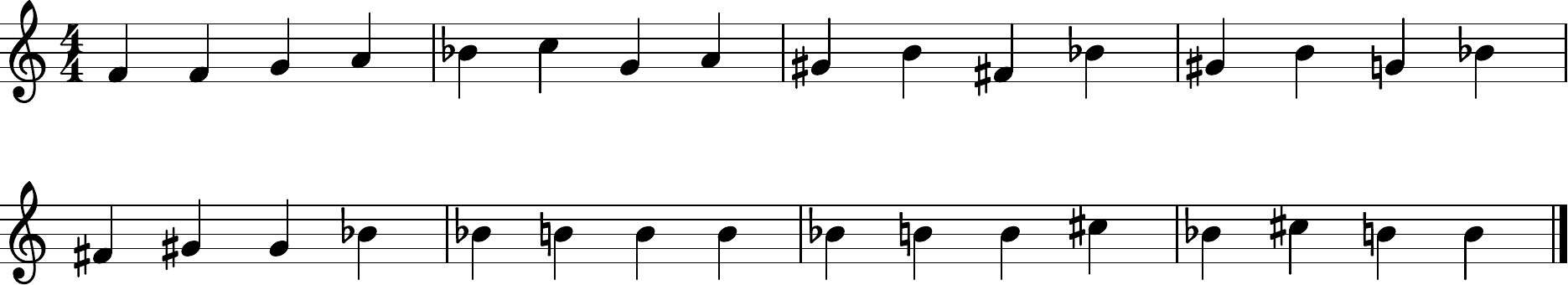}
	\caption{A generated Bach chorale initialized with the first seven tones of \textit{Meinen Jesum laß' ich nicht, Jesu} (BWV 379).}
	\label{fig:generated-chorale-7-init}
\end{figure}
\clearpage

\section{Generated Chinese Folk Melodies}
The model was learned from the Chinese folk melodies dataset, using the PI*C*KM$_K$-LTM. The sequences were generated using iterative random walk with a threshold level of $0.65$.

\vspace{1cm}
\begin{figure}[htb]
	\centering
	\includegraphics[width=\textwidth]{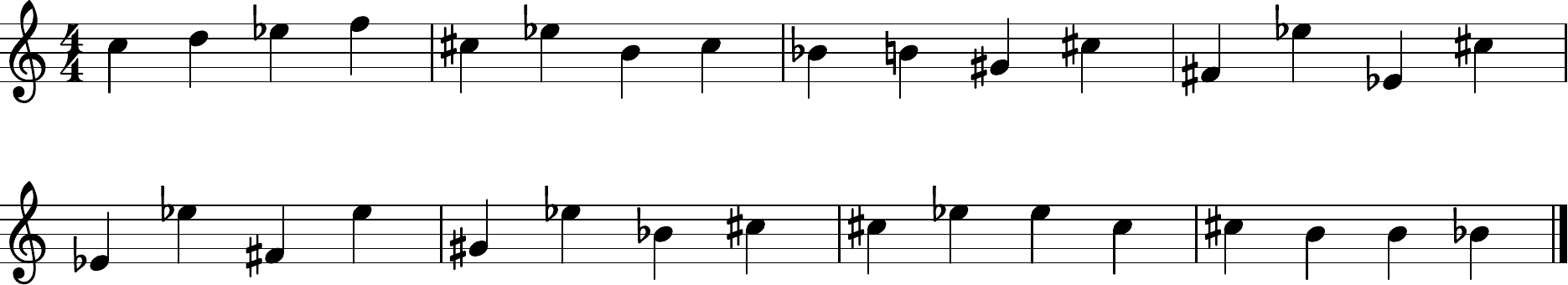}
	\caption{A generated Chinese folk melody initialized with start pitch value 72.}
	\label{fig:generated-chinese-no-init}
\end{figure}
\vspace{1.2cm}
\begin{figure}[htb]
	\centering
	\includegraphics[width=\textwidth]{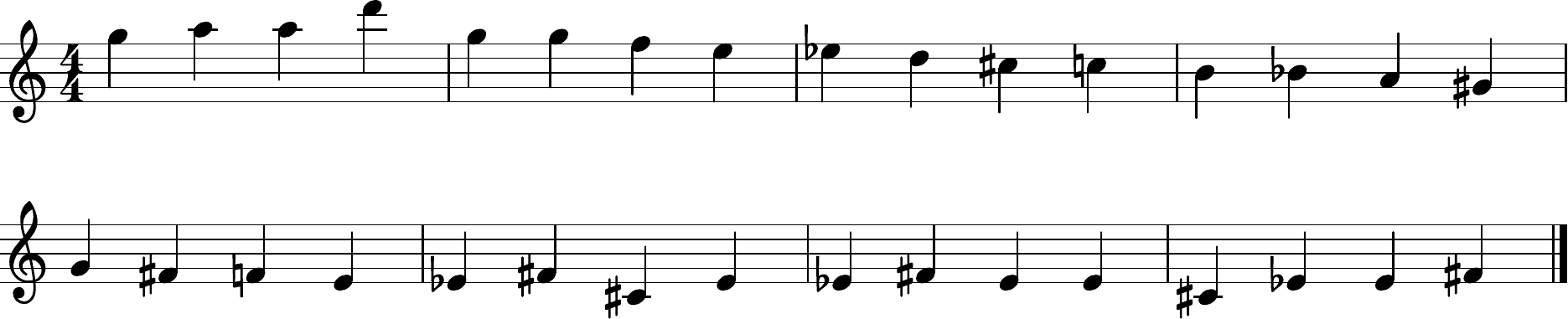}
	\caption{A generated Chinese folk melody initialized with the first seven tones of the tune \textit{Yila yiche hao nanhuo} from Hequ County, Shanxi.}
	\label{fig:generated-chinese-7-init}
\end{figure}

\end{document}